%% file: main.tex
\documentclass{article}

\usepackage[final]{corl_2026} 

\usepackage{amsfonts}
\usepackage{amssymb}
\usepackage{amsthm}
\usepackage{amsmath}
\usepackage{mathtools}
\usepackage{algorithm}
\usepackage{algpseudocode}
\usepackage{graphicx}
\usepackage{subcaption}
\usepackage{wrapfig}
\usepackage{tikz}
\usepackage{dsfont}
\usepackage{multicol}
\usepackage{multirow}
\usepackage[shortlabels]{enumitem}
\usepackage{bm}
\usepackage{xr}
\usepackage[table]{xcolor}
\usepackage{booktabs}
\usepackage{adjustbox}
\usepackage{url}
\usepackage{verbatim}

\usepackage[most]{tcolorbox}
\definecolor{promptgray}{gray}{0.95}
\definecolor{promptborder}{gray}{0.0}

\newtcolorbox{promptbox}{
  colback=promptgray,        
  colframe=promptborder,     
  boxrule=1.0pt,             
  arc=2pt,                   
  left=1em,
  right=1em,
  top=0.8em,
  bottom=0.8em,
  breakable,
}

\newcommand{\accepted}{\textit{accepted}}
\newcommand{\rejected}{\textit{rejected}}
\newcommand{\opplabel}{\operatorname{opp}}
\newcommand{\indicator}{\mathbf{1}}
\newcommand{\implied}{IMPLIED}
\newcommand{\pie}{PIE}

\newcommand{\final}[1]{\textcolor{black}{#1}}

\newcommand{\examplesymbol}[1][CDCDEE]{%
\begin{tikzpicture}[baseline=-0.3em]
    \definecolor{examplesymbolcolor}{HTML}{#1}%
    \draw[examplesymbolcolor, line width=2pt] (0,0) -- (0.3,0);
\end{tikzpicture}%
}

\title{Rethinking the Implications of Human Feedback for Preference Learning in Human-Robot Collaboration}

\author{
  Qiping Zhang,
  Kate Candon,
  Debasmita Ghose,
  Marynel V\'azquez \\
  Department of Computer Science\\
  Yale University, New Haven, CT, United States \\
  \texttt{qiping.zhang@yale.edu}
}

\begin{document}
\maketitle


\begingroup
\renewcommand{\thefootnote}{\fnsymbol{footnote}}%
\footnotetext[1]{Project website: \final{https://implied.interactive-machines.com}}%
\endgroup


\begin{abstract}
In Human-Robot Interaction, the standard approach to learn a reward model that represents human preferences for robot behavior consists of three steps. First, the robot collects limited \textit{direct} evidence from human feedback (e.g., positive or negative binary feedback). Then, the robot utilizes the direct evidence to derive accepted or rejected labels to feasible but unchosen actions using fixed implication rules. Finally, the robot updates the reward model with both the direct and derived evidence. Unfortunately,   
the fixed rule can hinder preference learning: in a user study with two collaborative simulation environments, human-provided implication labels often differed from the standard fixed rule, and using the human labels substantially improved reward learning with the Preference Learning from Implicit and Explicit Feedback (\pie{}) algorithm. Consequently, we propose \implied{}, an implication modeling method that treats fixed-rule implications as an initial guide while learning to infer and revise accepted and rejected action labels over time. Across evaluations on recorded human-robot interaction trajectories and a physical robot pizza-making study, \implied{} predicts human implications more accurately than the fixed rule approach and LLM baselines, approaching the performance of a human-label oracle. In turn, \implied{} reduces preference-estimation error and leads to robot actions that are more often rational with respect to a combined reward (which includes the true preference reward and a task-specific reward) compared to baselines. By learning to reason about the implications of human feedback, this work enables more faithful and efficient robot behavior adaptation during human-robot collaboration.

\end{abstract}


\input{sections/1.introduction}

\input{sections/2.preliminaries}

\input{sections/3.hypothesis_validation}

\input{sections/4.method}
\input{sections/5.evaluation}

\input{sections/6.related_work}

\input{sections/7.limitations}



\acknowledgments{This work was partially supported by the National Science Foundation (Grant No. IIS-2143109, IIS-2106690, and OIA-2535195). Any opinions, findings, and conclusions or recommendations expressed in this paper are those of the author(s) and do not necessarily reflect the views of the National Science Foundation (NSF). The authors are also thankful to Sasha Lew for his assistance with the robot system that was used for the real-world evaluation presented in Sec.~\ref{sec:eval_physical_robot}.}


\bibliography{references}  
\clearpage

\appendix
\normalsize \input{appendix.tex}

\end{document}

%% file: sections/1.introduction.tex
\section{Introduction}
\label{sec:intro}

There are many valid ways to complete everyday tasks, and people have different preferences about aspects such as action ordering, object placement, or division of labor.
Because preferences are often personal and difficult to specify in advance \cite{agrawal2022task}, collaborative robots must not only learn how to help a person achieve a task goal, but also how the person prefers the task to be done. 

Prior work in Human-Robot Interaction (HRI) typically assumes that robots obtain \textit{direct} evidence about a person’s preferences from two sources. A person can give \textit{explicit feedback} about robot actions to indicate if actions matched their preferences via modalities like evaluative feedback \cite{fitzgerald2022inquire, tamer2009, mindmeld2022, akrour2011preference, wilde2020improving, wang2022skill, hejna2023few}, demonstrations \cite{argall2009survey, christiano2017deep}, or corrections \cite{losey2018including, wang2025effects, bajcsy2018learning, wang2026enhancing}. 
People can also provide \textit{implicit feedback} through other communicative signals in response to a particular action \cite{ghose2023tailoring, jeon2020reward, ghose2025ve, cui2021empathic, loftin2016learning, dennler2025hri, ghose2026robots, zhang2023self, candon2024react, zhang2025predicting}.
The direct evidence indicates if a particular action matches the user's preference and should be treated as \textit{accepted}, or did not and should be treated as \textit{rejected}.
This evidence is useful, but limited. It can be difficult to know what to infer about the other feasible actions that were not taken, which we call \textit{unchosen actions}.
Following prior work \cite{fitzgerald2022inquire, jeon2020reward, candon2026learning}, we call the accept and reject labels for all actions, both chosen and unchosen, \textit{implications}.

\begin{figure}[t]
    \centering
    \includegraphics[width=1\linewidth]{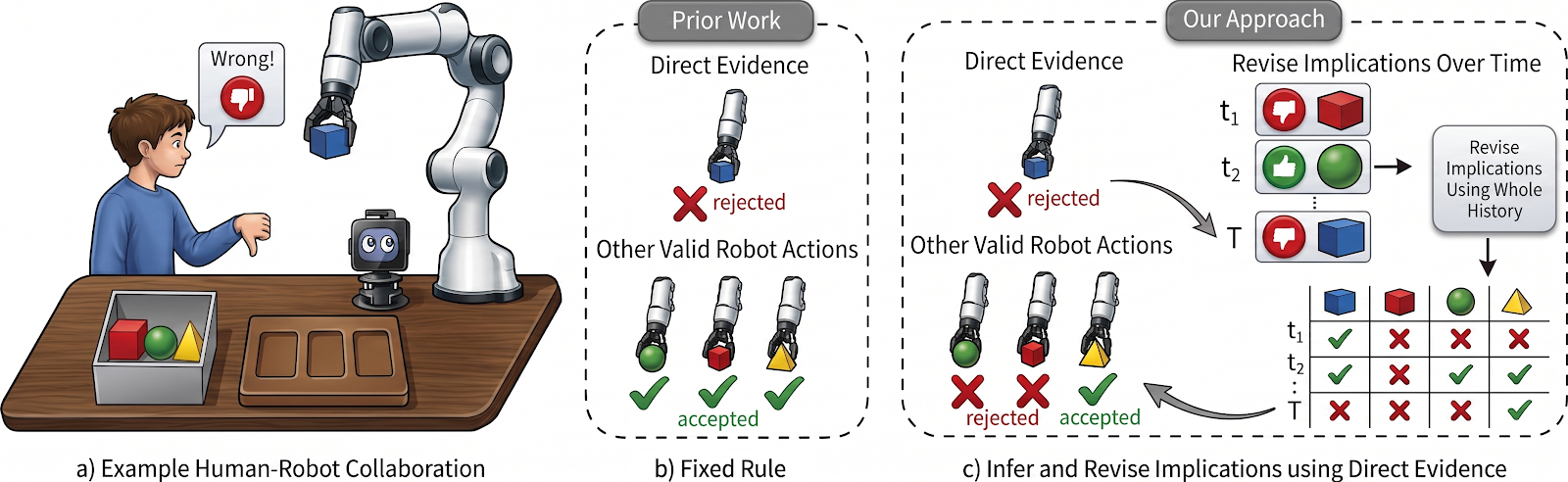}
    \caption{(a) In a collaboration, person provides direct evidence, rejecting the robot's executed action. (b) Typical implication applies fixed-rule to label other feasible actions given the direct evidence. (c) Our approach, \implied{}, uses direct evidence overtime to revise implication labels for unchosen actions.}
    \label{fig:pull}
\end{figure}

\final{Converting limited direct evidence into broader accepted and rejected feedback sets via fixed implication rules is a long-standing practice. For example, this practice is common for inverse reinforcement learning \cite{ziebart2008maximum}, modeling human choices for robot learning \cite{jeon2020reward}, interactive querying \cite{fitzgerald2022inquire}, and collaborative preference learning \cite{candon2026learning}.}
Intuitively, 
negative feedback for a robot action (Fig. \ref{fig:pull}a) could indicate that the executed action is rejected and other feasible actions are accepted (Fig. \ref{fig:pull}b). But if this assumption is wrong,  it could be difficult to learn correct preferences from noisy sets of accepted and rejected behavior. 
To investigate this possible shortcoming, we conducted a user study where people collaborated with a robot to perform two tasks in simulation. \final{The tasks involved high-level collaborative decision-making, where at each step the human and the robot each had to select from a discrete, enumerable set of high-level actions.} Participants provided direct evidence for preference learning via explicit and implicit feedback, and also labeled unchosen human and robot actions as accepted or rejected. We found evidence that the naive implication rules \final{commonly used in prior work} were inconsistent with human labels, suggesting that robots could better learn preferences if they reasoned more carefully about the unchosen actions.

To this end, we propose \implied{}, Introspective Modeling for Preference LearnIng with EviDence. \implied{} uses fixed-rule implications as a prior for generating implications for human feedback when direct evidence from the person is limited. As the interaction progresses, the robot relies more on a 
Gaussian Process classifier \cite{williams2006gaussian} trained on the direct evidence. 
Because the implications are re-computed whenever new human feedback is provided, the robot can revisit its prior decisions for which unchosen actions should be treated as accepted or rejected (see Fig. \ref{fig:pull}c). In this sense, \implied{} is introspective \cite{fox2006robot, liang2024introspective, daftry2016introspective} because it uses new evidence to revise the derived implications of past unchosen actions. 

\final{We evaluated \implied{}} across recorded human-robot interaction trajectories from our initial user study with simulation environments and in a second physical robot pizza-making study. \final{We found that } \implied{} \final{predicted} human implications more accurately than \final{the traditional} fixed-rule approach and \final{several} LLM baselines, approaching the performance of a human-label oracle.
The improved implications \final{reduced} preference-estimation error and \final{led} to robot actions that \final{were} more often rational in relation to a reward that \final{considered} the human preferences and the task's objective goal. 
\final{In summary, our contributions are threefold}:
\begin{enumerate}[labelindent=0em,leftmargin=10pt,noitemsep]
    \item We formulate implication modeling as a learnable aspect of preference learning from feedback \final{for high-level collaborative decision-making}, where robots infer accepted/rejected labels for  actions that are available but not taken.

    \item We conduct a user study to collect human implication labels for unchosen actions, showing that fixed implications often diverge from human judgments and affect preference learning.

\item We propose \implied{}, the first approach that, to our knowledge, revises implication labels for unchosen actions from accumulated direct evidence. We evaluate \implied{} against state-of-the-art baselines showing that it can facilitate preference learning.


\end{enumerate}

%% file: sections/2.preliminaries.tex
\section{Preliminaries: Learning Preferences Over a Human-Robot Collaboration}
\label{sec:preliminaries}


Consider a collaboration between a human ($H$) and a robot ($R$) on a task, where each has fixed roles, e.g., the robot may pass pizza ingredients to the human, and the human may use them to assemble a pizza. As in \citet{candon2026learning}, we study the problem of learning a reward function that captures the human's preferences over the collaboration. 
This learning setup can be seen as a generalization of learning preferences over robot behavior only (e.g., as in ~\cite{fitzgerald2022inquire, jeon2020reward}), as it considers both (explicit) human feedback about the robot's behavior and (implicit) feedback about the human's behavior.

\textbf{Collaborative Setup. } We view the interaction as a Markov Decision Process (MDP) $ \langle \mathcal{S}, \mathcal{A}, \mathcal{T}, R \rangle$, where  
$\mathbf{s} \in \mathcal{S}$ is the state space, 
$\mathcal{A} = \mathcal{A}_H \times \mathcal{A}_R$ is the action space with $a_H \in \mathcal{A}_H$ and $a_R \in \mathcal{A}_R$  the simultaneous high-level actions taken by the human and robot,  
$\mathcal{T}(\mathbf{s}'|\mathbf{s}, a_H, a_R)$ is a deterministic transition function from $\mathbf{s}$ to $\mathbf{s}'$ given the actions, and $R(\mathbf{s}, a_H, a_R)$ is a shared, total reward that the team seeks to maximize.
Building off of Cooperative Inverse Reinforcement Learning \cite{hadfield2016cooperative}, 
$R(\mathbf{s},a_R,a_H) = R^\text{goal}(\mathbf{s},a_R,a_H) + \gamma R^\text{pref}(\mathbf{s},a_R,a_H; \mathbf{w})$, where $\gamma$ controls the relative importance of a task-specific goal reward ($R^\text{goal}$) and a reward that describes the human's preferences for the collaboration ($R^\text{pref}$), which is parameterized by $\mathbf{w} \in \mathbb{R}^d$. Because of the fixed-role assignments, the goal and preference reward are assumed to be decomposable into individual rewards~\cite{candon2026learning}, such that the human acts to maximize $R_H(\mathbf{s},a_H) = R_H^\text{goal}(\mathbf{s},a_H) + \gamma R_H^\text{pref}(\mathbf{s},a_H; \mathbf{w})$, and the robot should maximize $R_R(\mathbf{s},a_R) = R_R^\text{goal}(\mathbf{s},a_R) + \gamma R_R^\text{pref}(\mathbf{s},a_R; \mathbf{w})$. 
Importantly, the task goal is public information, so both $H$ and $R$ know  $R_H^\text{goal}$ and $R_R^\text{goal}$. Further,  
the human and robot also know which features $\phi$ matter to describe the preference reward, but only the human knows the preference parameters $\mathbf{w}$ for the collaboration. During the interaction, the robot has to learn $R^\text{pref}(\mathbf{s},a_R,a_H;\mathbf{w}) = 
R^\text{pref}_R(\mathbf{s},a_R;\mathbf{w}) + R^\text{pref}_H(\mathbf{s},a_H;\mathbf{w}) = 
\mathbf{w}^\intercal (\phi_R(\mathbf{s},a_R) + \phi_H(\mathbf{s},a_H))$ by estimating $\mathbf{w}$, where $\phi_R$ and $\phi_H$ map the state and individual actions to the preference feature space. 



\textbf{Preference Learning from Implicit and
Explicit Feedback.} \citet{candon2026learning} proposed the \pie{} algorithm for estimating the preferences considering: 1) \textit{explicit} human feedback about the robot's actions, provided as binary evaluative feedback (although other types of explicit feedback could be possible~\cite{fitzgerald2022inquire,jeon2020reward}); and 2) \textit{implicit} human feedback, provided via the human's actions in the interaction. 
The feedback generates what we term \textit{direct evidence} for chosen actions by the collaborators ($R$ or $H$): it leads to assigning the chosen actions in a given state $\mathbf{s}$  to an accepted $(f^+_{\mathbf{s},\text{m}})$ or rejected $(f^-_{\mathbf{s},\text{m}})$ feedback set based on the feedback modality, either explicit $(m={exp})$ or implicit $(m={imp})$. 
Because direct evidence is limited, preference learning algorithms, including \pie{}, also \textit{derive evidence} for unchosen actions. They assign other feasible actions that could have been taken in the state when the feedback was provided to the accepted or rejected sets. The process of classifying state-action observations into the feedback sets is often referred to as the ``implication" of human feedback~\cite{fitzgerald2022inquire}, and is  not considered a learnable aspect of preference learning in prior work. 

As is common in the literature (e.g.,~\cite{fitzgerald2022inquire,jeon2020reward,ziebart2008maximum}),  \pie{} uses a fixed set of rules to implement the implication of human feedback. 
%
Consider a timestep in the collaboration that leads to a state $\mathbf{s}$ where the robot takes action $a_R$ and the human takes  $a_H$. For explicit feedback, if the human indicates that $a_R$ is acceptable via a positive button press, then \pie{} adds that state-action observation $(\mathbf{s},a_R)$ to the accepted  set $(f^+_{\mathbf{s},\text{exp}})$. Also, all other possible state-action observations $(\mathbf{s},a)$ for unchosen but feasible robot actions in that state, $a \in \mathcal{A}_R(\mathbf{s}) \setminus \{a_R\}$, are added to the rejected set $(f^-_{\mathbf{s},\text{exp}})$. Conversely, if the human indicates unacceptable robot behavior via a negative button press,  $(\mathbf{s},a_R)$ is added to the rejected set  $(f^-_{\mathbf{s},\text{exp}})$, and the state-action observations for other possible robot actions are added to the accepted set $(f^+_{\mathbf{s},\text{exp}})$. For implicit feedback, the assumption is that the human acts optimally relative to $R_H$. The $(\mathbf{s},a_H)$ observation gets added to the accepted set $(f^+_{\mathbf{s},\text{imp}})$. Meanwhile, the rejected set $(f^-_{\mathbf{s},\text{imp}})$ consists of the observations $(\mathbf{s},a)$ for all other possible human actions, $a \in \mathcal{A}_H(\mathbf{s}_t) \setminus \{a_H\}$. 

Finally, \pie{} estimates the  weights $\mathbf{w}$ using a non-parametric belief distribution computed via gradient ascent, optimizing for the log-likelihood ($\log \mathcal{L}(\mathbf{w})$) of the accepted behavior: 
\begin{align}
{\small
    \mathcal{L}(\mathbf{w}) = \prod_{(f^+_{\mathbf{s},\text{m}},\, f^-_{\mathbf{s},\text{m}}) \in \mathbf{F}} P \big( f^+_{\mathbf{s},\text{m}}|\mathbf{w}\big) = \prod_{(f^+_{\mathbf{s},\text{m}},\, f^-_{\mathbf{s},\text{m}}) \in \mathbf{F}} \frac{\sum_{(\mathbf{s},a) \in f^+_{\mathbf{s},m}}B_\text{m}(\mathbf{s},a)}{\sum_{(\mathbf{s},a) \in f^+_{\mathbf{s},m} \cup f^-_{\mathbf{s},m} }B_\text{m}(\mathbf{s},a)}
}
    \label{eq:MLE-objective}
\end{align}
where $B_\text{m}(\mathbf{s}, a) = 
\exp({\hat{\beta}_\text{m}(R^\text{goal}_\text{c}(\mathbf{s}, a) + \gamma R^\text{pref}_\text{c}(\mathbf{s}, a) )})$ 
is the exponential component of the Boltzmann rationality model, the collaborator $c$ is either $H$ (for $m={imp}$ feedback) or $R$ (for $m={exp}$ feedback), and the action $a$ is the corresponding human or robot action in state $\mathbf{s}$.

%% file: sections/3.hypothesis_validation.tex
\section{The Implications of Real Human Feedback in H-R Collaborations}
\label{sec:study}

With approval from our Institutional Review Board (IRB), we recruited 18 participants to collaborate with a robot in two MuJoCo environments adapted from RoCo simulation components~\cite{mandi2024roco}. 
Half of the participants were assigned to a Fixed-Rule condition, where the robot estimated preferences using the standard implicature rule (as in Sec.~\ref{sec:preliminaries}) after each timestep. 
The other half were assigned to the Human Implication condition, where the robot estimated preferences using the same learning method, but with participant-provided implication labels. Thus, the condition affected the feedback sets used by the robot in the interaction, which in turn shaped its preference updates and behavior. 


\textbf{Study procedure.}
Participants completed sandwich-making and block-sorting tasks, 
each for three episodes with different initializations. \final{These tasks involved high-level decision-making, a setting that is common in human-robot collaboration \cite{ghose2025ve, candon2026learning, hadfield2016cooperative}, where interaction data is scarce and actions must be coarse enough for the partners to communicate and reason about them.}

In sandwich-making, the objective was to assemble a target sandwich; 
in block-sorting, the objective was to move all blocks into bins. 
Both tasks require sequential collaboration where task progress and latent preferences jointly determine desirable behavior, while differing in object semantics and action structure. Preferences could concern, for example, ingredient order, workspace occupancy, or color/shape-to-bin mappings. As commonly assumed in HRI (e.g., ~\cite{fitzgerald2022inquire,candon2026learning}), we assumed that each participant had stable preferences. 

Before each task, participants selected a true preference  (such as placing cheese before bread or sorting blue blocks into the red bin) and used it consistently when acting and evaluating the robot. The robot knew the task goal reward but not the weights that captured the participant's true preference. After each feedback step, the robot updated its preference estimate using gradient ascent on Eq.~(\ref{eq:MLE-objective}) and selected later actions based on its latest preference estimate.

At each timestep, the participant and robot each selected one feasible action. Then, after both actions were executed, the participant evaluated the robot's action as \accepted{} or \rejected{}, considering both task progress and alignment with their preference. The participant then labeled each of the robot's unchosen feasible actions as well, and finally labeled their own unchosen feasible actions using the same criterion. These labels served as human \textit{ground-truth} implications, indicating how direct evidence at that timestep should extend to feasible actions that were not taken by the agents. See Appx.~\ref{sec:mujuco_design} for further details on the study and task design.

\textbf{Implication inconsistency.}
We compared the fixed-rule and human-provided implication labels with two metrics:  \emph{timestep inconsistency}, which is the percentage of timesteps with at least one label mismatch for unchosen actions; and the \emph{label inconsistency}, which is the percentage of individual unchosen-action labels that do not match across the interaction history. Both were computed per participant and reported as mean $\pm$ standard error. Considering implications from both explicit feedback on robot actions and implicit feedback based on human actions, the sandwich-making task
had  $0.27\pm0.03$ timestep inconsistencies and $0.21\pm0.02$ label inconsistencies; meanwhile, block-sorting had $0.41\pm0.02$ and $0.22\pm0.01$, respectively. These results show that human judgments about unchosen actions differ substantially from the fixed implication rule.

\begin{wrapfigure}{r}{0.5\textwidth}
    \centering
    \includegraphics[width=\linewidth]{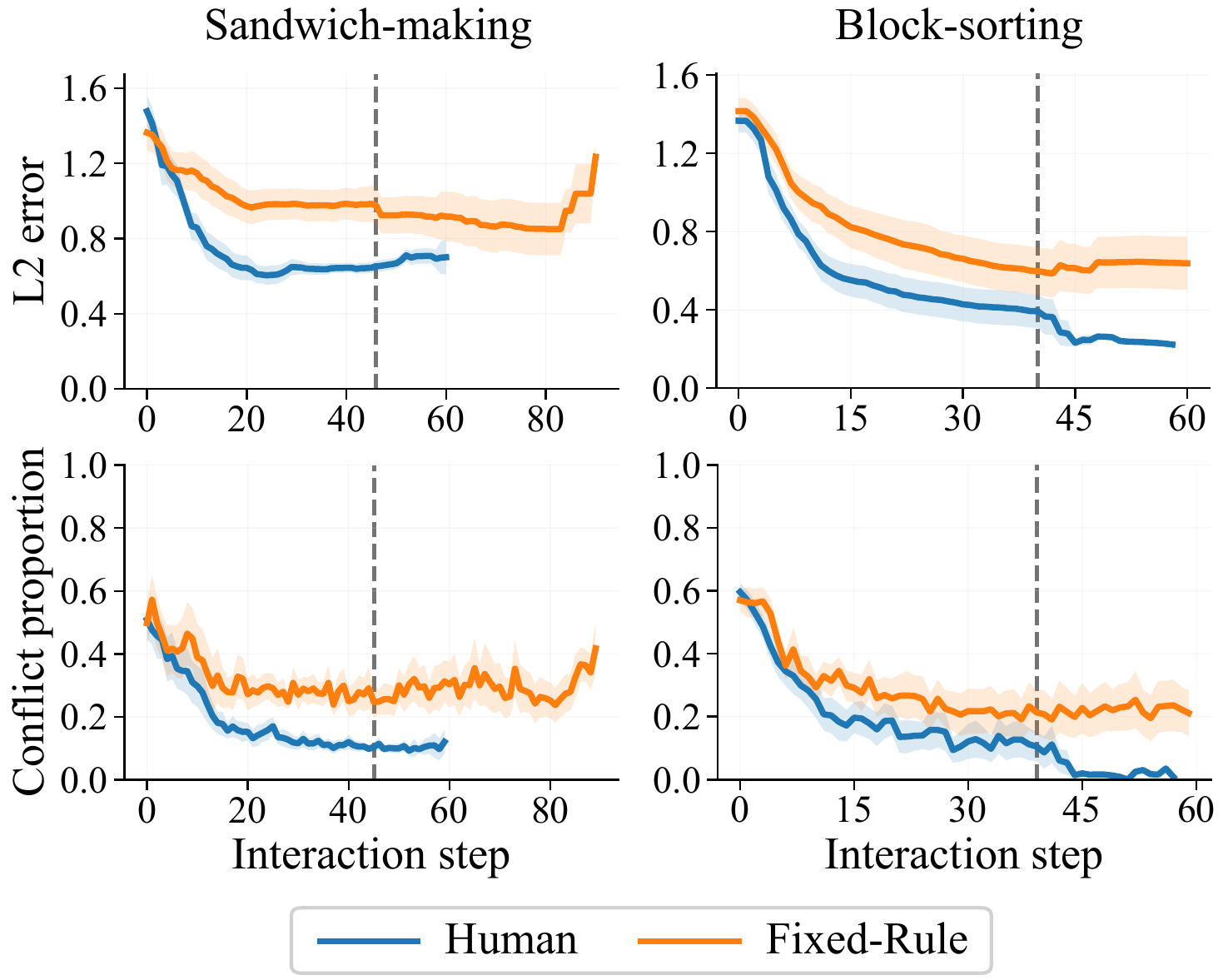}
    \caption{
    Preference-learning performance, with mean and std. error across participants. Human implications improved learning performance.
    The dashed line indicates the earliest task-completion step per task. 
    }
    \label{fig:task_metric_curves}
    \vspace{-1em}
\end{wrapfigure}

\textbf{Human implications improve preference learning.}
We tested whether differences in implications affected preference learning. We compared the two conditions using the \textit{L2  error} between the normalized preference weight estimates and the ground-truth preferences. We also computed the rollout \textit{conflict proportion}, i.e., the fraction of rollout robot actions that would not maximize the combined robot reward under the participant's true preferences.  
Lower values in Fig.~\ref{fig:task_metric_curves} indicate more accurate preferences and more aligned robot behavior.

At the earliest task-completion step across participants (vertical dashed line in Fig.~\ref{fig:task_metric_curves}) in the sandwich-making task, linear mixed-effects analyses 
showed that Human Implication significantly reduced sandwich-making L2 error relative to Fixed-Rule ($0.65\pm0.03$ vs. $0.97\pm0.10$, $p<0.01$). The same was observed for conflict proportion ($0.10\pm0.01$ vs. $0.25\pm0.05$, $p<0.05$). In block-sorting, Human Implication also reduced L2 error ($0.39\pm0.08$ vs. $0.60\pm0.12$) and conflict proportion ($0.11\pm0.05$ vs. $0.21\pm0.07$), though these differences were not statistically significant.
Together, these findings motivate considering the implications of human feedback as a learnable component of preference learning. 

%% file: sections/4.method.tex
\section{The IMPLIED Method for Inferring and Revising Implications}
\label{method}

In prior work~\cite{fitzgerald2022inquire,jeon2020reward,candon2026learning}, the feedback sets ($f^+_{\mathbf{s},\text{m}}, f^-_{\mathbf{s},\text{m}}$) used for preference learning via Eq.~(\ref{eq:MLE-objective}) are generated by a fixed implication rule; however, Sec.~\ref{sec:study} showed that these fixed-rule implications can differ from human judgments. Thus, we propose to instead treat the implications of unchosen actions as a learnable aspect of preference learning. 
Our insight is that \textit{the more direct evidence the robot receives during the collaboration, the better it should be at inferring accepted or rejected labels for unchosen actions}. To explain how this inference process works, we re-define the format of the elements in the accepted and rejected feedback sets: they now contain  state-action pairs, as in Sec.~\ref{sec:preliminaries}, along with metadata useful for reasoning about the implications of human feedback:
\begin{equation}
    \mathbf{o} = (\overbrace{\mathbf{s}, a,}^{\text{observation}} \overbrace{c, m, t, e}^{\text{metadata}} \ )
    \label{eq:element-in-feedback-set}
\end{equation}
The metadata  stores information about which collaborator $c \in \{H, R\}$ took or could take the action $a$, which feedback modality $m \in \{exp, imp\}$ led to the state-action observation, the timestep $t$ when the observation was generated, and the type of evidence $e \in \{direct, derived\}$ that was used to classify $\mathbf{o}$ into the feedback sets. Observations from chosen actions by the collaborators correspond to $e=direct$ evidence; those for unchosen actions are $e=derived$ evidence. 

Our method, Introspective Modeling for Preference LearnIng with EviDence (\implied{}), labels observations  $\mathbf{o}$ of chosen actions as accepted or rejected per the direct evidence. For example, for explicit feedback, if a human indicates a robot's action is acceptable, then the  observation is added to the accepted feedback set. However, for observations of actions that were not chosen by the collaborators (i.e., the derived evidence), \implied{} estimates the probability that any such observation $\mathbf{o}$ belongs in the accepted set ($f^+_{\mathbf{s},\text{m}}$) given \textit{all} prior direct evidence $\mathcal{D}$, $P(\mathbf{o} \in f^+_{\mathbf{s},\text{m}} | \mathcal{D})$. Importantly, the set $\mathcal{D}$ includes both observations and their corresponding labels, indicating whether the observations belong to the accepted or rejected feedback set. For example, if the interaction has reached the $T$-th timestep, then $\mathcal{D}$ is: 
\begin{align}
    \mathcal{D} &= \{ (\mathbf{o}', y') \,|\, \mathbf{o}' = (\mathbf{s}',a',c',m',t',e') \in f^+_{\mathbf{s}',m'} \cup f^-_{\mathbf{s}',m'} \wedge  1 \leq t' \leq T \wedge e' = direct\}\label{eq:D}\\ & \text{where}\ y' \in \{+, -\}\ \text{is the observation label for } \mathbf{o}' \text{ per the direct evidence}
    \nonumber
\end{align} 
Only if the probability $P(\mathbf{o} \in f^+_{\mathbf{s},\text{m}} | \mathcal{D})$ is greater than a nominal threshold ($\tau=0.5$ in our experiments), \implied{} assigns the observation $\mathbf{o}$ for an unchosen action to the accepted set; otherwise, it is assigned to the rejected set ($f^-_{\mathbf{s},\text{m}}$). This classification process happens every time new feedback is received, so the derived evidence can change over time as the interaction progresses.

\noindent
\textbf{Classifying observations of unchosen actions.} \implied{} defines $P(\mathbf{o} \in f^+_{\mathbf{s},\text{m}} | \mathcal{D})$ for an observation $\mathbf{o}$ corresponding to an unchosen action as a mixture distribution. This mixture uses the standard fixed-rule implication~\cite{ fitzgerald2022inquire, jeon2020reward, candon2026learning} as a structured prior. Also, it takes into consideration the probability that the observation belongs into the accepted set according to a learnable classifier trained on all the available direct evidence $\mathcal{D}$ that the robot has thus far in the collaboration. The mixture is:
\begin{equation}
    P(\mathbf{o} \in f^+_{\mathbf{s},\text{m}} | \mathcal{D}) = (1 - \lambda(\mathbf{o}))\, P_{\text{Fixed}}(\mathbf{o} \in f^+_{\mathbf{s},\text{m}} | \mathcal{D}) + \lambda(\mathbf{o})\, P_{\text{Learned}}(\mathbf{o} \in f^+_{\mathbf{s},\text{m}} | \mathcal{D})
    \label{eq:IMPLIED-prob}
\end{equation}
where $\lambda(\mathbf{o})$ is an observation-specific weight that balances the contributions of $P_{\text{Fixed}}$ and $P_{\text{Learned}}$. 

We implement $P_{\text{Fixed}}$ in Eq.~(\ref{eq:IMPLIED-prob}) via delta functions $\mathbb{I}$  whose output depends on the direct evidence obtained when the observation $\mathbf{o} = (\mathbf{s},a,c,m,t,e)$ for an unchosen action $a$ was generated. First, we search for the element $(\mathbf{o}',y')$ in $\mathcal{D}$ (as in Eq.~(\ref{eq:D})) that has $\mathbf{o'} = (\mathbf{s}',a',c',m',t',e')$ and the same timestep $t'=t$, state $\mathbf{s}'=\mathbf{s}$, feedback modality $m'=m$, and collaborator $c'=c$ as the observation $\mathbf{o}$. Then, if $a'=a$, we assign the label y' to $\mathbf{o}$, $P_{\text{Fixed}} = \mathbb{I}(y' = +)$. Otherwise, if $a'\not=a$, we assign the opposite label, 
$P_{\text{Fixed}} = \mathbb{I}(y' \not= +)$.


The probability $P_{\text{Learned}}$ in Eq.~(\ref{eq:IMPLIED-prob}) is computed with a Gaussian Process (GP) classifier \cite{williams2006gaussian,mackay1998introduction,matthews2017gpflow} trained using data derived from $\mathcal{D}$. We chose a GP because it can learn quickly during interactions and provides a principled mechanism to measure epistemic uncertainty, i.e., uncertainty due to a lack of
classifier knowledge~\cite{hullermeier2021aleatoric}. We use the uncertainty to adapt the weight $\lambda$ in Eq.~(\ref{eq:IMPLIED-prob}), such that when the GP uncertainty is high, \implied{} favors the fixed prior rather than the learned probability. 

Specifically, the GP represents $P_{\text{Learned}}$ via a function $h_\theta: \mathcal{Z} \rightarrow [0,1]$ that maps features $z(\mathbf{o})$ of an observation to the posterior probability that $\mathbf{o}$  should be treated as accepted, i.e., included in $f^+_{\mathbf{s},m}$. For the features, we use the goal reward derived from the state-action pair in $\mathbf{o}$ and the corresponding collaborator preference features $\phi_c$, $z(\mathbf{o}) = z((\mathbf{s},a,c,m,t,e)) = [R^\text{goal}_c(\mathbf{s}, a), \ \phi_c(\mathbf{s},a)]^T$. We made this choice because both the goal reward and $\phi$ matter for the MDP's combined reward $R$ (see Sec.~\ref{sec:preliminaries}). From an implementation perspective, the GP's function $h_\theta = \sigma \circ f$ is a composition of a sigmoid likelihood function, $\sigma(u)=1/(1+\exp(-u))$, and a latent function $f \sim \mathcal{N}(\mathbf{0},K)$ with a linear kernel $K$ with an additive bias component. For inference, because the sigmoid likelihood is non-Gaussian, we perform Monte Carlo integration with $M$ samples. We use their mean
$\mu(\mathbf{o})=\frac{1}{M}\sum_{r=1}^M h_\theta(z(\mathbf{o}))$ as $P_{\text{Learned}}$, and their variance
$v(\mathbf{o})=\frac{1}{M}\sum_{r=1}^M(h_\theta(z(\mathbf{o}))-\mu(\mathbf{o}))^2$ as the classifier's confidence.
Finally, we set $\lambda$ in Eq.~(\ref{eq:IMPLIED-prob}) such that it is inversely proportional to the classifier's uncertainty (modeled by the variance $v$) and bounded to $[0,1]$, $\lambda(\mathbf{o}) = 1-4v(\mathbf{o})$. See Appx.~\ref{sec:model_design} for the derivation of the equation for $\lambda$ and more implementation details for the GP.

%% file: sections/5.evaluation.tex
\section{Experiments and Results}
\label{sec:evaluation}

\subsection{Evaluation on Human-in-the-Loop Simulation Data}
\label{sec:eval_mujoco}

We first evaluate how well \implied{} predicts the implications of human feedback and whether it improves downstream preference learning using the interaction trajectories from the user study from Sec.~\ref{sec:study}. This evaluation considers both implicit and explicit human feedback, as described previously. However, unlike during data collection, we hold the observed states, actions, and direct evidence fixed, and compare how different implication methods (\implied{} and the baselines described below) reconstruct accepted and rejected feedback sets for preference learning compared to a human's ground-truth implication labels. 

\final{Note that IMPLIED only changes how the feedback sets are constructed, so it is not tied to PIE: it applies to any preference-learning algorithm that derives implications for feasible but unchosen actions. See Appx.~\ref{sec:implicit_explicit_only} for the results of \implied{} when preference learning considers only implicit human feedback, or only explicit feedback. Furthermore, it is possible to use a different classifier than a GP for revising implications. For example, Appx.~\ref{app:knn}  provides results with a K-Nearest Neighbors (KNN) classifier. However, we have got the best results with a GP, as described in Sec.~\ref{method}; thus, our evaluation focuses on comparing \implied{} using a GP classifier against the baselines.}

\textbf{Baselines.}
We compare \implied{} against several baselines for inferring the implications of human feedback. The first is the \texttt{Fixed-Rule}  implication approach from prior work (e.g., from \pie{}~\cite{candon2026learning}, as described in Sec.~\ref{sec:preliminaries}), which tests how implication prediction and downstream preference learning perform with a fixed heuristic for assigning observations to feedback sets. The second baseline is \texttt{No-Prior}. Since \implied{} combines a structured fixed-rule prior with a GP-based prediction in Eq.~(\ref{eq:IMPLIED-prob}), this ablation removes the structured prior and uses only the GP classifier to predict implications for unchosen actions. 
Further, 
we compare \implied{} against four LLM baselines (implemented using Gemini 3.0 Flash) that copy the accepted/rejected labels from the direct evidence for chosen actions, but predict with the large model the labels for unchosen actions. These baselines serve to evaluate if a general-purpose LLM can use task context and broader world knowledge about human preferences to generate derived evidence. The LLM baselines receive different amounts of information when classifying unchosen actions: \texttt{LLM} receives only the current 
world state $\mathbf{s}$, the executed $a_H$ and $a_R$ actions, and explicit feedback on the robot action at that timestep; the \texttt{LLM-B} baseline additionally receives a description of the preference-belief, feature by feature; \texttt{LLM-H} receives the interaction history, including previous states, actions, and feedback, on top of the information provided in the \texttt{LLM} baseline; and \texttt{LLM-HB} receives both the prior interaction history and a description of the preference-belief. See Appx.~\ref{sec:llm_prompts} for the prompts used for each LLM baseline.
Finally, we also compare \implied{} against using Human oracle implications for preference learning. The Human implications were collected directly from study participants (per Sec.~\ref{sec:study}).

\textbf{Evaluation metrics.} Since participants complete the task in different numbers of interaction steps, we evaluate all implication-prediction methods at the earliest task-completion timestep across participants, i.e., the largest timestep for which interaction data is available for every participant. We consider three metrics: 1) average \textit{F1 classification score} across the study participants, considering all the labels predicted for unchosen actions up to the earliest time-completion step;  2) the average \textit{L2  error} between the ground-truth preference weights $\mathbf{w}$ and the normalized, average preference weight from the preference-belief; and 3) the average \textit{rollout conflict proportion}, defined as the fraction of rollout robot actions that do not maximize the combined robot reward $R_R$ under the true preference. Higher values are better for F1, and lower values are better for the latter two metrics. 

\input{sections/simulation_table}

\textbf{How well do methods classify state-action observations (with unchosen actions) as accepted or rejected?}
Fig.~\ref{fig:implication_f1_curves} shows the 
F1 classification scores over time---the Figure excludes results for the Human implications as they involve no inference. \implied{} (orange line) generally led to higher F1 score faster than the non-human baselines.
Also, at the earliest task-completion step across participants (vertical line in Fig.~\ref{fig:implication_f1_curves}), \implied{} achieved the highest  F1-score  in both tasks. In sandwich-making, linear mixed-effects analysis showed that \implied{} ($0.84\pm0.01$) had significantly higher F1-score  at the earliest task-completion step 
($p<0.01$). Similarly, \implied{} ($0.87\pm0.01$) significantly outperformed the non-human baselines in block-sorting at that step 
($p<0.001$). 
Thus, our method inferred human judgments for unchosen actions more accurately than non-human baselines.

\textbf{How well do methods predict the human's preferences?}
 Table~\ref{tab:simulation_downstream_results} shows that preference learning (per Eq.~(\ref{eq:MLE-objective})) with \implied{} closely approaches learning with ground-truth human implications. 
 A linear mixed-effects analysis for the sandwich-making task ($p<0.001$) showed that \implied{} led to significantly lower L2 error than all non-human baselines, except for \texttt{LLM-HB}. Also, \implied{} led to significantly lower conflict proportion than all non-human baselines ($p<0.05$). In block-sorting, \implied{} 
 significantly outperformed all other non-human baselines on  L2 error ($p<0.001$) and conflict proportion ($p<0.05$). These results suggest that implication revision improves learning. 

\subsection{Evaluation in Real-World Human-Robot Collaboration}
\label{sec:eval_physical_robot}

\begin{wrapfigure}{r}{0.3\textwidth}
  \centering
  \vspace{-3.5em}
  \includegraphics[width=\linewidth]{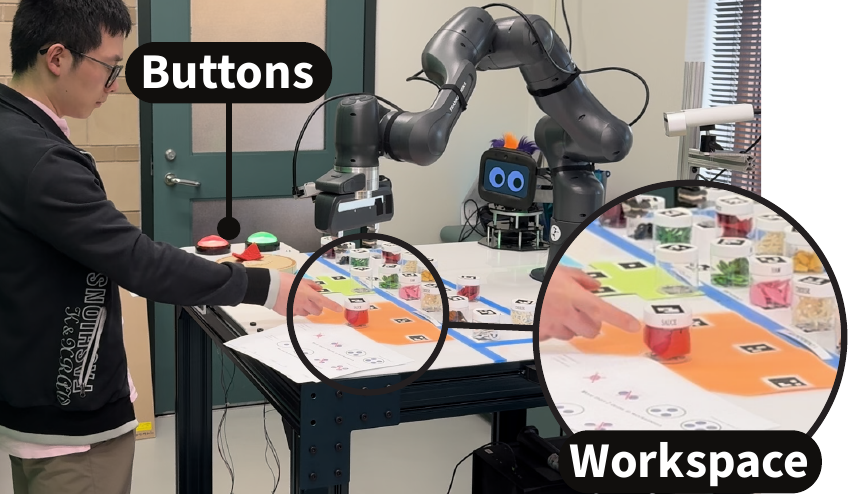}
  \caption{Collaboration setup.}
  \label{fig:online_lab_setup}
  \vspace{-1em}
\end{wrapfigure}

With approval from our IRB, we evaluated learning during physical collaborations with 20 participants (Fig.~\ref{fig:online_lab_setup}). We used a pizza-making setup validated in prior work~\cite{candon2026learning}, where a human collaborates with a Franka Panda arm to assemble toy pizzas in a lab. Different from~\cite{candon2026learning}, the robot updated its preference estimate during the interaction, rather than doing so offline. 
The study had a within-subjects design, where each participant experienced robot learning per Eq.~(\ref{eq:MLE-objective}) with the common \texttt{Fixed-Rule} implication as well as  
with \implied{}. 
The conditions were counterbalanced, and the robot's memory was cleared between them to avoid information leakage. For each condition, the team made two pizzas with different recipes---but the same were used for \texttt{Fixed-Rule} and \implied{}.  Participants decided on their own when to provide explicit feedback. Implicit feedback was received on every step based on the human's actions. 
See Appx.~\ref{sec:real_world_details} for more details. 

\begin{wraptable}{r}{0.45\textwidth}
  \centering
  \vspace{-1em}
  \caption{Online learning performance for the physical human-robot collaboration study.}
  \label{tab:online_lab_results}
  \small
  \begin{tabular}{lcc}
    \toprule
    Implication & L2 Error $\downarrow$ & Conflict Prop. $\downarrow$ \\
    \midrule
    IMPLIED  & $\mathbf{0.86 \pm 0.07}$ & $\mathbf{0.16 \pm 0.03}$ \\
    Fixed-Rule & $1.14 \pm 0.03$ & $0.25 \pm 0.03$ \\
    \bottomrule
  \end{tabular}
  \vspace{-1.0em}
\end{wraptable}

Table~\ref{tab:online_lab_results} reports L2 preference error and conflict proportion at the earliest shared task-completion step. Linear mixed-effects analyses showed that online learning with \implied{} achieved significantly lower L2 error ($p<0.01$) and conflict proportion ($p<0.05$) than learning with \texttt{Fixed-Rule}. \final{We also measured two objective interaction costs: the number of explicit feedback instances a participant provided, and the number of interaction steps taken to complete the task. Online learning with \implied{} required significantly fewer instances of explicit feedback than with \texttt{Fixed-Rule} ($34.10 \pm 0.88$ vs. $43.45 \pm 1.05$, $p < 0.0001$) and significantly fewer interaction steps to completion ($40.05 \pm 0.83$ vs. $45.35 \pm 1.15$, $p < 0.05$).}

%
%
%
Further, at the end of the study,  participants answered four comparison questions for the conditions. Two-tailed binomial tests showed that significantly more participants reported that \implied{} made the robot learn faster ($17/20$, $p<0.01$), required less teaching effort ($16/20$, $p<0.05$), led the robot to make fewer mistakes ($17/20$, $p<0.01$), and was the method they would prefer the robot to use in the future ($18/20$, $p<0.001$). 
These results suggest that users perceived \implied{} as more preferable than the \texttt{Fixed-Rule} approach. 

%% file: sections/simulation_table.tex
\definecolor{mplBlue}{rgb}{0.7176, 0.8431, 0.9490}
\definecolor{mplOrange}{rgb}{0.9843, 0.8902, 0.8353}

\begin{figure}[t!]
  \centering

  \begin{minipage}[t]{0.29\textwidth}
    \centering
    \vspace{0pt}
    \includegraphics[width=.92\linewidth]{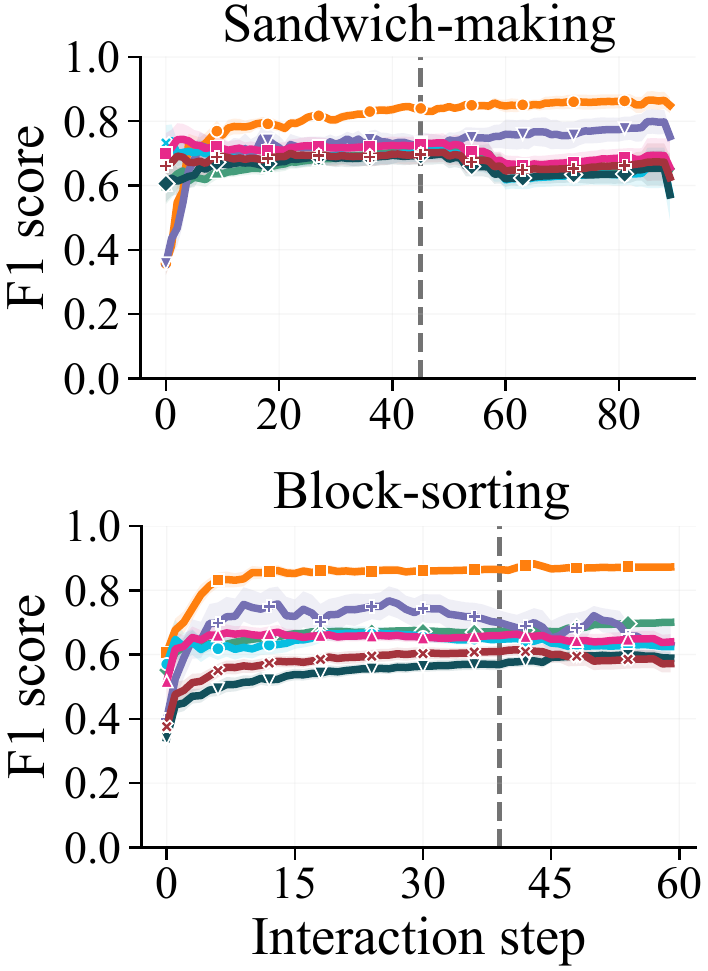}
    \captionof{figure}{
    F1 score of classifying unchosen actions over time.
    }
    \label{fig:implication_f1_curves}
  \end{minipage}
  \hfill
  \begin{minipage}[t]{0.7\textwidth}
    \centering
    \vspace{0pt}
    \captionof{table}{
    Downstream preference-learning performance using different implications.
    Values are mean $\pm$ standard error at the earliest task-completion timestep across participants. \protect\colorbox{mplBlue}{Human} is the oracle condition using ground-truth implication labels, and \protect\colorbox{mplOrange}{IMPLIED} highlights our proposed method.
    }
    \label{tab:simulation_downstream_results}

    \scriptsize
    \setlength{\tabcolsep}{3pt}
    \begin{adjustbox}{width=\linewidth}
    \begin{tabular}{lcccc}
      \toprule
      & \multicolumn{2}{c}{Sandwich-making}
      & \multicolumn{2}{c}{Block-sorting} \\
      \cmidrule(r){2-3} \cmidrule(r){4-5}
      Implication
      & L2 Error $\downarrow$
      & Conflict Prop. $\downarrow$
      & L2 Error $\downarrow$
      & Conflict Prop. $\downarrow$ \\
      \cmidrule(r){1-1} \cmidrule(r){2-2} \cmidrule(r){3-3} \cmidrule(r){4-4} \cmidrule(r){5-5}

      Human 
      & \cellcolor{mplBlue}$0.69 \pm 0.05$
      & \cellcolor{mplBlue}$0.11 \pm 0.01$
      & \cellcolor{mplBlue}$0.40 \pm 0.06$
      & \cellcolor{mplBlue}$0.10 \pm 0.03$ \\

      IMPLIED \examplesymbol[FF7F0E] 
      & \cellcolor{mplOrange}$0.71 \pm 0.06$
      & \cellcolor{mplOrange}$0.13 \pm 0.01$
      & \cellcolor{mplOrange}$0.42 \pm 0.07$
      & \cellcolor{mplOrange}$0.12 \pm 0.03$ \\

      Fixed-Rule \examplesymbol[449D7B]
      & $0.86 \pm 0.07$
      & $0.20 \pm 0.03$
      & $0.55 \pm 0.07$
      & $0.20 \pm 0.04$ \\

      No-Prior \examplesymbol[7570B3]
      & $0.93 \pm 0.06$
      & $0.20 \pm 0.03$
      & $0.63 \pm 0.09$
      & $0.19 \pm 0.05$ \\

      LLM \examplesymbol[07BEE1]
      & $0.87 \pm 0.08$
      & $0.19 \pm 0.04$
      & $0.59 \pm 0.07$
      & $0.20 \pm 0.04$ \\

      LLM-H \examplesymbol[E7298A]
      & $0.88 \pm 0.08$
      & $0.24 \pm 0.06$
      & $0.61 \pm 0.08$
      & $0.21 \pm 0.04$ \\

      LLM-B \examplesymbol[12505B]
      & $0.88 \pm 0.08$
      & $0.20 \pm 0.03$
      & $0.58 \pm 0.07$
      & $0.17 \pm 0.04$ \\

      LLM-HB \examplesymbol[A4333D]
      & $0.85 \pm 0.08$
      & $0.21 \pm 0.05$
      & $0.56 \pm 0.08$
      & $0.20 \pm 0.05$ \\

      \bottomrule
    \end{tabular}
    \end{adjustbox}
  \end{minipage}
\vspace{-1em}
\end{figure}

%% file: sections/6.related_work.tex
\section{Related Work} 
\label{sec:related_work}


There is a long history of work in preference learning. Early work (e.g.,  \cite{russell1998learning,ng2000algorithms}) focused on Inverse Reinforcement Learning (IRL),  based on the observation that in ``natural learning,'' a reward function is often not explicitly specified and instead must be learned. 
In HRI, learning human preferences typically involves a robot learning from human demonstrations of the task they want the robot to complete (e.g., \cite{argall2009survey,abbeel2004apprenticeship,billard2016learning}) or from evaluative feedback provided by a human observer (e.g., \cite{tamer2009, pmlr-v70-macglashan17a}). More related to our work, recent approaches have addressed how to combine different types of feedback: contributing a theoretical framework describing how different types of feedback can be mapped to a choice from a set of choices \cite{jeon2020reward}; integrating demonstrations together with preference queries \cite{biyik2022learning}; combining demonstrations, corrections, preferences, and binary feedback \cite{fitzgerald2022inquire}; reasoning about evaluative binary feedback alongside implicit feedback from human task actions \cite{candon2026learning}; or combining instructions and demonstrations \cite{zhu2026interpret}. 

\final{Beyond combining feedback types, several methods derive labels beyond direct human input. For instance, \citet{zhan2021human} train an adversarial network to take over the human's role in labeling trajectory pairs, SURF \cite{park2022surf} assigns confidence-based pseudo-labels to unlabeled segment pairs, and in text-to-image alignment, Semi-DPO \cite{liu2026learning} self-trains on a consensus-filtered subset to relabel conflicting preference pairs. Furthermore, D-REX \cite{brown2020better} and SSRR \cite{chen2021learning} instead synthesize preference rankings by injecting noise into a cloned policy. Closer to our setting, \citet{loftin2016learning} model the human's feedback strategy to interpret what the absence of feedback implies.}
Our work adds to this literature by showing that the implications of human feedback can be learned during preference learning, \final{revising the labels of feasible unchosen actions from accumulated interaction evidence.}




%% file: sections/7.limitations.tex
\section{Limitations and Future Work}
\label{sec:limitations}

We proposed \implied{}, a method that learns and revises implication labels from accumulated direct evidence. 
Our results suggest that \implied{} improves implication prediction, preference estimation, and robot behavior in comparison to using a common fixed-rule implication or using large models to derive evidence for preference learning. 

However, our work is not without limitations, which point to interesting avenues for future work.
First, we assumed that the relevant preference-feature dimensions are known, so learning focuses on estimating their weights ($\mathbf{w}$ in Eq.~(\ref{eq:MLE-objective})). This follows a long tradition of using linear reward functions in HRI, often due to sample efficiency concerns, but limits the method to tasks where designers can specify what humans may care about. Future work could investigate ways to jointly infer both the preference representation~\cite{dennler2025hri} and the weights. 

Second, we focused on learning from two types of feedback modalities: binary explicit feedback, and implicit feedback provided by the human's actions. In the future, it would be interesting to explore using more modalities, like demonstrations~\cite{fitzgerald2022inquire} or  language~\cite{jeon2020reward} for which unified learning formalisms exist. 

Third, our experiments used structured tasks with known goal rewards and relatively stable preferences. Extending implication modeling beyond these assumptions to more open-ended collaboration settings, where human preferences may shift over time \cite{ghose2025ve, ma2024goal, ghose2026open, jain2019probabilistic}, will require richer action representations and robustness to noisy or inconsistent feedback.
\final{Relatedly, IMPLIED assumes a non-adversarial collaborator. A user who gives systematically misleading feedback could induce an overconfident and wrong GP that dominates inference. Detecting such inconsistency and falling back to the prior is an open problem.}

Finally, we assumed \final{an enumerable set of} discrete high-level actions, which is also common in prior HRI work \cite{ghose2025ve, candon2026learning,  ghose2026open, brawerhri23-overlay}.
\final{Future work could study implication modeling in continuous action spaces and with vision-language-action (VLA) policies, which would first require deciding the unit of feedback. Specifically, Eq. (\ref{eq:MLE-objective}) reasons about one high-level action at a time, so implications would instead have to be defined over trajectories. INQUIRE \cite{fitzgerald2022inquire} does so, but assumes fixed, pre-segmented trajectories, whereas during a collaboration trajectory boundaries are not obvious. Non-collaborative VLA settings, where trajectory lengths can be pre-determined, may thus be a more tractable entry point for this future work direction.}

%% file: appendix.tex
\section{Details of the User Study in MuJoCo Simulation} 
\label{sec:mujuco_design}

\begin{figure}[h]
    \centering
    \includegraphics[width=0.95\linewidth]{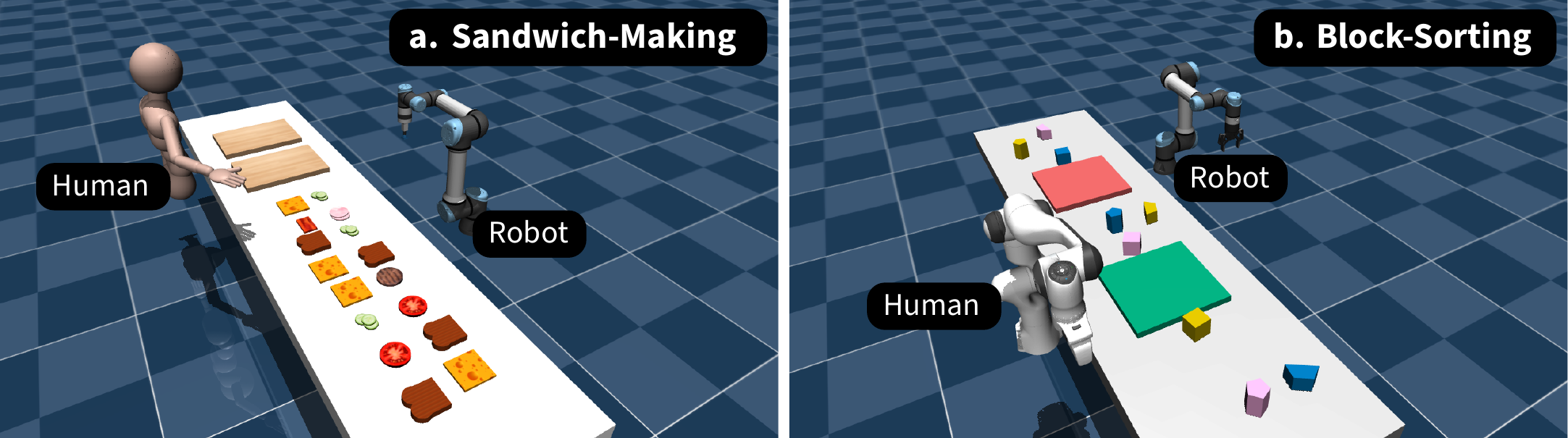}
    \caption{MuJoCo tasks. For both tasks, a human controls an agent (shown on the left side of both images). The other agent (the robot) is autonomous. It estimates preferences and acts according to them during our user study. These MuJoCo tasks were implemented by adapting and extending RoCo simulation components~\cite{mandi2024roco}.}
    \label{fig:mujuco}
\end{figure}

In this section, we describe in more detail the two simulation environments used for our initial user study (Sec.~\ref{sec:study}), which gathered evidence that the fixed-rule implication does not always match human judgments for unchosen actions. Fig.~\ref{fig:mujuco} shows a screenshot of each of the simulated tasks. For both, the robot and the human have fixed roles. 

\subsection{Study Protocol}

We collected human feedback and implication labels in two MuJoCo human-robot collaboration
environments: sandwich-making (Fig.~\ref{fig:mujuco}(a)) and block-sorting
(Fig.~\ref{fig:mujuco}(b)). The study was approved by our local Institutional Review Board.
In both environments, the human and robot acted in a shared task state. Before each task,
participants selected a true preference from the available options and were instructed to use this
preference consistently when choosing their own actions and evaluating the robot. The robot knew the
task goal reward but did not know the weights corresponding to the participant's true preference.
Further details on the available task initializations and preference options are provided in
Appx.~\ref{sec:sandwich_details} and \ref{sec:sort_details}.

\begin{figure}[t]
    \centering
    \includegraphics[width=0.85\linewidth]{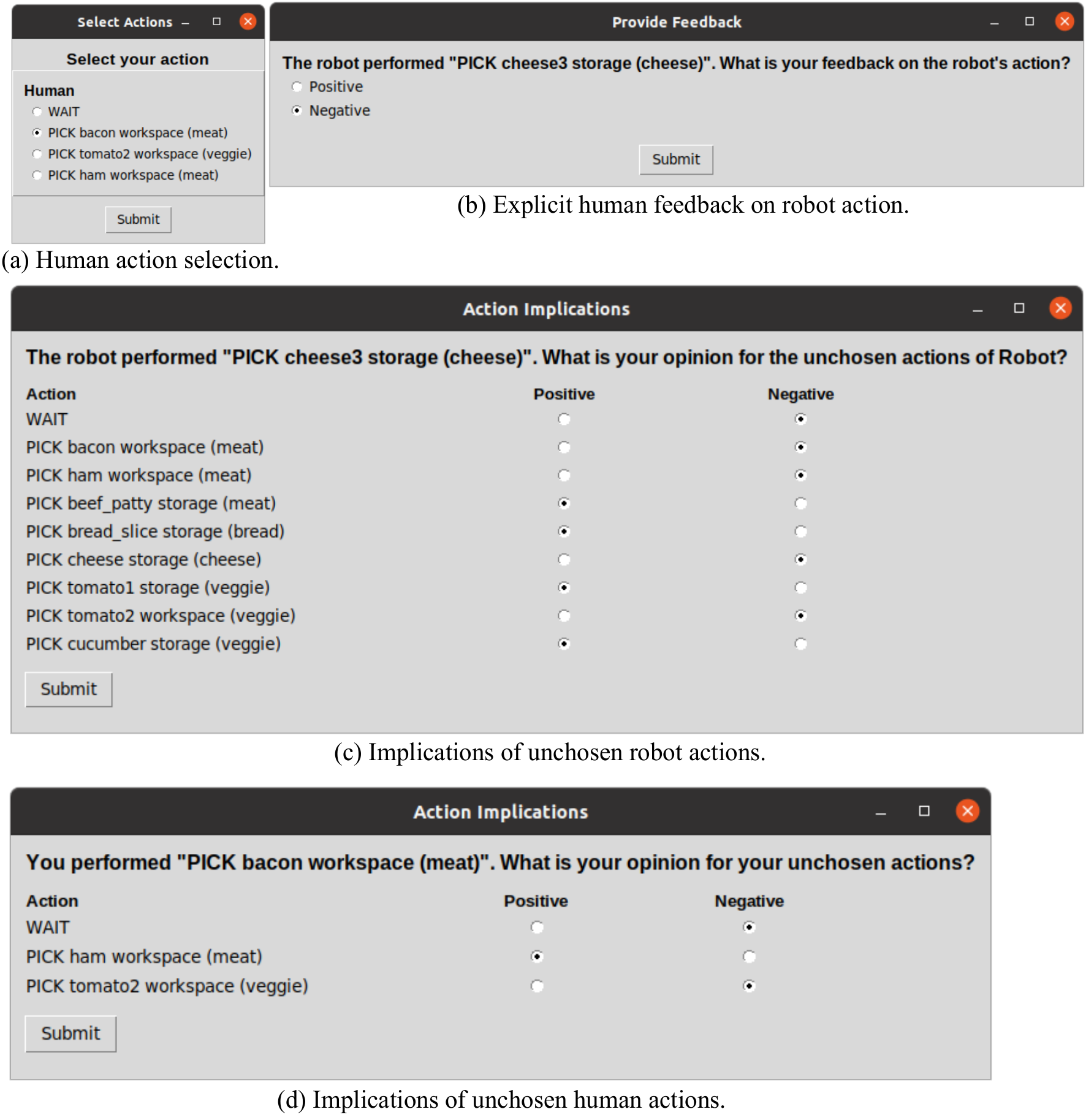}
    \caption{
    Graphical user interface used in the MuJoCo simulation study. At each timestep, participants (a) selected their own action, (b) provided explicit feedback on the robot's action, (c) labeled implications for the robot's unchosen actions, and (d) labeled implications for their own unchosen actions.
    }
    \label{fig:simulation-labeling-gui}
\end{figure}

At each timestep, the participant first selected their own action from the available human actions
through the graphical user interface shown in Fig.~\ref{fig:simulation-labeling-gui}(a).
The robot also selected one action from its available action set according to the current policy
$\pi_R(a_R|\mathbf{s}) = P(a_R|\mathbf{s}) \propto \exp(\beta_R R_R(\mathbf{s}, a_R))$,
where the combined robot reward $R_R$ is determined by the latest preference estimate $\mathbf{w}$,
and $\beta_R$ controls how rational the robot's actions are. The joint human-robot actions were then
rendered in the simulation.

After the transition in simulation, the participant used the interface to give explicit binary feedback on the robot's action taken
(Fig.~\ref{fig:simulation-labeling-gui}(b)). The participant was instructed to judge whether the
robot's action helped complete the task while respecting how they preferred the task to be performed.
The interface then asked the participant to label the implications of the robot's unchosen actions as
positive or negative (Fig.~\ref{fig:simulation-labeling-gui}(c)). Finally, the participant labeled the
implications of their own unchosen actions (Fig.~\ref{fig:simulation-labeling-gui}(d)).

Half of the participants were assigned to a Fixed-Rule condition, in which the robot estimated
preferences using the standard implicature rule described in Sec.~\ref{sec:preliminaries} after each
timestep. The other half were assigned to the Human Implication condition, in which the robot used the
same preference-learning method but with the participant-provided implication labels. Thus, the
condition determined which feedback sets were used by the robot during the interaction. After each
feedback step, the robot updated its preference estimate using gradient ascent on
Eq.~(\ref{eq:MLE-objective}) and selected later actions based on its latest preference estimate.

Each participant completed both tasks in the simulation, and the order of the two tasks was
counterbalanced between participants. For each task, the participant completed three episodes. The three
episodes used different initializations, randomly sampled from the available task initialization options.
Each episode continued until the task was completed. Across episodes, the participant interacted with the
robot, observed the rendered simulation state, and provided the action selections (implicit feedback), explicit feedback, and implication labels described above.

A total of 18 participants completed the study. Each study session typically lasted 45 min, and each
participant was compensated \$15. Participants were recruited through flyers, online postings, and word
of mouth, and were required to be at least 18 years old, fluent in English, and have normal or
corrected-to-normal hearing and vision.


\subsection{Sandwich-Making Task}
\label{sec:sandwich_details}

\textbf{Domain basics.}
Fig.~\ref{fig:mujuco}(a) shows the sandwich-making task, in which a robot and a human assemble a target sandwich from 16 ingredients:
\begin{itemize}[wide]
    \item Bread (4): \texttt{bread\_slice1}, \texttt{bread\_slice2}, \texttt{bread\_slice3}, \texttt{bread\_slice4}
    \item Cheese (4): \texttt{cheese1}, \texttt{cheese2}, \texttt{cheese3}, \texttt{cheese4}
    \item Meat (3): \texttt{bacon}, \texttt{ham}, \texttt{beef\_patty}
    \item Veggie (5): \texttt{tomato1}, \texttt{tomato2}, \texttt{cucumber1}, \texttt{cucumber2}, \texttt{cucumber3}
\end{itemize}
The task locations are \texttt{storage}, \texttt{workspace}, and \texttt{sandwich}, with additional
robot- and human-hand states. Initially, all ingredients are in storage, and the workspace,
sandwich, and both hands are empty. The workspace can contain at most four ingredients. The robot
can move ingredients between storage and workspace, while the human can move ingredients between
workspace and sandwich. Each episode specifies a target recipe, and completion is evaluated using
required counts over goal ingredient classes
(bread, cheese, bacon, ham, beef, tomato, and cucumber).

\textbf{State space.}
The task state is represented by the allocation of ingredients across the five relevant locations:
\texttt{storage}, \texttt{workspace}, \texttt{sandwich}, the robot's hand, and the human's hand.
Together with the target recipe, this allocation determines the relevant task progress: which
required goal-class counts have already been satisfied on the sandwich, which ingredients are
available to each agent, whether either agent is holding an ingredient, and whether the workspace
capacity constraint is active.

\textbf{Action space and transitions.}
The robot's action space has a total of 19 actions: 16 \texttt{PICK(ingredient)} actions, \texttt{PUT(ingredient, storage)},
\texttt{PUT(ingredient, workspace)}, and \texttt{WAIT}. If the robot is not holding an ingredient,
it may pick an ingredient from storage or workspace; if it is holding an ingredient, it may put it
in storage or, when the workspace is not full, put it in the workspace. The human's action space also has a total of 19
actions: 16 \texttt{PICK(ingredient)} actions, \texttt{PUT(ingredient, workspace)},
\texttt{PUT(ingredient, sandwich)}, and \texttt{WAIT}. The human may pick only from the workspace;
if holding an ingredient, the human may put it on the sandwich or return it to the workspace when
space is available. \texttt{WAIT} is always available for both agents. At each step, the robot and
human choose simultaneous actions; a pick action transfers an ingredient to the acting agent's hand,
and a put action transfers a held ingredient to the chosen target location.

\textbf{Preference space.}
Human preferences are represented by a 10-dimensional weight vector,
\[
\begin{aligned}
\mathbf{w} = [&\mathtt{cheese\_before\_bread},\ \mathtt{cheese\_before\_meat},\\
     &\mathtt{cheese\_before\_veggie},\ \mathtt{bread\_before\_meat},\\
     &\mathtt{bread\_before\_veggie},\ \mathtt{meat\_before\_veggie},\\
     &\mathtt{workspace\_size\_le0},\ \mathtt{workspace\_size\_le1},\\
     &\mathtt{workspace\_size\_le2},\ \mathtt{workspace\_size\_le3}]
\end{aligned}
\]
The first six dimensions encode pairwise ordering preferences over the four broad ingredient types
cheese, bread, meat, and veggie. For example, a positive value of
\texttt{cheese\_before\_bread} means that all target cheese should be added before any target bread;
a negative value represents the reverse preference, and a value near zero represents no preference
between the two types. The last four dimensions encode preferences over workspace occupancy:
dimension \(k\) is positive when the post-action workspace size is at most \(k\) and negative
otherwise. Weights are normalized to unit Euclidean norm, and sampled preferences are constrained to
have acyclic ordering preferences and consistent workspace-size preferences.

\textbf{True preference options and environment initializations.}
Before the sandwich-making task, participants selected one of six preference choices and used it
consistently across all sandwich-making trials. The true preference options were:
\begin{itemize}[wide]
    \item \textit{Preference A:} Bread before cheese before veggie before meat.
    \item \textit{Preference B:} Bread before meat before veggie.
    \item \textit{Preference C:} Cheese before meat before bread and veggie.
    \item \textit{Preference D:} Veggie before meat.
    \item \textit{Preference E:} Prefer more than two ingredients in the workspace.
    \item \textit{Preference F:} Prefer at most two ingredients in the workspace.
\end{itemize}
Each environment initialization specified one target recipe. The recipe options were:
\begin{itemize}[wide]
    \item \textit{Initial state 1:} 2 bread, 1 cheese, 1 ham, 1 bacon, 1 tomato, 1 cucumber.
    \item \textit{Initial state 2:} 2 bread, 2 cheese, 1 beef, 1 tomato, 1 cucumber.
    \item \textit{Initial state 3:} 2 bread, 2 cheese, 1 ham, 1 bacon, 1 beef.
    \item \textit{Initial state 4:} 2 bread, 2 cheese, 1 ham, 1 bacon, 1 tomato.
    \item \textit{Initial state 5:} 2 bread, 1 cheese, 2 tomato, 1 cucumber.
    \item \textit{Initial state 6:} 1 bread, 1 cheese, 1 ham, 1 tomato.
\end{itemize}

\textbf{Reward function.}
The shared reward is decomposed into goal and preference terms,
\[
R(s,a_R,a_H) = R_{\mathrm{goal}}(s,a_R,a_H) +
\gamma R_{\mathrm{pref}}(s,a_R,a_H),
\]
where \(\gamma\) controls the relative importance of the preference reward and goal reward.

For the goal reward, each agent receives a step cost of \(-0.3\), and completing the sandwich gives a \(+10\) bonus. The robot is rewarded for moving still-needed target ingredients
toward the human: picking a needed target from storage gives \(+2\), and placing a needed target
from storage into the workspace gives \(+3\). To incentivize efficient collaboration, regressive or redundant robot actions are penalized:
picking a needed ingredient that is already in the workspace gives \(-2\) (or \(-0.1\) if the
workspace is full), returning a needed target to storage gives \(-1\), cycling a workspace item
back to the workspace gives \(-4\), handling already-satisfied target ingredients gives penalties
of \(-3\) or \(-5\), and handling non-target ingredients gives penalties of \(-2\) or \(-10\)
depending on whether the action is moving them away from or toward the main task area. Waiting while
holding an ingredient, or while a needed target remains unavailable to the human, gives \(-5\).
The human receives \(+2\) for picking a still-needed target from the workspace and \(+3\) for
placing it on the sandwich. Returning a needed target to the workspace gives \(-1\), acting on an
already-satisfied target gives \(-3\) or \(-5\), acting on a non-target ingredient gives \(-2\) or
\(-10\), and attempting to place without holding an ingredient gives \(-1\) for the workspace or
\(-2\) for the sandwich.

The preference reward is linear in interpretable feature vectors:
\[
R_{\mathrm{pref}}(s,a_R,a_H)
= \mathbf{w}^\top \phi_R(s,a_R) + \mathbf{w}^\top \phi_H(s,a_H)
= \mathbf{w}^\top(\phi_R(s,a_R)+\phi_H(s,a_H)).
\]
The feature vectors $\phi_R$ and $\phi_H$ have the same 10 dimensions as \(\mathbf{w}\). Ordering features are $+1$ when an
action supports the preferred ingredient ordering, $-1$ when it violates that ordering, and $0$
when the action is irrelevant. Workspace features evaluate the workspace size induced by the
action, with features set to zero for size thresholds that are no longer relevant once too few
target ingredients remain.

\subsection{Block-Sorting Task}
\label{sec:sort_details}

\textbf{Domain basics.}
Fig.~\ref{fig:mujuco}(b) shows the block-sorting task, in which a robot and a human sort nine blocks into two target bins. The
blocks are the Cartesian product of three colors and three shapes:
\begin{itemize}[wide]
    \item Blue: \texttt{blue\_square}, \texttt{blue\_polygon}, \texttt{blue\_trapezoid}
    \item Pink: \texttt{pink\_square}, \texttt{pink\_polygon}, \texttt{pink\_trapezoid}
    \item Yellow: \texttt{yellow\_square}, \texttt{yellow\_polygon}, \texttt{yellow\_trapezoid}
\end{itemize}
From left to right, the five table regions are \texttt{left}, \texttt{red}, \texttt{mid},
\texttt{green}, and \texttt{right}. The \texttt{red} and \texttt{green} regions are the target bins,
while \texttt{left}, \texttt{mid}, and \texttt{right} are staging regions. Initially, each block is
placed in one of the staging regions. The robot can reach \texttt{left}, \texttt{red},
and \texttt{mid}; the human can reach \texttt{mid}, \texttt{green}, and \texttt{right}. The table
regions do not impose capacity limits. The task is complete when every block is in either the red or
green bin and neither agent is holding a block.

\textbf{State space.}
The task state is represented by the allocation of blocks across the five table regions
\texttt{left}, \texttt{red}, \texttt{mid}, \texttt{green}, and \texttt{right}, plus the two hand
locations of the robot and human. This allocation determines whether each block has already reached
a target bin, which blocks are reachable by each agent, whether either agent is holding a block, and
whether the task is complete.

\textbf{Action space and transitions.}
The action space of each agent has 15 possible actions: nine \texttt{PICK(block)} actions, five
\texttt{PLACE(block, location)} actions for \texttt{left}, \texttt{red}, \texttt{mid},
\texttt{green}, and \texttt{right}, and \texttt{WAIT}. An agent can take only actions involving
regions it can reach. Thus,
when the robot is empty-handed, it may pick a block from \texttt{left}, \texttt{red}, or
\texttt{mid}; when holding a block, it may place it in one of those same regions. The human follows
the analogous rule over \texttt{mid}, \texttt{green}, and \texttt{right}. The middle region is
reachable by both agents and enables handoff between the robot side and the human side. At each
step, simultaneous actions move picked blocks into the acting agent's hand or move held blocks to
the selected reachable region. \texttt{WAIT} is always available.

\textbf{Preference space.}
Human preferences specify which bin each block type should ultimately be sorted into. The preference
weight vector has nine dimensions,
\[
\begin{aligned}
\mathbf{w} = [&\mathtt{blue\_square},\ \mathtt{blue\_trapezoid},\ \mathtt{blue\_polygon},\\
     &\mathtt{pink\_square},\ \mathtt{pink\_trapezoid},\ \mathtt{pink\_polygon},\\
     &\mathtt{yellow\_square},\ \mathtt{yellow\_trapezoid},\ \mathtt{yellow\_polygon}]
\end{aligned}
\]
A positive weight means that the corresponding block type is preferred in the red bin, a negative
weight means that it is preferred in the green bin, and a value near zero means either bin is
acceptable. Weights are normalized to unit Euclidean norm; sampled preferences are constrained so
that not all block-bin dimensions have the same sign.

\textbf{True preference options and environment initializations.}
Before the block-sorting task, participants selected one of six preference choices and used it
consistently across all block-sorting trials. The true preference options were:
\begin{itemize}[wide]
    \item \textit{Preference A:} Blue and yellow blocks to red; pink blocks to green.
    \item \textit{Preference B:} Blue and yellow blocks to green; pink blocks to red.
    \item \textit{Preference C:} Squares and polygons to red; trapezoids to green.
    \item \textit{Preference D:} Squares and polygons to green; trapezoids to red.
    \item \textit{Preference E:} Blue square to red; pink polygon and yellow trapezoid to green; all other blocks either.
    \item \textit{Preference F:} Blue polygon to green; pink trapezoid and yellow square to red; all other blocks either.
\end{itemize}
Each environment initialization specified one initial table layout, i.e., the blocks initially placed
in the left, middle, and right staging regions:
\begin{itemize}[wide]
    \item \textit{Initial state 1:} Left: blue square, pink trapezoid, yellow polygon; Mid: pink square, blue polygon, yellow trapezoid; Right: yellow square, blue trapezoid, pink polygon.
    \item \textit{Initial state 2:} Left: blue square, pink square; Mid: blue polygon, pink polygon, yellow polygon, yellow square, pink trapezoid; Right: blue trapezoid, yellow trapezoid.
    \item \textit{Initial state 3:} Left: blue square, pink square, yellow square, blue polygon; Mid: pink polygon; Right: yellow polygon, pink trapezoid, blue trapezoid, yellow trapezoid.
    \item \textit{Initial state 4:} Left: blue square, pink trapezoid; Mid: yellow square, pink square, blue polygon; Right: yellow polygon, blue trapezoid, pink polygon, yellow trapezoid.
    \item \textit{Initial state 5:} Left: blue square, pink square, yellow square; Mid: blue polygon, pink polygon, yellow polygon; Right: blue trapezoid, pink trapezoid, yellow trapezoid.
    \item \textit{Initial state 6:} Left: blue square, blue polygon, blue trapezoid; Mid: pink square, pink polygon, pink trapezoid; Right: yellow square, yellow polygon, yellow trapezoid.
\end{itemize}

\textbf{Reward function.}
As in the sandwich task, the shared reward decomposes into goal and preference terms:
\[
R(s,a_R,a_H) = R_{\mathrm{goal}}(s,a_R,a_H) +
\gamma R_{\mathrm{pref}}(s,a_R,a_H),
\text{ with }
R_{\mathrm{pref}} = \mathbf{w}^\top(\phi_R(s,a_R)+\phi_H(s,a_H))
\]
The goal reward uses the same cases for the robot and the human. Each agent
receives a step cost of \(-0.3\), and completing the sort gives a \(+10\) bonus. Goal progress is
encouraged by rewarding picks from staging regions (\(+0.8\)) and placements into either target bin
(\(+1.5\)). Placing a block into the shared middle region receives a smaller reward (\(+0.3\))
because it can support handoff between agents. Actions that undo or delay progress are penalized:
picking from a target bin gives \(-0.6\), placing into a side staging region gives \(-0.5\),
attempting to place without holding a block gives \(-1.0\), and waiting while holding a block gives
\(-1.0\). Waiting without holding a block has no additional goal reward.

The feature vector \(\phi_{\mathrm{agent}}\in\{-1,0,1\}^9\), where $\mathrm{agent}$ is either the human or robot, marks the block type affected
by the action. Placing a block in the red bin, or removing it from the green bin, gives a \(+1\)
feature for that block; placing it in the green bin, or removing it from the red bin, gives a \(-1\)
feature. Actions involving the shared middle region are signed according to which side can next
make progress: robot actions that take a block from \texttt{mid} or human actions that place a block
in \texttt{mid} contribute \(+1\), while human actions that take a block from \texttt{mid} or robot
actions that place a block in \texttt{mid} contribute \(-1\). Otherwise, the feature will be $0$.

\subsection{Additional Preference-Learning Parameters}

Table~\ref{tab:sim-pref-learning-params} summarizes the shared preference-learning parameters used
for our study. Following the learning setup used by \citet{candon2026learning}, we used
one shared set of preference-learning parameters across different collaboration tasks and implication sources. This choice keeps the preference-learning procedure general and makes the comparisons focus on the source of implication labels. 

\begin{table}[h]
\centering
\small
\caption{Preference-learning parameters for user study in MuJoCo simulation.}
\vspace{1em}
\label{tab:sim-pref-learning-params}
\begin{tabular}{lll}
\hline
Parameter & Value & Description \\
\hline
learning\_rate & \(1\times 10^{-4}\) & Step size for preference-weight updates \\
weight\_samples & 10 & Number of samples representing the belief over weight \\
max\_iteration & 1000 & Maximum iterations for each preference update \\
convergence\_threshold & \(1\times 10^{-3}\) & Threshold for preference update convergence \\
\(\beta_R\) & 1.0 & Robot action rationality coefficient \\
\(\beta_{\mathrm{exp}}\) & 1.0 & Explicit-feedback rationality coefficient \\
\(\beta_{\mathrm{imp}}\) & 1.0 & Implicit-feedback rationality coefficient \\
\(\gamma\) & 5.0 & Relative weight on preference reward vs. goal reward \\
\hline
\end{tabular}
\vspace{-1em}
\end{table}

\subsection{Demographic Information}

Participants completed a demographic survey at the beginning of the study. The statistics for the
18 participants are summarized below:
\begin{itemize}[leftmargin=1.5em]
    \item Gender: 12 participants were male and 6 were female.
    \item Age: The participants' mean age was 24 years old
    ($\mathrm{SD}=7.2$), with ages ranging from 18 to 50.
    \item Familiarity with robotics: Participants rated their familiarity with robotics on a
    7-point Likert scale, where 1 indicated ``not familiar at all'' and 7 indicated ``very familiar.''
    The mean rating was 4.78 ($\mathrm{SD}=1.40$).
    \item Frequency of interacting with robots: 6 participants reported interacting with
    robots less than once a month, 6 reported once a month, 1 reported once a week, 3 reported
    2--3 times a week, 1 reported 4--6 times a week, and 1 reported daily.
\end{itemize}

\section{Details of the GP Model for IMPLIED}
\label{sec:model_design}

\subsection{Derivation of $\lambda$ in Eq.~(\ref{eq:IMPLIED-prob})}

We use the variance $v(\mathbf{o})$ to derive $\lambda(\mathbf{o})$ in Eq.~(\ref{eq:IMPLIED-prob}). The function $\lambda(\mathbf{o})$ returns an observation-specific weight that balances the contributions of the fixed-rule prior and the learned GP. Since implication labels are binary, the learned GP outputs a probability for the positive implication label, which can be interpreted as the parameter of a Bernoulli distribution over accepted/rejected labels. The most uncertain case occurs when this probability is $p=0.5$, where the two labels are equally likely. The corresponding Bernoulli variance is therefore maximized at
$v_{\max}=p(1-p)=0.25$. We use this maximum variance to normalize the GP uncertainty and let:
\begin{equation}
    \lambda(\mathbf{o})=1-v(\mathbf{o})/v_{\max}=1-4v(\mathbf{o}) \in[0,1]
\end{equation}
This choice gives a bounded weight in $[0,1]$ and makes the confidence in the learned GP higher when the model is more certain, i.e., when the variance is lower. This confidence also reflects distance from the decision boundary, which we set to $0.5$ probability in our experiments. If the probability samples behave like Bernoulli draws with mean $\mu(\mathbf{o})$, then:
\begin{align}
    v(\mathbf{o}) &\approx \mu(\mathbf{o})(1-\mu(\mathbf{o}))\\
    \lambda(\mathbf{o}) 
    &= 1 - 4v(\mathbf{o}) \notag\\
    &\approx 1 - 4\mu(\mathbf{o})(1-\mu(\mathbf{o})) \notag\\
    &= 1 - 4\mu(\mathbf{o}) + 4\mu(\mathbf{o})^2 \notag\\
    &= 4(\mu(\mathbf{o})-0.5)^2 .
\end{align}

Thus, $\lambda(\mathbf{o})$ is small when $\mu(\mathbf{o})$ is near $0.5$ and large when $\mu(\mathbf{o})$ is near $0$ or $1$.

\subsection{GP Hyperparameter Search}
\label{app:gp-hyper}

For the GP classifier in Sec.~\ref{method}, the covariance matrix $K$ is computed from Z-score normalized observation features. Let $\tilde{z}(\mathbf{o})$ denote the standardized version of $z(\mathbf{o})$. For two observations $\mathbf{o}_i$ and $\mathbf{o}_j$, we use
$K_{ij}=k_\theta(\tilde{z}(\mathbf{o}_i),\tilde{z}(\mathbf{o}_j))$, with
$k_\theta(\tilde{z},\tilde{z}')=\sigma_{\ell}^{2}\tilde{z}^{\top}\tilde{z}'+\sigma_b^2$.
Here, the parameters $\theta=(\sigma_{\ell}^{2},\sigma_b^2)$, where $\sigma_{\ell}^{2}$ controls the scale of the linear dependence on the feature inputs, and $\sigma_b^2$ is the additive bias variance, allowing the latent acceptance function to include a global offset. Thus, the latent function values used by $h_\theta=\sigma\circ f$ follow the GP prior induced by this kernel, $f\sim\mathcal{N}(\mathbf{0},K)$.

For experiments with human-in-the-loop simulation data, we used leave-one-participant-out cross-validation, as explained in Appx.~\ref{sec:implicit_explicit_only}. For the GP, the train/val split served to select the GP kernel hyperparameters $\theta=(\sigma_{\ell}^{2},\sigma_b^2)$ separately within each leave-one-participant-out fold. 
We selected $\theta$ from a finite grid formed by the Cartesian product of
$\sigma_{\ell}^{2}\in\{0.25,0.5,1.0,2.0,4.0\}$ and
$\sigma_b^2\in\{0.05,0.1,0.2,0.5,1.0\}$.
Specifically, in each fold, one participant was held out for testing, and the remaining participants were split into $80\%$ training and $20\%$ validation participants. For each candidate $\theta$, we fit the GP classifier using cumulative direct evidence from the training participants and evaluate implication-prediction performance on the validation participants. The parameter setting with the best validation performance is then fixed and used for making predictions on the held-out test participant. Implication labels from the held-out participant are never used for hyperparameter selection.

For experiments with real-world human-robot collaboration, we use a single parameter setting selected from the search performed for the simulation data. Specifically, we chose the parameters most frequently selected as optimal across the simulation folds, yielding $\sigma_{\ell}^{2}=1.0$ and $\sigma_b^2=0.2$. These values were used for all participants in the real-world human-robot collaboration (Sec.~\ref{sec:eval_physical_robot}). 

\subsection{Sensitivity to the Threshold $\tau$}
\label{app:tau}

We set $\tau = 0.5$ in Sec.~\ref{method} because it is the standard decision boundary for
binary classification. To verify that the choice is reasonable, we conducted a sensitivity analysis by re-running the evaluation of Sec.~\ref{sec:eval_mujoco} on data from Sec.~\ref{sec:study} with $\tau \in \{0.25, 0.5, 0.75\}$. Table~\ref{tab:tau}
reports the results. In sandwich-making, $\tau = 0.5$ significantly outperformed both alternatives on
all three metrics ($p < 0.05$). In block-sorting, $\tau = 0.5$ significantly
outperformed $\tau = 0.25$ and $\tau = 0.75$ on F1 ($p < 0.01$), and
outperformed $\tau = 0.25$ on L2 error ($p < 0.05$). We further swept $\tau$ in
increments of $0.05$ for both tasks and found no significant difference from
$\tau = 0.5$ for $\tau \in [0.40, 0.60]$.

\begin{table}[h]
\centering
\small
\caption{Sensitivity analysis for the threshold $\tau$.}
\vspace{0.5em}
\label{tab:tau}
\begin{tabular}{llccc}
\toprule
Task & $\tau$ & F1 $\uparrow$ & L2 Error $\downarrow$ & Conflict Prop. $\downarrow$ \\
\midrule
\multirow{3}{*}{Sandwich-making}
 & 0.25 & $0.77 \pm 0.03$ & $0.84 \pm 0.06$ & $0.18 \pm 0.04$ \\
 & 0.50 & $\mathbf{0.84 \pm 0.01}$ & $\mathbf{0.71 \pm 0.06}$ & $\mathbf{0.13 \pm 0.01}$ \\
 & 0.75 & $0.76 \pm 0.02$ & $0.80 \pm 0.07$ & $0.17 \pm 0.03$ \\
\midrule
\multirow{3}{*}{Block-sorting}
 & 0.25 & $0.72 \pm 0.04$ & $0.56 \pm 0.08$ & $0.16 \pm 0.05$ \\
 & 0.50 & $\mathbf{0.87 \pm 0.01}$ & $\mathbf{0.42 \pm 0.07}$ & $\mathbf{0.12 \pm 0.03}$ \\
 & 0.75 & $0.74 \pm 0.01$ & $0.49 \pm 0.07$ & $0.16 \pm 0.04$ \\
\bottomrule
\end{tabular}
\end{table}

\subsection{Comparison with a Simpler Learned Classifier}
\label{app:knn}

We tested whether a simpler learner for IMPLIED could lead to performance close to that of the GP. We swapped $P_{\text{Learned}}$ for a K-Nearest Neighbors (KNN)
classifier, keeping the same set of features $z(\mathbf{o})$, the Z-score normalization, and the mixture in Eq.~(\ref{eq:IMPLIED-prob}). We selected $k \in \{3, 5, 7, 9\}$ using the leave-one-participant-out cross validation mentioned in Appx.~\ref{app:gp-hyper}. We set $\mu(\mathbf{o})$ to the fraction of neighbors voting for the accepted label and used the corresponding Bernoulli variance $v(\mathbf{o}) = \mu(\mathbf{o})(1-\mu(\mathbf{o}))$ to compute $\lambda(\mathbf{o})= 1-4v(\mathbf{o})$.

On the recorded interaction data from Sec.~\ref{sec:study}, GP outperformed KNN at the earliest shared
task-completion step. In sandwich-making, GP led to significantly higher F1 ($0.84 \pm 0.01$ vs.
$0.75 \pm 0.03$, $p < 0.01$), lower L2 error ($0.71 \pm 0.06$ vs.
$0.89 \pm 0.07$, $p < 0.001$), and lower conflict proportion ($0.13 \pm 0.01$ vs.
$0.19 \pm 0.03$, $p < 0.05$) than KNN. In block-sorting, GP also led to significantly higher F1
($0.87 \pm 0.01$ vs. $0.71 \pm 0.03$, $p < 0.001$), lower L2 error ($0.42 \pm 0.07$ vs.
$0.60 \pm 0.08$, $p < 0.001$), and lower conflict proportion ($0.12 \pm 0.03$ vs. $0.20 \pm 0.05$, $p < 0.05$) than KNN.

\section{Learning with Human-in-the-Loop Simulation Data} 
\label{sec:implicit_explicit_only}

\input{sections/simulation_table_explicit}
\input{sections/simulation_table_implicit}

In Sec.~\ref{sec:eval_mujoco}, we evaluate implication prediction and preference learning using both implicit feedback from human actions and explicit feedback on robot actions. For these experiments, we use leave-one-participant-out cross-validation to measure generalization performance. In each fold, one participant was held out for testing, and the remaining participants were split into $80\%$ training and $20\%$ validation participants. The training and validation data served to fit hyperparameters, e.g., for the GP in IMPLIED and the No-Prior baseline, as mentioned in Appx.~\ref{sec:model_design}. 

For the Large Language Model baselines, we used Gemini 3.0 Flash, with a temperature of 0 (making the output the most deterministic possible) and a high thinking level. We chose this particular model because it is faster than other frontier ``Pro" models, like Gemini 3.0 Pro, allowing for learning \textit{during} collaborations as opposed to afterwards.

The next sections provide additional analyses that isolate each feedback modality. The evaluation protocol is the same as described before and in Sec.~\ref{sec:eval_mujoco}: we hold the recorded MuJoCo interaction trajectories fixed, apply each implication method to construct accepted and rejected feedback sets, and evaluate both implication-label prediction and downstream preference learning.

\textbf{Explicit feedback only.}
Fig.~\ref{fig:explicit_f1} and Table~\ref{tab:downstream_explicit} show results when preference learning uses only explicit feedback on the robot's executed actions and the corresponding implications for unchosen robot actions. The overall trend is consistent with the combined-feedback results: \implied{} led to a higher F1 score faster than the non-human baselines. Also, at the earliest task-completion step across participants (vertical dashed line in Fig.~\ref{fig:explicit_f1}), \implied{} achieved the highest F1-score in both tasks. In sandwich-making, linear mixed-effects analyses showed that \implied{} ($0.86\pm0.02$) achieved significantly higher F1-score than the non-human baselines at the earliest task-completion step ($p < 0.05$). Likewise, in block-sorting, \implied{} ($0.84\pm0.01$) got significantly higher F1 at that step ($p < 0.05$). In downstream learning (Table~\ref{tab:downstream_explicit}), a linear mixed-effects analysis for the sandwich-making task showed that \implied{} achieved significantly lower L2 error ($p < 0.01$) than all non-human baselines, except for \texttt{No-Prior}, and significantly lower conflict proportion ($p < 0.05$) than all non-human baselines. For block-sorting, \implied{} also significantly outperformed all other non-human baselines on L2 error ($p < 0.05$) and
conflict proportion ($p < 0.01$). These results suggest that, even when only explicit robot-action feedback is available, revising implications from accumulated direct evidence improves the feedback sets used for preference learning.

\textbf{Implicit feedback only.}
Fig.~\ref{fig:implicit_f1} and Table~\ref{tab:downstream_implicit} show results when preference learning uses only implicit feedback from the human's executed task actions and the corresponding implications for unchosen human actions. In block-sorting, the results again follow the main trend: \implied{} led to a higher F1 score faster than the non-human baselines. A linear mixed-effects analysis showed that \implied{} ($0.86\pm0.01$) achieved significantly higher F1-score than the non-human baselines at the earliest task-completion step (vertical dashed line in Fig.~\ref{fig:implicit_f1}), with $p < 0.0001$. In downstream learning, a linear mixed-effects analysis showed that \implied{} achieved significantly lower L2 error ($p < 0.05$) and conflict proportion ($p < 0.05$) than all non-human baselines in block sorting. This indicates that the method can also revise implications derived from human action choices, not only explicit evaluations of robot actions.

When learning in the sandwich-making task with implicit feedback only, we found different results than all of our prior results: \implied{} underperformed the Fixed-Rule baseline in terms of F1 classification score (Fig.~\ref{fig:implicit_f1}). 
We attribute this result to the design of the sandwich-making task. In this task, the implicit-feedback action space available to the human was more constrained than for the block-sorting task, and human implication labels for implicit feedback had very low inconsistency against the fixed rule ($0.04 \pm 0.01$ timestep inconsistencies and $0.03 \pm 0.01$ label inconsistencies). Despite this, \implied{} led to $0.84 \pm 0.05$ L2 Error on preference learning versus $0.86 \pm 0.06$ at the earliest task-completion step (which were not significantly different,  $p=0.3945$). Also, \implied{} led to very similar Conflict Proportion performance ($0.25 \pm 0.03$ vs. $0.24 \pm 0.03$), suggesting that  \implied{} does not hinder preference learning in a meaningful way when the fixed-rule prior is close to the optimal implications for a given task.



\section{Prompts for LLM Baselines}
\label{sec:llm_prompts}

Sec.~\ref{sec:eval_mujoco} compares \implied{} with four LLM baselines that differ in whether the model receives prior interaction history and/or the current preference-belief summary. Below, we show representative prompts from the most informative condition, \texttt{LLM-HB}, which includes both the interaction history (\texttt{H}) and the preference-belief description (\texttt{B}). The first prompt corresponds to the sandwich-making task; the second one to the block-sorting task. The other LLM baselines use the same prompt structure but remove the corresponding information parts: \texttt{LLM} removes both history and belief, \texttt{LLM-H} removes only the belief part, and \texttt{LLM-B} removes only the history part. Each prompt asks the LLM to label the current timestep's unchosen robot and human actions as \accepted{} or \rejected{} and to return strict JSON.

\begin{promptbox}
\textbf{Example \texttt{LLM-HB} prompt for the sandwich-making task.}
\par\medskip
\begingroup
\small
\linespread{0.95}\selectfont
\raggedright
\noindent\hspace*{0.00em}{\ttfamily You are a robot collaborating with a human to make a sandwich together.}\par
\par\smallskip
\noindent\hspace*{0.00em}{\ttfamily In the environment, there are three areas: the storage, the workspace, and the sandwich area.}\par
\par\smallskip
\noindent\hspace*{0.00em}{\ttfamily The storage initially contains ingredient instances from four ingredient types: cheese, bread, meat, and veggies.}\par
\noindent\hspace*{0.00em}{\ttfamily These ingredients are selected and used to make the sandwich.}\par
\par\smallskip
\noindent\hspace*{0.00em}{\ttfamily The workspace starts empty. You (Robot) can:}\par
\noindent\hspace*{0.00em}{\ttfamily - PICK ingredients from storage to your hand,}\par
\noindent\hspace*{0.00em}{\ttfamily - PUT ingredients from your hand to workspace,}\par
\noindent\hspace*{0.00em}{\ttfamily - PICK ingredients from workspace to your hand,}\par
\noindent\hspace*{0.00em}{\ttfamily - PUT ingredients from your hand back to storage,}\par
\noindent\hspace*{0.00em}{\ttfamily - WAIT.}\par
\noindent\hspace*{0.00em}{\ttfamily The workspace can contain at most 4 ingredients.}\par
\par\smallskip
\noindent\hspace*{0.00em}{\ttfamily The human can:}\par
\noindent\hspace*{0.00em}{\ttfamily - PICK ingredients from workspace,}\par
\noindent\hspace*{0.00em}{\ttfamily - PUT ingredients from hand to sandwich,}\par
\noindent\hspace*{0.00em}{\ttfamily - PUT ingredients from hand back to workspace,}\par
\noindent\hspace*{0.00em}{\ttfamily - WAIT.}\par
\par\smallskip
\noindent\hspace*{0.00em}{\ttfamily You can only reach storage/workspace. The human can only reach workspace/sandwich.}\par
\noindent\hspace*{0.00em}{\ttfamily In each step, you (Robot) and the human act simultaneously.}\par
\noindent\hspace*{0.00em}{\ttfamily The human also gives explicit binary feedback (positive or negative) on your executed action.}\par
\par\smallskip
\noindent\hspace*{0.00em}{\ttfamily The objective of the collaboration is to achieve the task goal efficiently while also ensuring that robot and human actions align with the human's preference for how the task should be performed.}\par
\par\smallskip
\noindent\hspace*{0.00em}{\ttfamily In the collaboration, the human has a fixed latent preference that defines how the goal should be achieved, and the human consistently follows this preference when selecting actions and providing feedback.}\par
\noindent\hspace*{0.00em}{\ttfamily The human's actions and explicit binary feedback to you reflect both task-goal progress and this preference.}\par
\par\smallskip
\noindent\hspace*{0.00em}{\ttfamily The goal of each sandwich-making task is to complete a specific target sandwich in as few steps as possible, and the task goal may change when a new task begins.}\par
\par\smallskip
\noindent\hspace*{0.00em}{\ttfamily The human's preference over the collaboration specifies how the target sandwich should be made, and this preference does not change across tasks.}\par
\noindent\hspace*{0.00em}{\ttfamily In particular, the human cares about the following aspects of the collaboration:}\par
\par\smallskip
\noindent\hspace*{0.00em}{\ttfamily 1) The order in which ingredient types are used (6 dimensions):}\par
\noindent\hspace*{0.00em}{\ttfamily - cheese before bread}\par
\noindent\hspace*{0.00em}{\ttfamily - cheese before meat}\par
\noindent\hspace*{0.00em}{\ttfamily - cheese before veggie}\par
\noindent\hspace*{0.00em}{\ttfamily - bread before meat}\par
\noindent\hspace*{0.00em}{\ttfamily - bread before veggie}\par
\noindent\hspace*{0.00em}{\ttfamily - meat before veggie}\par
\noindent\hspace*{0.00em}{\ttfamily Here, ingredient types are cheese, bread, meat, and veggie. Each dimension reflects whether one ingredient type should generally be completed before another.}\par
\noindent\hspace*{0.00em}{\ttfamily Example: 'cheese before bread' means required cheese ingredients should be placed before required bread ingredients.}\par
\par\smallskip
\noindent\hspace*{0.00em}{\ttfamily 2) How full the workspace should be during collaboration (4 dimensions):}\par
\noindent\hspace*{0.00em}{\ttfamily - whether the human prefers the workspace to contain at most 0 ingredients,}\par
\noindent\hspace*{0.00em}{\ttfamily - at most 1 ingredient,}\par
\noindent\hspace*{0.00em}{\ttfamily - at most 2 ingredients,}\par
\noindent\hspace*{0.00em}{\ttfamily - or at most 3 ingredients, whenever possible.}\par
\par\smallskip
\noindent\hspace*{0.00em}{\ttfamily In each task, you know the target sandwich recipe goal, but you do not know the human's latent preference. Infer it from interaction history and feedback.}\par
\par\smallskip
\noindent\hspace*{0.00em}{\normalfont\bfseries Below is the history of the collaboration tasks between you and the human:}\par
\par\smallskip
\noindent\hspace*{0.00em}{\normalfont\bfseries Task 1:}\par
\noindent\hspace*{0.00em}{\normalfont\bfseries Sandwich recipe goal: 1 cheese, 2 breads, 1 bacon, 1 ham, 1 tomato, 1 cucumber}\par
\par\smallskip
\noindent\hspace*{0.00em}{\ttfamily Step 0}\par
\noindent\hspace*{0.64em}{\ttfamily State:}\par
\noindent\hspace*{1.28em}{\ttfamily - human\_hand: (empty)}\par
\noindent\hspace*{1.28em}{\ttfamily - robot\_hand: (empty)}\par
\noindent\hspace*{1.28em}{\ttfamily - sandwich: (empty)}\par
\noindent\hspace*{1.28em}{\ttfamily - storage: bacon, beef\_patty, bread\_slice1, bread\_slice2, bread\_slice3, bread\_slice4, cheese1, cheese2, cheese3, cheese4, cucumber1, cucumber2, cucumber3, ham, tomato1, tomato2}\par
\noindent\hspace*{1.28em}{\ttfamily - workspace: (empty)}\par
\noindent\hspace*{0.64em}{\ttfamily Executed actions:}\par
\noindent\hspace*{1.28em}{\ttfamily - You (Robot): PICK cucumber2 storage (veggie)}\par
\noindent\hspace*{1.28em}{\ttfamily - Human: WAIT}\par
\noindent\hspace*{0.64em}{\ttfamily Human explicit feedback for your action: negative}\par
\par\smallskip
\noindent\hspace*{0.00em}{\ttfamily Step 1}\par
\noindent\hspace*{0.64em}{\ttfamily State:}\par
\noindent\hspace*{1.28em}{\ttfamily - human\_hand: (empty)}\par
\noindent\hspace*{1.28em}{\ttfamily - robot\_hand: cucumber2}\par
\noindent\hspace*{1.28em}{\ttfamily - sandwich: (empty)}\par
\noindent\hspace*{1.28em}{\ttfamily - storage: bacon, beef\_patty, bread\_slice1, bread\_slice2, bread\_slice3, bread\_slice4, cheese1, cheese2, cheese3, cheese4, cucumber1, cucumber3, ham, tomato1, tomato2}\par
\noindent\hspace*{1.28em}{\ttfamily - workspace: (empty)}\par
\noindent\hspace*{0.64em}{\ttfamily Executed actions:}\par
\noindent\hspace*{1.28em}{\ttfamily - You (Robot): PUT cucumber2 workspace (veggie)}\par
\noindent\hspace*{1.28em}{\ttfamily - Human: WAIT}\par
\noindent\hspace*{0.64em}{\ttfamily Human explicit feedback for your action: negative}\par
\par\smallskip
\noindent\hspace*{0.00em}{\ttfamily Step 2}\par
\noindent\hspace*{0.64em}{\ttfamily State:}\par
\noindent\hspace*{1.28em}{\ttfamily - human\_hand: (empty)}\par
\noindent\hspace*{1.28em}{\ttfamily - robot\_hand: (empty)}\par
\noindent\hspace*{1.28em}{\ttfamily - sandwich: (empty)}\par
\noindent\hspace*{1.28em}{\ttfamily - storage: bacon, beef\_patty, bread\_slice1, bread\_slice2, bread\_slice3, bread\_slice4, cheese1, cheese2, cheese3, cheese4, cucumber1, cucumber3, ham, tomato1, tomato2}\par
\noindent\hspace*{1.28em}{\ttfamily - workspace: cucumber2}\par
\noindent\hspace*{0.64em}{\ttfamily Executed actions:}\par
\noindent\hspace*{1.28em}{\ttfamily - You (Robot): PICK ham storage (meat)}\par
\noindent\hspace*{1.28em}{\ttfamily - Human: WAIT}\par
\noindent\hspace*{0.64em}{\ttfamily Human explicit feedback for your action: negative}\par
\par\smallskip
\noindent\hspace*{0.00em}{\ttfamily Step 3}\par
\noindent\hspace*{0.64em}{\ttfamily State:}\par
\noindent\hspace*{1.28em}{\ttfamily - human\_hand: (empty)}\par
\noindent\hspace*{1.28em}{\ttfamily - robot\_hand: ham}\par
\noindent\hspace*{1.28em}{\ttfamily - sandwich: (empty)}\par
\noindent\hspace*{1.28em}{\ttfamily - storage: bacon, beef\_patty, bread\_slice1, bread\_slice2, bread\_slice3, bread\_slice4, cheese1, cheese2, cheese3, cheese4, cucumber1, cucumber3, tomato1, tomato2}\par
\noindent\hspace*{1.28em}{\ttfamily - workspace: cucumber2}\par
\noindent\hspace*{0.64em}{\ttfamily Executed actions:}\par
\noindent\hspace*{1.28em}{\ttfamily - You (Robot): PUT ham workspace (meat)}\par
\noindent\hspace*{1.28em}{\ttfamily - Human: WAIT}\par
\noindent\hspace*{0.64em}{\ttfamily Human explicit feedback for your action: negative}\par
\par\smallskip
\noindent\hspace*{0.00em}{\ttfamily Step 4}\par
\noindent\hspace*{0.64em}{\ttfamily State:}\par
\noindent\hspace*{1.28em}{\ttfamily - human\_hand: (empty)}\par
\noindent\hspace*{1.28em}{\ttfamily - robot\_hand: (empty)}\par
\noindent\hspace*{1.28em}{\ttfamily - sandwich: (empty)}\par
\noindent\hspace*{1.28em}{\ttfamily - storage: bacon, beef\_patty, bread\_slice1, bread\_slice2, bread\_slice3, bread\_slice4, cheese1, cheese2, cheese3, cheese4, cucumber1, cucumber3, tomato1, tomato2}\par
\noindent\hspace*{1.28em}{\ttfamily - workspace: cucumber2, ham}\par
\noindent\hspace*{0.64em}{\ttfamily Executed actions:}\par
\noindent\hspace*{1.28em}{\ttfamily - You (Robot): PICK cheese3 storage (cheese)}\par
\noindent\hspace*{1.28em}{\ttfamily - Human: WAIT}\par
\noindent\hspace*{0.64em}{\ttfamily Human explicit feedback for your action: positive}\par
\par\smallskip
\noindent\hspace*{0.00em}{\ttfamily Step 5}\par
\noindent\hspace*{0.64em}{\ttfamily State:}\par
\noindent\hspace*{1.28em}{\ttfamily - human\_hand: (empty)}\par
\noindent\hspace*{1.28em}{\ttfamily - robot\_hand: cheese3}\par
\noindent\hspace*{1.28em}{\ttfamily - sandwich: (empty)}\par
\noindent\hspace*{1.28em}{\ttfamily - storage: bacon, beef\_patty, bread\_slice1, bread\_slice2, bread\_slice3, bread\_slice4, cheese1, cheese2, cheese4, cucumber1, cucumber3, tomato1, tomato2}\par
\noindent\hspace*{1.28em}{\ttfamily - workspace: cucumber2, ham}\par
\noindent\hspace*{0.64em}{\ttfamily Executed actions:}\par
\noindent\hspace*{1.28em}{\ttfamily - You (Robot): PUT cheese3 workspace (cheese)}\par
\noindent\hspace*{1.28em}{\ttfamily - Human: WAIT}\par
\noindent\hspace*{0.64em}{\ttfamily Human explicit feedback for your action: positive}\par
\par\smallskip
\noindent\hspace*{0.00em}{\ttfamily Step 6}\par
\noindent\hspace*{0.64em}{\ttfamily State:}\par
\noindent\hspace*{1.28em}{\ttfamily - human\_hand: (empty)}\par
\noindent\hspace*{1.28em}{\ttfamily - robot\_hand: (empty)}\par
\noindent\hspace*{1.28em}{\ttfamily - sandwich: (empty)}\par
\noindent\hspace*{1.28em}{\ttfamily - storage: bacon, beef\_patty, bread\_slice1, bread\_slice2, bread\_slice3, bread\_slice4, cheese1, cheese2, cheese4, cucumber1, cucumber3, tomato1, tomato2}\par
\noindent\hspace*{1.28em}{\ttfamily - workspace: cheese3, cucumber2, ham}\par
\noindent\hspace*{0.64em}{\ttfamily Executed actions:}\par
\noindent\hspace*{1.28em}{\ttfamily - You (Robot): PICK bread\_slice1 storage (bread)}\par
\noindent\hspace*{1.28em}{\ttfamily - Human: PICK cheese3 workspace (cheese)}\par
\noindent\hspace*{0.64em}{\ttfamily Human explicit feedback for your action: negative}\par
\par\smallskip
\noindent\hspace*{0.00em}{\ttfamily Step 7}\par
\noindent\hspace*{0.64em}{\ttfamily State:}\par
\noindent\hspace*{1.28em}{\ttfamily - human\_hand: cheese3}\par
\noindent\hspace*{1.28em}{\ttfamily - robot\_hand: bread\_slice1}\par
\noindent\hspace*{1.28em}{\ttfamily - sandwich: (empty)}\par
\noindent\hspace*{1.28em}{\ttfamily - storage: bacon, beef\_patty, bread\_slice2, bread\_slice3, bread\_slice4, cheese1, cheese2, cheese4, cucumber1, cucumber3, tomato1, tomato2}\par
\noindent\hspace*{1.28em}{\ttfamily - workspace: cucumber2, ham}\par
\noindent\hspace*{0.64em}{\ttfamily Executed actions:}\par
\noindent\hspace*{1.28em}{\ttfamily - You (Robot): PUT bread\_slice1 workspace (bread)}\par
\noindent\hspace*{1.28em}{\ttfamily - Human: PUT cheese3 sandwich (cheese)}\par
\noindent\hspace*{0.64em}{\ttfamily Human explicit feedback for your action: negative}\par
\par\smallskip
\noindent\hspace*{0.00em}{\ttfamily Step 8}\par
\noindent\hspace*{0.64em}{\ttfamily State:}\par
\noindent\hspace*{1.28em}{\ttfamily - human\_hand: (empty)}\par
\noindent\hspace*{1.28em}{\ttfamily - robot\_hand: (empty)}\par
\noindent\hspace*{1.28em}{\ttfamily - sandwich: cheese3}\par
\noindent\hspace*{1.28em}{\ttfamily - storage: bacon, beef\_patty, bread\_slice2, bread\_slice3, bread\_slice4, cheese1, cheese2, cheese4, cucumber1, cucumber3, tomato1, tomato2}\par
\noindent\hspace*{1.28em}{\ttfamily - workspace: bread\_slice1, cucumber2, ham}\par
\noindent\hspace*{0.64em}{\ttfamily Executed actions:}\par
\noindent\hspace*{1.28em}{\ttfamily - You (Robot): PICK bacon storage (meat)}\par
\noindent\hspace*{1.28em}{\ttfamily - Human: PICK ham workspace (meat)}\par
\noindent\hspace*{0.64em}{\ttfamily Human explicit feedback for your action: positive}\par
\par\smallskip
\noindent\hspace*{0.00em}{\ttfamily Step 9}\par
\noindent\hspace*{0.64em}{\ttfamily State:}\par
\noindent\hspace*{1.28em}{\ttfamily - human\_hand: ham}\par
\noindent\hspace*{1.28em}{\ttfamily - robot\_hand: bacon}\par
\noindent\hspace*{1.28em}{\ttfamily - sandwich: cheese3}\par
\noindent\hspace*{1.28em}{\ttfamily - storage: beef\_patty, bread\_slice2, bread\_slice3, bread\_slice4, cheese1, cheese2, cheese4, cucumber1, cucumber3, tomato1, tomato2}\par
\noindent\hspace*{1.28em}{\ttfamily - workspace: bread\_slice1, cucumber2}\par
\noindent\hspace*{0.64em}{\ttfamily Executed actions:}\par
\noindent\hspace*{1.28em}{\ttfamily - You (Robot): PUT bacon workspace (meat)}\par
\noindent\hspace*{1.28em}{\ttfamily - Human: PUT ham sandwich (meat)}\par
\noindent\hspace*{0.64em}{\ttfamily Human explicit feedback for your action: positive}\par
\par\smallskip
\noindent\hspace*{0.00em}{\ttfamily Step 10}\par
\noindent\hspace*{0.64em}{\ttfamily State:}\par
\noindent\hspace*{1.28em}{\ttfamily - human\_hand: (empty)}\par
\noindent\hspace*{1.28em}{\ttfamily - robot\_hand: (empty)}\par
\noindent\hspace*{1.28em}{\ttfamily - sandwich: cheese3, ham}\par
\noindent\hspace*{1.28em}{\ttfamily - storage: beef\_patty, bread\_slice2, bread\_slice3, bread\_slice4, cheese1, cheese2, cheese4, cucumber1, cucumber3, tomato1, tomato2}\par
\noindent\hspace*{1.28em}{\ttfamily - workspace: bacon, bread\_slice1, cucumber2}\par
\noindent\hspace*{0.64em}{\ttfamily Executed actions:}\par
\noindent\hspace*{1.28em}{\ttfamily - You (Robot): PICK bread\_slice2 storage (bread)}\par
\noindent\hspace*{1.28em}{\ttfamily - Human: PICK bacon workspace (meat)}\par
\noindent\hspace*{0.64em}{\ttfamily Human explicit feedback for your action: positive}\par
\par\smallskip
\noindent\hspace*{0.00em}{\ttfamily Step 11}\par
\noindent\hspace*{0.64em}{\ttfamily State:}\par
\noindent\hspace*{1.28em}{\ttfamily - human\_hand: bacon}\par
\noindent\hspace*{1.28em}{\ttfamily - robot\_hand: bread\_slice2}\par
\noindent\hspace*{1.28em}{\ttfamily - sandwich: cheese3, ham}\par
\noindent\hspace*{1.28em}{\ttfamily - storage: beef\_patty, bread\_slice3, bread\_slice4, cheese1, cheese2, cheese4, cucumber1, cucumber3, tomato1, tomato2}\par
\noindent\hspace*{1.28em}{\ttfamily - workspace: bread\_slice1, cucumber2}\par
\noindent\hspace*{0.64em}{\ttfamily Executed actions:}\par
\noindent\hspace*{1.28em}{\ttfamily - You (Robot): PUT bread\_slice2 workspace (bread)}\par
\noindent\hspace*{1.28em}{\ttfamily - Human: PUT bacon sandwich (meat)}\par
\noindent\hspace*{0.64em}{\ttfamily Human explicit feedback for your action: positive}\par
\par\smallskip
\noindent\hspace*{0.00em}{\ttfamily Step 12}\par
\noindent\hspace*{0.64em}{\ttfamily State:}\par
\noindent\hspace*{1.28em}{\ttfamily - human\_hand: (empty)}\par
\noindent\hspace*{1.28em}{\ttfamily - robot\_hand: (empty)}\par
\noindent\hspace*{1.28em}{\ttfamily - sandwich: cheese3, ham, bacon}\par
\noindent\hspace*{1.28em}{\ttfamily - storage: beef\_patty, bread\_slice3, bread\_slice4, cheese1, cheese2, cheese4, cucumber1, cucumber3, tomato1, tomato2}\par
\noindent\hspace*{1.28em}{\ttfamily - workspace: bread\_slice1, bread\_slice2, cucumber2}\par
\noindent\hspace*{0.64em}{\ttfamily Executed actions:}\par
\noindent\hspace*{1.28em}{\ttfamily - You (Robot): PICK tomato2 storage (veggie)}\par
\noindent\hspace*{1.28em}{\ttfamily - Human: PICK bread\_slice1 workspace (bread)}\par
\noindent\hspace*{0.64em}{\ttfamily Human explicit feedback for your action: positive}\par
\par\smallskip
\noindent\hspace*{0.00em}{\ttfamily Step 13}\par
\noindent\hspace*{0.64em}{\ttfamily State:}\par
\noindent\hspace*{1.28em}{\ttfamily - human\_hand: bread\_slice1}\par
\noindent\hspace*{1.28em}{\ttfamily - robot\_hand: tomato2}\par
\noindent\hspace*{1.28em}{\ttfamily - sandwich: cheese3, ham, bacon}\par
\noindent\hspace*{1.28em}{\ttfamily - storage: beef\_patty, bread\_slice3, bread\_slice4, cheese1, cheese2, cheese4, cucumber1, cucumber3, tomato1}\par
\noindent\hspace*{1.28em}{\ttfamily - workspace: bread\_slice2, cucumber2}\par
\noindent\hspace*{0.64em}{\ttfamily Executed actions:}\par
\noindent\hspace*{1.28em}{\ttfamily - You (Robot): PUT tomato2 workspace (veggie)}\par
\noindent\hspace*{1.28em}{\ttfamily - Human: PUT bread\_slice1 sandwich (bread)}\par
\noindent\hspace*{0.64em}{\ttfamily Human explicit feedback for your action: positive}\par
\par\smallskip
\noindent\hspace*{0.00em}{\ttfamily Step 14}\par
\noindent\hspace*{0.64em}{\ttfamily State:}\par
\noindent\hspace*{1.28em}{\ttfamily - human\_hand: (empty)}\par
\noindent\hspace*{1.28em}{\ttfamily - robot\_hand: (empty)}\par
\noindent\hspace*{1.28em}{\ttfamily - sandwich: cheese3, ham, bacon, bread\_slice1}\par
\noindent\hspace*{1.28em}{\ttfamily - storage: beef\_patty, bread\_slice3, bread\_slice4, cheese1, cheese2, cheese4, cucumber1, cucumber3, tomato1}\par
\noindent\hspace*{1.28em}{\ttfamily - workspace: bread\_slice2, cucumber2, tomato2}\par
\noindent\hspace*{0.64em}{\ttfamily Executed actions:}\par
\noindent\hspace*{1.28em}{\ttfamily - You (Robot): WAIT}\par
\noindent\hspace*{1.28em}{\ttfamily - Human: PICK tomato2 workspace (veggie)}\par
\noindent\hspace*{0.64em}{\ttfamily Human explicit feedback for your action: positive}\par
\par\smallskip
\noindent\hspace*{0.00em}{\ttfamily Step 15}\par
\noindent\hspace*{0.64em}{\ttfamily State:}\par
\noindent\hspace*{1.28em}{\ttfamily - human\_hand: tomato2}\par
\noindent\hspace*{1.28em}{\ttfamily - robot\_hand: (empty)}\par
\noindent\hspace*{1.28em}{\ttfamily - sandwich: cheese3, ham, bacon, bread\_slice1}\par
\noindent\hspace*{1.28em}{\ttfamily - storage: beef\_patty, bread\_slice3, bread\_slice4, cheese1, cheese2, cheese4, cucumber1, cucumber3, tomato1}\par
\noindent\hspace*{1.28em}{\ttfamily - workspace: bread\_slice2, cucumber2}\par
\noindent\hspace*{0.64em}{\ttfamily Executed actions:}\par
\noindent\hspace*{1.28em}{\ttfamily - You (Robot): WAIT}\par
\noindent\hspace*{1.28em}{\ttfamily - Human: PUT tomato2 sandwich (veggie)}\par
\noindent\hspace*{0.64em}{\ttfamily Human explicit feedback for your action: positive}\par
\par\smallskip
\noindent\hspace*{0.00em}{\ttfamily Step 16}\par
\noindent\hspace*{0.64em}{\ttfamily State:}\par
\noindent\hspace*{1.28em}{\ttfamily - human\_hand: (empty)}\par
\noindent\hspace*{1.28em}{\ttfamily - robot\_hand: (empty)}\par
\noindent\hspace*{1.28em}{\ttfamily - sandwich: cheese3, ham, bacon, bread\_slice1, tomato2}\par
\noindent\hspace*{1.28em}{\ttfamily - storage: beef\_patty, bread\_slice3, bread\_slice4, cheese1, cheese2, cheese4, cucumber1, cucumber3, tomato1}\par
\noindent\hspace*{1.28em}{\ttfamily - workspace: bread\_slice2, cucumber2}\par
\noindent\hspace*{0.64em}{\ttfamily Executed actions:}\par
\noindent\hspace*{1.28em}{\ttfamily - You (Robot): PICK bread\_slice2 workspace (bread)}\par
\noindent\hspace*{1.28em}{\ttfamily - Human: PICK cucumber2 workspace (veggie)}\par
\noindent\hspace*{0.64em}{\ttfamily Human explicit feedback for your action: negative}\par
\par\smallskip
\noindent\hspace*{0.00em}{\ttfamily Step 17}\par
\noindent\hspace*{0.64em}{\ttfamily State:}\par
\noindent\hspace*{1.28em}{\ttfamily - human\_hand: cucumber2}\par
\noindent\hspace*{1.28em}{\ttfamily - robot\_hand: bread\_slice2}\par
\noindent\hspace*{1.28em}{\ttfamily - sandwich: cheese3, ham, bacon, bread\_slice1, tomato2}\par
\noindent\hspace*{1.28em}{\ttfamily - storage: beef\_patty, bread\_slice3, bread\_slice4, cheese1, cheese2, cheese4, cucumber1, cucumber3, tomato1}\par
\noindent\hspace*{1.28em}{\ttfamily - workspace: (empty)}\par
\noindent\hspace*{0.64em}{\ttfamily Executed actions:}\par
\noindent\hspace*{1.28em}{\ttfamily - You (Robot): PUT bread\_slice2 workspace (bread)}\par
\noindent\hspace*{1.28em}{\ttfamily - Human: PUT cucumber2 sandwich (veggie)}\par
\noindent\hspace*{0.64em}{\ttfamily Human explicit feedback for your action: positive}\par
\par\smallskip
\noindent\hspace*{0.00em}{\ttfamily Step 18}\par
\noindent\hspace*{0.64em}{\ttfamily State:}\par
\noindent\hspace*{1.28em}{\ttfamily - human\_hand: (empty)}\par
\noindent\hspace*{1.28em}{\ttfamily - robot\_hand: (empty)}\par
\noindent\hspace*{1.28em}{\ttfamily - sandwich: cheese3, ham, bacon, bread\_slice1, tomato2, cucumber2}\par
\noindent\hspace*{1.28em}{\ttfamily - storage: beef\_patty, bread\_slice3, bread\_slice4, cheese1, cheese2, cheese4, cucumber1, cucumber3, tomato1}\par
\noindent\hspace*{1.28em}{\ttfamily - workspace: bread\_slice2}\par
\noindent\hspace*{0.64em}{\ttfamily Executed actions:}\par
\noindent\hspace*{1.28em}{\ttfamily - You (Robot): WAIT}\par
\noindent\hspace*{1.28em}{\ttfamily - Human: PICK bread\_slice2 workspace (bread)}\par
\noindent\hspace*{0.64em}{\ttfamily Human explicit feedback for your action: positive}\par
\par\smallskip
\noindent\hspace*{0.00em}{\ttfamily Step 19}\par
\noindent\hspace*{0.64em}{\ttfamily State:}\par
\noindent\hspace*{1.28em}{\ttfamily - human\_hand: bread\_slice2}\par
\noindent\hspace*{1.28em}{\ttfamily - robot\_hand: (empty)}\par
\noindent\hspace*{1.28em}{\ttfamily - sandwich: cheese3, ham, bacon, bread\_slice1, tomato2, cucumber2}\par
\noindent\hspace*{1.28em}{\ttfamily - storage: beef\_patty, bread\_slice3, bread\_slice4, cheese1, cheese2, cheese4, cucumber1, cucumber3, tomato1}\par
\noindent\hspace*{1.28em}{\ttfamily - workspace: (empty)}\par
\noindent\hspace*{0.64em}{\ttfamily Executed actions:}\par
\noindent\hspace*{1.28em}{\ttfamily - You (Robot): WAIT}\par
\noindent\hspace*{1.28em}{\ttfamily - Human: PUT bread\_slice2 sandwich (bread)}\par
\noindent\hspace*{0.64em}{\ttfamily Human explicit feedback for your action: positive}\par
\par\smallskip
\noindent\hspace*{0.00em}{\normalfont\bfseries Task 2:}\par
\noindent\hspace*{0.00em}{\normalfont\bfseries Sandwich recipe goal: 2 cheeses, 2 breads, 1 beef patty, 1 tomato, 1 cucumber}\par
\par\smallskip
\noindent\hspace*{0.00em}{\ttfamily Step 0}\par
\noindent\hspace*{0.64em}{\ttfamily State:}\par
\noindent\hspace*{1.28em}{\ttfamily - human\_hand: (empty)}\par
\noindent\hspace*{1.28em}{\ttfamily - robot\_hand: (empty)}\par
\noindent\hspace*{1.28em}{\ttfamily - sandwich: (empty)}\par
\noindent\hspace*{1.28em}{\ttfamily - storage: bacon, beef\_patty, bread\_slice1, bread\_slice2, bread\_slice3, bread\_slice4, cheese1, cheese2, cheese3, cheese4, cucumber1, cucumber2, cucumber3, ham, tomato1, tomato2}\par
\noindent\hspace*{1.28em}{\ttfamily - workspace: (empty)}\par
\noindent\hspace*{0.64em}{\ttfamily Executed actions:}\par
\noindent\hspace*{1.28em}{\ttfamily - You (Robot): PICK cheese2 storage (cheese)}\par
\noindent\hspace*{1.28em}{\ttfamily - Human: WAIT}\par
\noindent\hspace*{0.64em}{\ttfamily Human explicit feedback for your action: positive}\par
\par\smallskip
\noindent\hspace*{0.00em}{\ttfamily Step 1}\par
\noindent\hspace*{0.64em}{\ttfamily State:}\par
\noindent\hspace*{1.28em}{\ttfamily - human\_hand: (empty)}\par
\noindent\hspace*{1.28em}{\ttfamily - robot\_hand: cheese2}\par
\noindent\hspace*{1.28em}{\ttfamily - sandwich: (empty)}\par
\noindent\hspace*{1.28em}{\ttfamily - storage: bacon, beef\_patty, bread\_slice1, bread\_slice2, bread\_slice3, bread\_slice4, cheese1, cheese3, cheese4, cucumber1, cucumber2, cucumber3, ham, tomato1, tomato2}\par
\noindent\hspace*{1.28em}{\ttfamily - workspace: (empty)}\par
\noindent\hspace*{0.64em}{\ttfamily Executed actions:}\par
\noindent\hspace*{1.28em}{\ttfamily - You (Robot): PUT cheese2 workspace (cheese)}\par
\noindent\hspace*{1.28em}{\ttfamily - Human: WAIT}\par
\noindent\hspace*{0.64em}{\ttfamily Human explicit feedback for your action: positive}\par
\par\smallskip
\noindent\hspace*{0.00em}{\ttfamily Step 2}\par
\noindent\hspace*{0.64em}{\ttfamily State:}\par
\noindent\hspace*{1.28em}{\ttfamily - human\_hand: (empty)}\par
\noindent\hspace*{1.28em}{\ttfamily - robot\_hand: (empty)}\par
\noindent\hspace*{1.28em}{\ttfamily - sandwich: (empty)}\par
\noindent\hspace*{1.28em}{\ttfamily - storage: bacon, beef\_patty, bread\_slice1, bread\_slice2, bread\_slice3, bread\_slice4, cheese1, cheese3, cheese4, cucumber1, cucumber2, cucumber3, ham, tomato1, tomato2}\par
\noindent\hspace*{1.28em}{\ttfamily - workspace: cheese2}\par
\noindent\hspace*{0.64em}{\ttfamily Executed actions:}\par
\noindent\hspace*{1.28em}{\ttfamily - You (Robot): PICK cheese3 storage (cheese)}\par
\noindent\hspace*{1.28em}{\ttfamily - Human: PICK cheese2 workspace (cheese)}\par
\noindent\hspace*{0.64em}{\ttfamily Human explicit feedback for your action: positive}\par
\par\smallskip
\noindent\hspace*{0.00em}{\ttfamily Step 3}\par
\noindent\hspace*{0.64em}{\ttfamily State:}\par
\noindent\hspace*{1.28em}{\ttfamily - human\_hand: cheese2}\par
\noindent\hspace*{1.28em}{\ttfamily - robot\_hand: cheese3}\par
\noindent\hspace*{1.28em}{\ttfamily - sandwich: (empty)}\par
\noindent\hspace*{1.28em}{\ttfamily - storage: bacon, beef\_patty, bread\_slice1, bread\_slice2, bread\_slice3, bread\_slice4, cheese1, cheese4, cucumber1, cucumber2, cucumber3, ham, tomato1, tomato2}\par
\noindent\hspace*{1.28em}{\ttfamily - workspace: (empty)}\par
\noindent\hspace*{0.64em}{\ttfamily Executed actions:}\par
\noindent\hspace*{1.28em}{\ttfamily - You (Robot): PUT cheese3 workspace (cheese)}\par
\noindent\hspace*{1.28em}{\ttfamily - Human: PUT cheese2 sandwich (cheese)}\par
\noindent\hspace*{0.64em}{\ttfamily Human explicit feedback for your action: positive}\par
\par\smallskip
\noindent\hspace*{0.00em}{\normalfont\bfseries Current step (Task 2, Step 4):}\par
\par\smallskip
\noindent\hspace*{0.00em}{\normalfont\bfseries Sandwich recipe goal: 2 cheeses, 2 breads, 1 beef patty, 1 tomato, 1 cucumber}\par
\par\smallskip
\noindent\hspace*{0.00em}{\ttfamily - State:}\par
\noindent\hspace*{1.28em}{\ttfamily - human\_hand: (empty)}\par
\noindent\hspace*{1.28em}{\ttfamily - robot\_hand: (empty)}\par
\noindent\hspace*{1.28em}{\ttfamily - sandwich: cheese2}\par
\noindent\hspace*{1.28em}{\ttfamily - storage: bacon, beef\_patty, bread\_slice1, bread\_slice2, bread\_slice3, bread\_slice4, cheese1, cheese4, cucumber1, cucumber2, cucumber3, ham, tomato1, tomato2}\par
\noindent\hspace*{1.28em}{\ttfamily - workspace: cheese3}\par
\noindent\hspace*{0.00em}{\ttfamily - Executed actions:}\par
\noindent\hspace*{1.28em}{\ttfamily - You (Robot): WAIT}\par
\noindent\hspace*{1.28em}{\ttfamily - Human: PICK cheese3 workspace (cheese)}\par
\noindent\hspace*{0.00em}{\ttfamily - Human explicit feedback for your action: negative}\par
\par\smallskip
\noindent\hspace*{0.00em}{\normalfont\bfseries Latest belief over preference before incorporating the CURRENT step:}\par
\noindent\hspace*{0.00em}{\ttfamily - cheese before bread: 10/10 strong accept, 0/10 weak accept, 0/10 weak reject, 0/10 strong reject}\par
\noindent\hspace*{0.00em}{\ttfamily - cheese before meat: 3/10 strong accept, 7/10 weak accept, 0/10 weak reject, 0/10 strong reject}\par
\noindent\hspace*{0.00em}{\ttfamily - cheese before veggie: 0/10 strong accept, 9/10 weak accept, 1/10 weak reject, 0/10 strong reject}\par
\noindent\hspace*{0.00em}{\ttfamily - bread before meat: 0/10 strong accept, 0/10 weak accept, 10/10 weak reject, 0/10 strong reject}\par
\noindent\hspace*{0.00em}{\ttfamily - bread before veggie: 0/10 strong accept, 1/10 weak accept, 9/10 weak reject, 0/10 strong reject}\par
\noindent\hspace*{0.00em}{\ttfamily - meat before veggie: 3/10 strong accept, 7/10 weak accept, 0/10 weak reject, 0/10 strong reject}\par
\noindent\hspace*{0.00em}{\ttfamily - workspace at most 0 ingredients: 0/10 strong accept, 3/10 weak accept, 7/10 weak reject, 0/10 strong reject}\par
\noindent\hspace*{0.00em}{\ttfamily - workspace at most 1 ingredient: 0/10 strong accept, 0/10 weak accept, 10/10 weak reject, 0/10 strong reject}\par
\noindent\hspace*{0.00em}{\ttfamily - workspace at most 2 ingredients: 0/10 strong accept, 0/10 weak accept, 10/10 weak reject, 0/10 strong reject}\par
\noindent\hspace*{0.00em}{\ttfamily - workspace at most 3 ingredients: 0/10 strong accept, 3/10 weak accept, 7/10 weak reject, 0/10 strong reject}\par
\par\smallskip
\noindent\hspace*{0.00em}{\normalfont\bfseries - The unchosen actions:}\par
\noindent\hspace*{0.64em}{\ttfamily Your unchosen actions: PICK bacon storage (meat), PICK ham storage (meat), PICK beef\_patty storage (meat), PICK bread\_slice storage (bread), PICK cheese storage (cheese), PICK cheese3 workspace (cheese), PICK tomato storage (veggie), PICK cucumber storage (veggie)}\par
\noindent\hspace*{0.64em}{\ttfamily Human unchosen actions: WAIT}\par
\par\smallskip
\noindent\hspace*{0.00em}{\normalfont\bfseries Now, based on the collaboration history (states, actions, human feedback), and the latest belief over the human's preference before the CURRENT step, predict the human feedback for each action that you and the human did not choose, in the CURRENT step only, considering both the task goal and the human's preference.}\par
\par\smallskip
\noindent\hspace*{0.00em}{\normalfont\bfseries Label each unchosen action as:}\par
\noindent\hspace*{0.00em}{\ttfamily - accepted: accepted behavior with respect to both task-goal progress and inferred human preference.}\par
\noindent\hspace*{0.00em}{\ttfamily - rejected: rejected behavior with respect to task-goal progress and/or inferred human preference.}\par
\par\smallskip
\noindent\hspace*{0.00em}{\normalfont\bfseries Rules:}\par
\noindent\hspace*{0.00em}{\ttfamily 1) Only include unique action keys from the provided unchosen actions of you ("Robot") and the human ("Human").}\par
\noindent\hspace*{0.00em}{\ttfamily 2) Every label must be exactly 'accepted' or 'rejected'.}\par
\noindent\hspace*{0.00em}{\ttfamily 3) Return JSON only.}\par
\par\smallskip
\noindent\hspace*{0.00em}{\normalfont\bfseries Return STRICT JSON only with this shape:}\par
\noindent\hspace*{0.00em}{\ttfamily \{}\par
\noindent\hspace*{0.64em}{\ttfamily "Robot": \{}\par
\noindent\hspace*{1.28em}{\ttfamily "PICK bacon storage (meat)": "accepted\textbar{}rejected",}\par
\noindent\hspace*{1.28em}{\ttfamily "PICK ham storage (meat)": "accepted\textbar{}rejected",}\par
\noindent\hspace*{1.28em}{\ttfamily "PICK beef\_patty storage (meat)": "accepted\textbar{}rejected",}\par
\noindent\hspace*{1.28em}{\ttfamily "PICK bread\_slice storage (bread)": "accepted\textbar{}rejected",}\par
\noindent\hspace*{1.28em}{\ttfamily "PICK cheese storage (cheese)": "accepted\textbar{}rejected",}\par
\noindent\hspace*{1.28em}{\ttfamily "PICK cheese3 workspace (cheese)": "accepted\textbar{}rejected",}\par
\noindent\hspace*{1.28em}{\ttfamily "PICK tomato storage (veggie)": "accepted\textbar{}rejected",}\par
\noindent\hspace*{1.28em}{\ttfamily "PICK cucumber storage (veggie)": "accepted\textbar{}rejected"}\par
\noindent\hspace*{0.64em}{\ttfamily \},}\par
\noindent\hspace*{0.64em}{\ttfamily "Human": \{}\par
\noindent\hspace*{1.28em}{\ttfamily "WAIT": "accepted\textbar{}rejected"}\par
\noindent\hspace*{0.64em}{\ttfamily \}}\par
\noindent\hspace*{0.00em}{\ttfamily \}}\par
\endgroup
\end{promptbox}


\begin{promptbox}
\textbf{Example \texttt{LLM-HB} prompt for the block-sorting task.}
\par\medskip
\begingroup
\small
\linespread{0.95}\selectfont
\raggedright
\noindent\hspace*{0.00em}{\ttfamily You are a robot collaborating with a human to sort the blocks on the table together.}\par
\par\smallskip
\noindent\hspace*{0.00em}{\ttfamily In the environment, there are nine unique blocks:}\par
\noindent\hspace*{0.00em}{\ttfamily - blue square}\par
\noindent\hspace*{0.00em}{\ttfamily - blue polygon}\par
\noindent\hspace*{0.00em}{\ttfamily - blue trapezoid}\par
\noindent\hspace*{0.00em}{\ttfamily - pink square}\par
\noindent\hspace*{0.00em}{\ttfamily - pink polygon}\par
\noindent\hspace*{0.00em}{\ttfamily - pink trapezoid}\par
\noindent\hspace*{0.00em}{\ttfamily - yellow square}\par
\noindent\hspace*{0.00em}{\ttfamily - yellow polygon}\par
\noindent\hspace*{0.00em}{\ttfamily - yellow trapezoid}\par
\par\smallskip
\noindent\hspace*{0.00em}{\ttfamily From left to right, there are five areas: left panel, red bin, middle panel, green bin, right panel.}\par
\par\smallskip
\noindent\hspace*{0.00em}{\ttfamily The blocks are initially placed in the left panel, the middle panel, or the right panel.}\par
\par\smallskip
\noindent\hspace*{0.00em}{\ttfamily You can only reach the left panel, the middle panel, and the red bin.}\par
\noindent\hspace*{0.00em}{\ttfamily The human can only reach the middle panel, the right panel, and the green bin.}\par
\noindent\hspace*{0.00em}{\ttfamily The middle panel can be reached by both you and the human.}\par
\par\smallskip
\noindent\hspace*{0.00em}{\ttfamily Both you and the human can pick blocks from and place blocks into reachable areas. All areas can contain any number of blocks.}\par
\par\smallskip
\noindent\hspace*{0.00em}{\ttfamily In each step, you (Robot) and the human act simultaneously.}\par
\noindent\hspace*{0.00em}{\ttfamily The human also gives explicit binary feedback (positive or negative) on your executed action.}\par
\par\smallskip
\noindent\hspace*{0.00em}{\ttfamily The objective of the collaboration is to achieve the task goal efficiently while also ensuring that robot and human actions align with the human's preference for how the task should be performed.}\par
\par\smallskip
\noindent\hspace*{0.00em}{\ttfamily In the collaboration, the human has a fixed latent preference that defines how the goal should be achieved, and the human consistently follows this preference when selecting actions and providing feedback.}\par
\noindent\hspace*{0.00em}{\ttfamily The human's actions and explicit binary feedback to you reflect both task-goal progress and this preference.}\par
\par\smallskip
\noindent\hspace*{0.00em}{\ttfamily The goal of each block-sorting task is to place all blocks into either the red bin or the green bin in as few steps as possible.}\par
\par\smallskip
\noindent\hspace*{0.00em}{\ttfamily The human's preference over the collaboration specifies how the blocks should be sorted, and this preference does not change across tasks.}\par
\noindent\hspace*{0.00em}{\ttfamily In particular, the human cares about which bin each block should be sorted into.}\par
\par\smallskip
\noindent\hspace*{0.00em}{\ttfamily There are 9 preference dimensions, one for each unique block type:}\par
\noindent\hspace*{0.00em}{\ttfamily - blue square}\par
\noindent\hspace*{0.00em}{\ttfamily - blue trapezoid}\par
\noindent\hspace*{0.00em}{\ttfamily - blue polygon}\par
\noindent\hspace*{0.00em}{\ttfamily - pink square}\par
\noindent\hspace*{0.00em}{\ttfamily - pink trapezoid}\par
\noindent\hspace*{0.00em}{\ttfamily - pink polygon}\par
\noindent\hspace*{0.00em}{\ttfamily - yellow square}\par
\noindent\hspace*{0.00em}{\ttfamily - yellow trapezoid}\par
\noindent\hspace*{0.00em}{\ttfamily - yellow polygon}\par
\noindent\hspace*{0.00em}{\ttfamily For each block type, a positive preference means the human prefers that block in the red bin, while a negative preference means the human prefers that block in the green bin.}\par
\par\smallskip
\noindent\hspace*{0.00em}{\ttfamily In each task, you know the sorting goal, but you do not know the human's latent preference. Infer it from interaction history and feedback.}\par
\par\smallskip
\noindent\hspace*{0.00em}{\normalfont\bfseries Below is the history of the collaboration tasks between you and the human:}\par
\par\smallskip
\noindent\hspace*{0.00em}{\normalfont\bfseries Task 1:}\par
\noindent\hspace*{0.00em}{\normalfont\bfseries Block-sorting goal: place all blocks into either the red bin or the green bin}\par
\par\smallskip
\noindent\hspace*{0.00em}{\ttfamily Step 0}\par
\noindent\hspace*{0.64em}{\ttfamily State:}\par
\noindent\hspace*{1.28em}{\ttfamily - human\_hand: (empty)}\par
\noindent\hspace*{1.28em}{\ttfamily - robot\_hand: (empty)}\par
\noindent\hspace*{1.28em}{\ttfamily - left: blue\_square, pink\_trapezoid}\par
\noindent\hspace*{1.28em}{\ttfamily - red: (empty)}\par
\noindent\hspace*{1.28em}{\ttfamily - mid: blue\_polygon, pink\_square, yellow\_square}\par
\noindent\hspace*{1.28em}{\ttfamily - green: (empty)}\par
\noindent\hspace*{1.28em}{\ttfamily - right: blue\_trapezoid, pink\_polygon, yellow\_polygon, yellow\_trapezoid}\par
\noindent\hspace*{0.64em}{\ttfamily Executed actions:}\par
\noindent\hspace*{1.28em}{\ttfamily - You (Robot): PICK pink\_square mid}\par
\noindent\hspace*{1.28em}{\ttfamily - Human: PICK yellow\_trapezoid right}\par
\noindent\hspace*{0.64em}{\ttfamily Human explicit feedback for your action: negative}\par
\par\smallskip
\noindent\hspace*{0.00em}{\ttfamily Step 1}\par
\noindent\hspace*{0.64em}{\ttfamily State:}\par
\noindent\hspace*{1.28em}{\ttfamily - human\_hand: yellow\_trapezoid}\par
\noindent\hspace*{1.28em}{\ttfamily - robot\_hand: pink\_square}\par
\noindent\hspace*{1.28em}{\ttfamily - left: blue\_square, pink\_trapezoid}\par
\noindent\hspace*{1.28em}{\ttfamily - red: (empty)}\par
\noindent\hspace*{1.28em}{\ttfamily - mid: blue\_polygon, yellow\_square}\par
\noindent\hspace*{1.28em}{\ttfamily - green: (empty)}\par
\noindent\hspace*{1.28em}{\ttfamily - right: blue\_trapezoid, pink\_polygon, yellow\_polygon}\par
\noindent\hspace*{0.64em}{\ttfamily Executed actions:}\par
\noindent\hspace*{1.28em}{\ttfamily - You (Robot): PLACE pink\_square red}\par
\noindent\hspace*{1.28em}{\ttfamily - Human: PLACE yellow\_trapezoid mid}\par
\noindent\hspace*{0.64em}{\ttfamily Human explicit feedback for your action: negative}\par
\par\smallskip
\noindent\hspace*{0.00em}{\ttfamily Step 2}\par
\noindent\hspace*{0.64em}{\ttfamily State:}\par
\noindent\hspace*{1.28em}{\ttfamily - human\_hand: (empty)}\par
\noindent\hspace*{1.28em}{\ttfamily - robot\_hand: (empty)}\par
\noindent\hspace*{1.28em}{\ttfamily - left: blue\_square, pink\_trapezoid}\par
\noindent\hspace*{1.28em}{\ttfamily - red: pink\_square}\par
\noindent\hspace*{1.28em}{\ttfamily - mid: blue\_polygon, yellow\_square, yellow\_trapezoid}\par
\noindent\hspace*{1.28em}{\ttfamily - green: (empty)}\par
\noindent\hspace*{1.28em}{\ttfamily - right: blue\_trapezoid, pink\_polygon, yellow\_polygon}\par
\noindent\hspace*{0.64em}{\ttfamily Executed actions:}\par
\noindent\hspace*{1.28em}{\ttfamily - You (Robot): PICK yellow\_trapezoid mid}\par
\noindent\hspace*{1.28em}{\ttfamily - Human: PICK blue\_trapezoid right}\par
\noindent\hspace*{0.64em}{\ttfamily Human explicit feedback for your action: positive}\par
\par\smallskip
\noindent\hspace*{0.00em}{\ttfamily Step 3}\par
\noindent\hspace*{0.64em}{\ttfamily State:}\par
\noindent\hspace*{1.28em}{\ttfamily - human\_hand: blue\_trapezoid}\par
\noindent\hspace*{1.28em}{\ttfamily - robot\_hand: yellow\_trapezoid}\par
\noindent\hspace*{1.28em}{\ttfamily - left: blue\_square, pink\_trapezoid}\par
\noindent\hspace*{1.28em}{\ttfamily - red: pink\_square}\par
\noindent\hspace*{1.28em}{\ttfamily - mid: blue\_polygon, yellow\_square}\par
\noindent\hspace*{1.28em}{\ttfamily - green: (empty)}\par
\noindent\hspace*{1.28em}{\ttfamily - right: pink\_polygon, yellow\_polygon}\par
\noindent\hspace*{0.64em}{\ttfamily Executed actions:}\par
\noindent\hspace*{1.28em}{\ttfamily - You (Robot): PLACE yellow\_trapezoid red}\par
\noindent\hspace*{1.28em}{\ttfamily - Human: PLACE blue\_trapezoid mid}\par
\noindent\hspace*{0.64em}{\ttfamily Human explicit feedback for your action: positive}\par
\par\smallskip
\noindent\hspace*{0.00em}{\ttfamily Step 4}\par
\noindent\hspace*{0.64em}{\ttfamily State:}\par
\noindent\hspace*{1.28em}{\ttfamily - human\_hand: (empty)}\par
\noindent\hspace*{1.28em}{\ttfamily - robot\_hand: (empty)}\par
\noindent\hspace*{1.28em}{\ttfamily - left: blue\_square, pink\_trapezoid}\par
\noindent\hspace*{1.28em}{\ttfamily - red: pink\_square, yellow\_trapezoid}\par
\noindent\hspace*{1.28em}{\ttfamily - mid: blue\_polygon, blue\_trapezoid, yellow\_square}\par
\noindent\hspace*{1.28em}{\ttfamily - green: (empty)}\par
\noindent\hspace*{1.28em}{\ttfamily - right: pink\_polygon, yellow\_polygon}\par
\noindent\hspace*{0.64em}{\ttfamily Executed actions:}\par
\noindent\hspace*{1.28em}{\ttfamily - You (Robot): PICK pink\_square red}\par
\noindent\hspace*{1.28em}{\ttfamily - Human: PICK blue\_polygon mid}\par
\noindent\hspace*{0.64em}{\ttfamily Human explicit feedback for your action: positive}\par
\par\smallskip
\noindent\hspace*{0.00em}{\ttfamily Step 5}\par
\noindent\hspace*{0.64em}{\ttfamily State:}\par
\noindent\hspace*{1.28em}{\ttfamily - human\_hand: blue\_polygon}\par
\noindent\hspace*{1.28em}{\ttfamily - robot\_hand: pink\_square}\par
\noindent\hspace*{1.28em}{\ttfamily - left: blue\_square, pink\_trapezoid}\par
\noindent\hspace*{1.28em}{\ttfamily - red: yellow\_trapezoid}\par
\noindent\hspace*{1.28em}{\ttfamily - mid: blue\_trapezoid, yellow\_square}\par
\noindent\hspace*{1.28em}{\ttfamily - green: (empty)}\par
\noindent\hspace*{1.28em}{\ttfamily - right: pink\_polygon, yellow\_polygon}\par
\noindent\hspace*{0.64em}{\ttfamily Executed actions:}\par
\noindent\hspace*{1.28em}{\ttfamily - You (Robot): PLACE pink\_square mid}\par
\noindent\hspace*{1.28em}{\ttfamily - Human: PLACE blue\_polygon green}\par
\noindent\hspace*{0.64em}{\ttfamily Human explicit feedback for your action: positive}\par
\par\smallskip
\noindent\hspace*{0.00em}{\ttfamily Step 6}\par
\noindent\hspace*{0.64em}{\ttfamily State:}\par
\noindent\hspace*{1.28em}{\ttfamily - human\_hand: (empty)}\par
\noindent\hspace*{1.28em}{\ttfamily - robot\_hand: (empty)}\par
\noindent\hspace*{1.28em}{\ttfamily - left: blue\_square, pink\_trapezoid}\par
\noindent\hspace*{1.28em}{\ttfamily - red: yellow\_trapezoid}\par
\noindent\hspace*{1.28em}{\ttfamily - mid: blue\_trapezoid, pink\_square, yellow\_square}\par
\noindent\hspace*{1.28em}{\ttfamily - green: blue\_polygon}\par
\noindent\hspace*{1.28em}{\ttfamily - right: pink\_polygon, yellow\_polygon}\par
\noindent\hspace*{0.64em}{\ttfamily Executed actions:}\par
\noindent\hspace*{1.28em}{\ttfamily - You (Robot): PICK blue\_trapezoid mid}\par
\noindent\hspace*{1.28em}{\ttfamily - Human: PICK pink\_square mid}\par
\noindent\hspace*{0.64em}{\ttfamily Human explicit feedback for your action: positive}\par
\par\smallskip
\noindent\hspace*{0.00em}{\ttfamily Step 7}\par
\noindent\hspace*{0.64em}{\ttfamily State:}\par
\noindent\hspace*{1.28em}{\ttfamily - human\_hand: pink\_square}\par
\noindent\hspace*{1.28em}{\ttfamily - robot\_hand: blue\_trapezoid}\par
\noindent\hspace*{1.28em}{\ttfamily - left: blue\_square, pink\_trapezoid}\par
\noindent\hspace*{1.28em}{\ttfamily - red: yellow\_trapezoid}\par
\noindent\hspace*{1.28em}{\ttfamily - mid: yellow\_square}\par
\noindent\hspace*{1.28em}{\ttfamily - green: blue\_polygon}\par
\noindent\hspace*{1.28em}{\ttfamily - right: pink\_polygon, yellow\_polygon}\par
\noindent\hspace*{0.64em}{\ttfamily Executed actions:}\par
\noindent\hspace*{1.28em}{\ttfamily - You (Robot): PLACE blue\_trapezoid red}\par
\noindent\hspace*{1.28em}{\ttfamily - Human: PLACE pink\_square green}\par
\noindent\hspace*{0.64em}{\ttfamily Human explicit feedback for your action: positive}\par
\par\smallskip
\noindent\hspace*{0.00em}{\ttfamily Step 8}\par
\noindent\hspace*{0.64em}{\ttfamily State:}\par
\noindent\hspace*{1.28em}{\ttfamily - human\_hand: (empty)}\par
\noindent\hspace*{1.28em}{\ttfamily - robot\_hand: (empty)}\par
\noindent\hspace*{1.28em}{\ttfamily - left: blue\_square, pink\_trapezoid}\par
\noindent\hspace*{1.28em}{\ttfamily - red: blue\_trapezoid, yellow\_trapezoid}\par
\noindent\hspace*{1.28em}{\ttfamily - mid: yellow\_square}\par
\noindent\hspace*{1.28em}{\ttfamily - green: blue\_polygon, pink\_square}\par
\noindent\hspace*{1.28em}{\ttfamily - right: pink\_polygon, yellow\_polygon}\par
\noindent\hspace*{0.64em}{\ttfamily Executed actions:}\par
\noindent\hspace*{1.28em}{\ttfamily - You (Robot): WAIT}\par
\noindent\hspace*{1.28em}{\ttfamily - Human: PICK pink\_polygon right}\par
\noindent\hspace*{0.64em}{\ttfamily Human explicit feedback for your action: negative}\par
\par\smallskip
\noindent\hspace*{0.00em}{\ttfamily Step 9}\par
\noindent\hspace*{0.64em}{\ttfamily State:}\par
\noindent\hspace*{1.28em}{\ttfamily - human\_hand: pink\_polygon}\par
\noindent\hspace*{1.28em}{\ttfamily - robot\_hand: (empty)}\par
\noindent\hspace*{1.28em}{\ttfamily - left: blue\_square, pink\_trapezoid}\par
\noindent\hspace*{1.28em}{\ttfamily - red: blue\_trapezoid, yellow\_trapezoid}\par
\noindent\hspace*{1.28em}{\ttfamily - mid: yellow\_square}\par
\noindent\hspace*{1.28em}{\ttfamily - green: blue\_polygon, pink\_square}\par
\noindent\hspace*{1.28em}{\ttfamily - right: yellow\_polygon}\par
\noindent\hspace*{0.64em}{\ttfamily Executed actions:}\par
\noindent\hspace*{1.28em}{\ttfamily - You (Robot): PICK blue\_square left}\par
\noindent\hspace*{1.28em}{\ttfamily - Human: PLACE pink\_polygon green}\par
\noindent\hspace*{0.64em}{\ttfamily Human explicit feedback for your action: positive}\par
\par\smallskip
\noindent\hspace*{0.00em}{\ttfamily Step 10}\par
\noindent\hspace*{0.64em}{\ttfamily State:}\par
\noindent\hspace*{1.28em}{\ttfamily - human\_hand: (empty)}\par
\noindent\hspace*{1.28em}{\ttfamily - robot\_hand: blue\_square}\par
\noindent\hspace*{1.28em}{\ttfamily - left: pink\_trapezoid}\par
\noindent\hspace*{1.28em}{\ttfamily - red: blue\_trapezoid, yellow\_trapezoid}\par
\noindent\hspace*{1.28em}{\ttfamily - mid: yellow\_square}\par
\noindent\hspace*{1.28em}{\ttfamily - green: blue\_polygon, pink\_polygon, pink\_square}\par
\noindent\hspace*{1.28em}{\ttfamily - right: yellow\_polygon}\par
\noindent\hspace*{0.64em}{\ttfamily Executed actions:}\par
\noindent\hspace*{1.28em}{\ttfamily - You (Robot): PLACE blue\_square red}\par
\noindent\hspace*{1.28em}{\ttfamily - Human: PICK yellow\_polygon right}\par
\noindent\hspace*{0.64em}{\ttfamily Human explicit feedback for your action: negative}\par
\par\smallskip
\noindent\hspace*{0.00em}{\ttfamily Step 11}\par
\noindent\hspace*{0.64em}{\ttfamily State:}\par
\noindent\hspace*{1.28em}{\ttfamily - human\_hand: yellow\_polygon}\par
\noindent\hspace*{1.28em}{\ttfamily - robot\_hand: (empty)}\par
\noindent\hspace*{1.28em}{\ttfamily - left: pink\_trapezoid}\par
\noindent\hspace*{1.28em}{\ttfamily - red: blue\_square, blue\_trapezoid, yellow\_trapezoid}\par
\noindent\hspace*{1.28em}{\ttfamily - mid: yellow\_square}\par
\noindent\hspace*{1.28em}{\ttfamily - green: blue\_polygon, pink\_polygon, pink\_square}\par
\noindent\hspace*{1.28em}{\ttfamily - right: (empty)}\par
\noindent\hspace*{0.64em}{\ttfamily Executed actions:}\par
\noindent\hspace*{1.28em}{\ttfamily - You (Robot): PICK pink\_trapezoid left}\par
\noindent\hspace*{1.28em}{\ttfamily - Human: PLACE yellow\_polygon green}\par
\noindent\hspace*{0.64em}{\ttfamily Human explicit feedback for your action: positive}\par
\par\smallskip
\noindent\hspace*{0.00em}{\ttfamily Step 12}\par
\noindent\hspace*{0.64em}{\ttfamily State:}\par
\noindent\hspace*{1.28em}{\ttfamily - human\_hand: (empty)}\par
\noindent\hspace*{1.28em}{\ttfamily - robot\_hand: pink\_trapezoid}\par
\noindent\hspace*{1.28em}{\ttfamily - left: (empty)}\par
\noindent\hspace*{1.28em}{\ttfamily - red: blue\_square, blue\_trapezoid, yellow\_trapezoid}\par
\noindent\hspace*{1.28em}{\ttfamily - mid: yellow\_square}\par
\noindent\hspace*{1.28em}{\ttfamily - green: blue\_polygon, pink\_polygon, pink\_square, yellow\_polygon}\par
\noindent\hspace*{1.28em}{\ttfamily - right: (empty)}\par
\noindent\hspace*{0.64em}{\ttfamily Executed actions:}\par
\noindent\hspace*{1.28em}{\ttfamily - You (Robot): PLACE pink\_trapezoid mid}\par
\noindent\hspace*{1.28em}{\ttfamily - Human: PICK yellow\_square mid}\par
\noindent\hspace*{0.64em}{\ttfamily Human explicit feedback for your action: negative}\par
\par\smallskip
\noindent\hspace*{0.00em}{\ttfamily Step 13}\par
\noindent\hspace*{0.64em}{\ttfamily State:}\par
\noindent\hspace*{1.28em}{\ttfamily - human\_hand: yellow\_square}\par
\noindent\hspace*{1.28em}{\ttfamily - robot\_hand: (empty)}\par
\noindent\hspace*{1.28em}{\ttfamily - left: (empty)}\par
\noindent\hspace*{1.28em}{\ttfamily - red: blue\_square, blue\_trapezoid, yellow\_trapezoid}\par
\noindent\hspace*{1.28em}{\ttfamily - mid: pink\_trapezoid}\par
\noindent\hspace*{1.28em}{\ttfamily - green: blue\_polygon, pink\_polygon, pink\_square, yellow\_polygon}\par
\noindent\hspace*{1.28em}{\ttfamily - right: (empty)}\par
\noindent\hspace*{0.64em}{\ttfamily Executed actions:}\par
\noindent\hspace*{1.28em}{\ttfamily - You (Robot): PICK pink\_trapezoid mid}\par
\noindent\hspace*{1.28em}{\ttfamily - Human: PLACE yellow\_square green}\par
\noindent\hspace*{0.64em}{\ttfamily Human explicit feedback for your action: positive}\par
\par\smallskip
\noindent\hspace*{0.00em}{\ttfamily Step 14}\par
\noindent\hspace*{0.64em}{\ttfamily State:}\par
\noindent\hspace*{1.28em}{\ttfamily - human\_hand: (empty)}\par
\noindent\hspace*{1.28em}{\ttfamily - robot\_hand: pink\_trapezoid}\par
\noindent\hspace*{1.28em}{\ttfamily - left: (empty)}\par
\noindent\hspace*{1.28em}{\ttfamily - red: blue\_square, blue\_trapezoid, yellow\_trapezoid}\par
\noindent\hspace*{1.28em}{\ttfamily - mid: (empty)}\par
\noindent\hspace*{1.28em}{\ttfamily - green: blue\_polygon, pink\_polygon, pink\_square, yellow\_polygon, yellow\_square}\par
\noindent\hspace*{1.28em}{\ttfamily - right: (empty)}\par
\noindent\hspace*{0.64em}{\ttfamily Executed actions:}\par
\noindent\hspace*{1.28em}{\ttfamily - You (Robot): PLACE pink\_trapezoid red}\par
\noindent\hspace*{1.28em}{\ttfamily - Human: WAIT}\par
\noindent\hspace*{0.64em}{\ttfamily Human explicit feedback for your action: positive}\par
\par\smallskip
\noindent\hspace*{0.00em}{\normalfont\bfseries Task 2:}\par
\noindent\hspace*{0.00em}{\normalfont\bfseries Block-sorting goal: place all blocks into either the red bin or the green bin}\par
\par\smallskip
\noindent\hspace*{0.00em}{\ttfamily Step 0}\par
\noindent\hspace*{0.64em}{\ttfamily State:}\par
\noindent\hspace*{1.28em}{\ttfamily - human\_hand: (empty)}\par
\noindent\hspace*{1.28em}{\ttfamily - robot\_hand: (empty)}\par
\noindent\hspace*{1.28em}{\ttfamily - left: blue\_polygon, blue\_square, pink\_square, yellow\_square}\par
\noindent\hspace*{1.28em}{\ttfamily - red: (empty)}\par
\noindent\hspace*{1.28em}{\ttfamily - mid: pink\_polygon}\par
\noindent\hspace*{1.28em}{\ttfamily - green: (empty)}\par
\noindent\hspace*{1.28em}{\ttfamily - right: blue\_trapezoid, pink\_trapezoid, yellow\_polygon, yellow\_trapezoid}\par
\noindent\hspace*{0.64em}{\ttfamily Executed actions:}\par
\noindent\hspace*{1.28em}{\ttfamily - You (Robot): PICK pink\_square left}\par
\noindent\hspace*{1.28em}{\ttfamily - Human: PICK blue\_trapezoid right}\par
\noindent\hspace*{0.64em}{\ttfamily Human explicit feedback for your action: positive}\par
\par\smallskip
\noindent\hspace*{0.00em}{\ttfamily Step 1}\par
\noindent\hspace*{0.64em}{\ttfamily State:}\par
\noindent\hspace*{1.28em}{\ttfamily - human\_hand: blue\_trapezoid}\par
\noindent\hspace*{1.28em}{\ttfamily - robot\_hand: pink\_square}\par
\noindent\hspace*{1.28em}{\ttfamily - left: blue\_polygon, blue\_square, yellow\_square}\par
\noindent\hspace*{1.28em}{\ttfamily - red: (empty)}\par
\noindent\hspace*{1.28em}{\ttfamily - mid: pink\_polygon}\par
\noindent\hspace*{1.28em}{\ttfamily - green: (empty)}\par
\noindent\hspace*{1.28em}{\ttfamily - right: pink\_trapezoid, yellow\_polygon, yellow\_trapezoid}\par
\noindent\hspace*{0.64em}{\ttfamily Executed actions:}\par
\noindent\hspace*{1.28em}{\ttfamily - You (Robot): WAIT}\par
\noindent\hspace*{1.28em}{\ttfamily - Human: PLACE blue\_trapezoid mid}\par
\noindent\hspace*{0.64em}{\ttfamily Human explicit feedback for your action: negative}\par
\par\smallskip
\noindent\hspace*{0.00em}{\ttfamily Step 2}\par
\noindent\hspace*{0.64em}{\ttfamily State:}\par
\noindent\hspace*{1.28em}{\ttfamily - human\_hand: (empty)}\par
\noindent\hspace*{1.28em}{\ttfamily - robot\_hand: pink\_square}\par
\noindent\hspace*{1.28em}{\ttfamily - left: blue\_polygon, blue\_square, yellow\_square}\par
\noindent\hspace*{1.28em}{\ttfamily - red: (empty)}\par
\noindent\hspace*{1.28em}{\ttfamily - mid: blue\_trapezoid, pink\_polygon}\par
\noindent\hspace*{1.28em}{\ttfamily - green: (empty)}\par
\noindent\hspace*{1.28em}{\ttfamily - right: pink\_trapezoid, yellow\_polygon, yellow\_trapezoid}\par
\noindent\hspace*{0.64em}{\ttfamily Executed actions:}\par
\noindent\hspace*{1.28em}{\ttfamily - You (Robot): PLACE pink\_square mid}\par
\noindent\hspace*{1.28em}{\ttfamily - Human: PICK pink\_trapezoid right}\par
\noindent\hspace*{0.64em}{\ttfamily Human explicit feedback for your action: positive}\par
\par\smallskip
\noindent\hspace*{0.00em}{\ttfamily Step 3}\par
\noindent\hspace*{0.64em}{\ttfamily State:}\par
\noindent\hspace*{1.28em}{\ttfamily - human\_hand: pink\_trapezoid}\par
\noindent\hspace*{1.28em}{\ttfamily - robot\_hand: (empty)}\par
\noindent\hspace*{1.28em}{\ttfamily - left: blue\_polygon, blue\_square, yellow\_square}\par
\noindent\hspace*{1.28em}{\ttfamily - red: (empty)}\par
\noindent\hspace*{1.28em}{\ttfamily - mid: blue\_trapezoid, pink\_polygon, pink\_square}\par
\noindent\hspace*{1.28em}{\ttfamily - green: (empty)}\par
\noindent\hspace*{1.28em}{\ttfamily - right: yellow\_polygon, yellow\_trapezoid}\par
\noindent\hspace*{0.64em}{\ttfamily Executed actions:}\par
\noindent\hspace*{1.28em}{\ttfamily - You (Robot): PICK blue\_polygon left}\par
\noindent\hspace*{1.28em}{\ttfamily - Human: PLACE pink\_trapezoid mid}\par
\noindent\hspace*{0.64em}{\ttfamily Human explicit feedback for your action: positive}\par
\par\smallskip
\noindent\hspace*{0.00em}{\normalfont\bfseries Current step (Task 2, Step 4):}\par
\par\smallskip
\noindent\hspace*{0.00em}{\normalfont\bfseries Block-sorting goal: place all blocks into either the red bin or the green bin}\par
\par\smallskip
\noindent\hspace*{0.00em}{\ttfamily - State:}\par
\noindent\hspace*{1.28em}{\ttfamily - human\_hand: (empty)}\par
\noindent\hspace*{1.28em}{\ttfamily - robot\_hand: blue\_polygon}\par
\noindent\hspace*{1.28em}{\ttfamily - left: blue\_square, yellow\_square}\par
\noindent\hspace*{1.28em}{\ttfamily - red: (empty)}\par
\noindent\hspace*{1.28em}{\ttfamily - mid: blue\_trapezoid, pink\_polygon, pink\_square, pink\_trapezoid}\par
\noindent\hspace*{1.28em}{\ttfamily - green: (empty)}\par
\noindent\hspace*{1.28em}{\ttfamily - right: yellow\_polygon, yellow\_trapezoid}\par
\noindent\hspace*{0.00em}{\ttfamily - Executed actions:}\par
\noindent\hspace*{1.28em}{\ttfamily - You (Robot): PLACE blue\_polygon mid}\par
\noindent\hspace*{1.28em}{\ttfamily - Human: PICK pink\_polygon mid}\par
\noindent\hspace*{0.00em}{\ttfamily - Human explicit feedback for your action: positive}\par
\par\smallskip
\noindent\hspace*{0.00em}{\normalfont\bfseries Latest belief over preference before incorporating the CURRENT step:}\par
\noindent\hspace*{0.00em}{\ttfamily - blue square to red bin: 3/10 strong accept, 3/10 weak accept, 3/10 weak reject, 1/10 strong reject}\par
\noindent\hspace*{0.00em}{\ttfamily - blue trapezoid to red bin: 2/10 strong accept, 8/10 weak accept, 0/10 weak reject, 0/10 strong reject}\par
\noindent\hspace*{0.00em}{\ttfamily - blue polygon to red bin: 0/10 strong accept, 0/10 weak accept, 9/10 weak reject, 1/10 strong reject}\par
\noindent\hspace*{0.00em}{\ttfamily - pink square to red bin: 0/10 strong accept, 0/10 weak accept, 5/10 weak reject, 5/10 strong reject}\par
\noindent\hspace*{0.00em}{\ttfamily - pink trapezoid to red bin: 3/10 strong accept, 7/10 weak accept, 0/10 weak reject, 0/10 strong reject}\par
\noindent\hspace*{0.00em}{\ttfamily - pink polygon to red bin: 0/10 strong accept, 3/10 weak accept, 7/10 weak reject, 0/10 strong reject}\par
\noindent\hspace*{0.00em}{\ttfamily - yellow square to red bin: 0/10 strong accept, 0/10 weak accept, 10/10 weak reject, 0/10 strong reject}\par
\noindent\hspace*{0.00em}{\ttfamily - yellow trapezoid to red bin: 1/10 strong accept, 9/10 weak accept, 0/10 weak reject, 0/10 strong reject}\par
\noindent\hspace*{0.00em}{\ttfamily - yellow polygon to red bin: 0/10 strong accept, 2/10 weak accept, 8/10 weak reject, 0/10 strong reject}\par
\par\smallskip
\noindent\hspace*{0.00em}{\normalfont\bfseries - The unchosen actions:}\par
\noindent\hspace*{0.64em}{\ttfamily Your unchosen actions: PLACE blue\_polygon left, PLACE blue\_polygon red, WAIT}\par
\noindent\hspace*{0.64em}{\ttfamily Human unchosen actions: PICK pink\_square mid, PICK yellow\_polygon right, PICK blue\_trapezoid mid, PICK yellow\_trapezoid right, PICK pink\_trapezoid mid, WAIT}\par
\par\smallskip
\noindent\hspace*{0.00em}{\normalfont\bfseries Now, based on the collaboration history (states, actions, human feedback), and the latest belief over the human's preference before the CURRENT step, predict the human feedback for each action that you and the human did not choose, in the CURRENT step only, considering both the task goal and the human's preference.}\par
\par\smallskip
\noindent\hspace*{0.00em}{\normalfont\bfseries Label each unchosen action as:}\par
\noindent\hspace*{0.00em}{\ttfamily - accepted: accepted behavior with respect to both task-goal progress and inferred human preference.}\par
\noindent\hspace*{0.00em}{\ttfamily - rejected: rejected behavior with respect to task-goal progress and/or inferred human preference.}\par
\par\smallskip
\noindent\hspace*{0.00em}{\normalfont\bfseries Rules:}\par
\noindent\hspace*{0.00em}{\ttfamily 1) Only include unique action keys from the provided unchosen actions of you ("Robot") and the human ("Human").}\par
\noindent\hspace*{0.00em}{\ttfamily 2) Every label must be exactly 'accepted' or 'rejected'.}\par
\noindent\hspace*{0.00em}{\ttfamily 3) Return JSON only.}\par
\par\smallskip
\noindent\hspace*{0.00em}{\normalfont\bfseries Return STRICT JSON only with this shape:}\par
\noindent\hspace*{0.00em}{\ttfamily \{}\par
\noindent\hspace*{0.64em}{\ttfamily "Robot": \{}\par
\noindent\hspace*{1.28em}{\ttfamily "PLACE blue\_polygon left": "accepted\textbar{}rejected",}\par
\noindent\hspace*{1.28em}{\ttfamily "PLACE blue\_polygon red": "accepted\textbar{}rejected",}\par
\noindent\hspace*{1.28em}{\ttfamily "WAIT": "accepted\textbar{}rejected"}\par
\noindent\hspace*{0.64em}{\ttfamily \},}\par
\noindent\hspace*{0.64em}{\ttfamily "Human": \{}\par
\noindent\hspace*{1.28em}{\ttfamily "PICK pink\_square mid": "accepted\textbar{}rejected",}\par
\noindent\hspace*{1.28em}{\ttfamily "PICK yellow\_polygon right": "accepted\textbar{}rejected",}\par
\noindent\hspace*{1.28em}{\ttfamily "PICK blue\_trapezoid mid": "accepted\textbar{}rejected",}\par
\noindent\hspace*{1.28em}{\ttfamily "PICK yellow\_trapezoid right": "accepted\textbar{}rejected",}\par
\noindent\hspace*{1.28em}{\ttfamily "PICK pink\_trapezoid mid": "accepted\textbar{}rejected",}\par
\noindent\hspace*{1.28em}{\ttfamily "WAIT": "accepted\textbar{}rejected"}\par
\noindent\hspace*{0.64em}{\ttfamily \}}\par
\noindent\hspace*{0.00em}{\ttfamily \}}\par
\endgroup
\end{promptbox}

\section{Details of the Real-World Human-Robot Collaboration Study}
\label{sec:real_world_details}

Prior to starting data collection, the study protocol was submitted to our local Institutional Review Board, and was exempt from standard, full-board review because it involved a benign behavioral intervention with adult subjects, who had to consent to taking part in the research.

\subsection{Study Setup and Procedure}

\begin{figure}[t]
    \centering
    \includegraphics[width=0.6\linewidth]{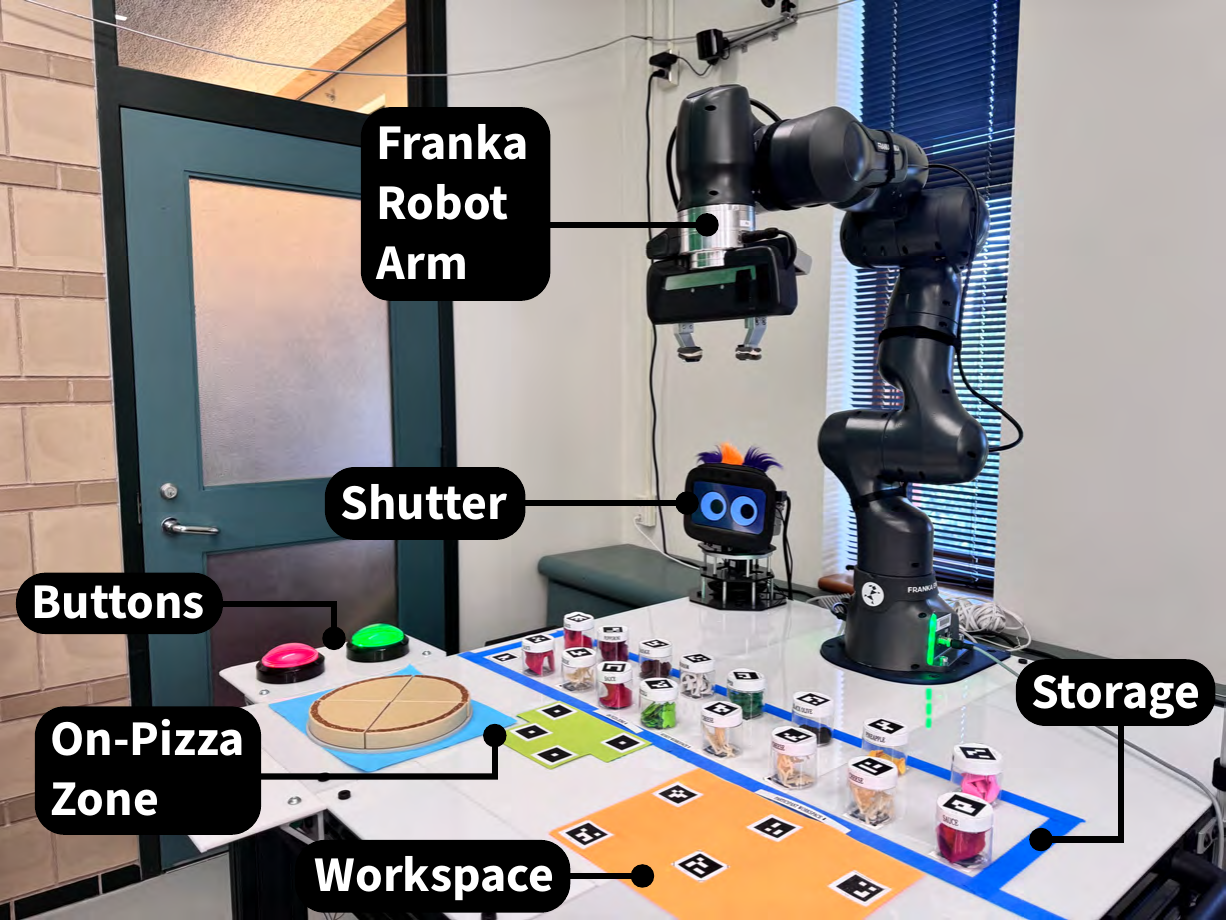}
    \caption{Layout of the tabletop pizza-making station in the physical human-robot collaboration study.}
    \label{fig:physical-pizza-setup}
\end{figure}

Participants collaborated with a robot system to assemble target pizzas at a tabletop station
(Fig.~\ref{fig:physical-pizza-setup}). The station contained three task areas: storage, workspace,
and on-pizza zone. Storage contained all candidate ingredients; the workspace served as the handoff
area between the robot and participant; and the on-pizza zone recorded ingredients that the
participant had already used. The robot's role in the collaboration was to choose ingredients from storage and place them
in the workspace. The participant's role was to decide which workspace ingredients to use for the
target pizza and when to use them. The workspace and on-pizza zone shared a capacity of four items,
so when these areas were full the robot first had to clear an item back to storage before it could
pass another ingredient.

The robot system included two physical robots with distinct roles. The Franka Emika Panda arm
handled manipulation of the ingredient containers, with pick-and-place motions generated through the
MoveIt Task Constructor framework~\cite{gorner2019moveit}. Shutter~\cite{thompson2024shutter}, a tabletop social robot, provided speech
and gaze cues during the interaction. Behavior trees~\cite{colledanchise2018behavior} coordinated the robots' task-level and
communicative actions.

The explicit feedback interface consisted of a green button for positive feedback and a red button for
negative feedback. Participants were instructed to base their feedback on both parts of the task: whether
the robot's behavior helped complete the target recipe and whether it respected their selected
preference over how the pizza should be made. Participants were free to provide feedback whenever
they considered it useful; they were not required to press a button at every step.

A study session began after participants consented to take part in the study and to be audio- and
video-recorded. Participants then completed a practice pizza-making task. The practice pizza contained two
cheeses and two sauces, and the practice preference was to use cheese before sauce. The practice task served to 
familiarize participants with the physical task, the handoff protocol, the capacity constraint of the shared workspace,
and the feedback buttons.
The practice was also used to fit participant-specific rationality coefficients for inferring preferences from implicit
and explicit human feedback, \(\hat{\beta}_{\mathrm{imp}}\) and
\(\hat{\beta}_{\mathrm{exp}}\) in Eq.~(\ref{eq:MLE-objective}), which were then used in the following pizza-making tasks for that participant.

After practice, participants experienced two experimental conditions: preference learning per Eq.~(\ref{eq:MLE-objective}) with \implied{}, and preference learning per Eq.~(\ref{eq:MLE-objective})  with the Fixed-Rule implication. The order of the conditions was counterbalanced between participants. 

For each condition, the participants completed two pizza-making tasks (i.e., two trials) sequentially. The
same pizza recipes were used across both conditions, but the robot's learned preference estimate
was reset between the conditions. Participants were not told which implication for preference learning was being used by the robot in a given
condition; they only knew that the robot was trying to learn their preferences during the collaboration using two learning methods. After completing both conditions, participants answered comparison questions about the two
learning methods.

Each lab session lasted about 45 minutes, and participants were compensated US\$15. Eligibility
required participants to be adults, fluent in English, and able to perceive the robot's speech and
visual cues with normal or corrected-to-normal hearing and vision.

\subsection{Pizza-Making Task Specification}

\textbf{Domain basics.}
The physical pizza-making task uses the same high-level collaboration structure as the pizza domain
described by ~\citet{candon2026learning}. There are 16 ingredients, divided into four broad ingredient types:
\begin{itemize}[wide]
    \item Sauce (4): \texttt{sauce1}, \texttt{sauce2}, \texttt{sauce3}, \texttt{sauce4}
    \item Cheese (4): \texttt{cheese1}, \texttt{cheese2}, \texttt{cheese3}, \texttt{cheese4}
    \item Meat (3): \texttt{ham}, \texttt{pepperoni}, \texttt{sausage}
    \item Veggie (5): \texttt{broccoli}, \texttt{greenpepper}, \texttt{blackolive}, \texttt{mushroom}, \texttt{pineapple}
\end{itemize}
The robot can access \texttt{storage} and the \texttt{workspace}, while the human can access the
\texttt{workspace} and \texttt{on-pizza} zone. A target recipe specifies how many ingredients of
each goal class are needed on the pizza. Cheese and sauce are counted by class, while the meat and
veggie ingredients are counted by their specific types (e.g., \texttt{ham}, \texttt{pepperoni},
\texttt{pineapple}). In each trial corresponding to making a given pizza, the robot was given the target recipe but did not know the
participant's true preference. The participant knew both.

\textbf{State space.}
The task state is represented by the allocation of ingredients across storage, the workspace, the
on-pizza zone, and the two agents' hands. Together with the target recipe, this allocation determines
which goal-class counts are satisfied, which ingredients are available to the robot or human,
whether either agent is holding an ingredient, and whether the shared workspace/on-pizza capacity is
active.

\textbf{Action space and transitions.}
The robot's action space has 19 actions in total: 16 \texttt{PICK(ingredient)} actions, \texttt{PUT(ingredient, storage)},
\texttt{PUT(ingredient, workspace)}, and \texttt{WAIT}. If empty-handed, the robot could pick an
ingredient from storage or the workspace; if holding an ingredient, it could put the ingredient in
storage or, when space was available, in the workspace. The human's action space has 33 actions in total: 16
\texttt{PICK(ingredient)} actions from the workspace, 16 \texttt{PLACE(ingredient)} actions that
mark a held ingredient as added to the pizza, and \texttt{WAIT}. At each step, the robot and human
act simultaneously. A pick action transfers an ingredient to the acting agent's hand; a robot put
action transfers a held ingredient to storage or the workspace; and a human place action marks a
held ingredient as used on the pizza.

\textbf{Preference space.}
Human preferences are represented by an 8-dimensional weight vector,
\[
\begin{aligned}
\mathbf{w} = [&\mathtt{cheese\_before\_sauce},\ \mathtt{cheese\_before\_meat},\\
     &\mathtt{cheese\_before\_veggie},\ \mathtt{sauce\_before\_meat},\\
     &\mathtt{sauce\_before\_veggie},\ \mathtt{meat\_before\_veggie},\\
     &\mathtt{workspace\_size\_le1},\ \mathtt{workspace\_size\_le2}]
\end{aligned}
\]
The first six dimensions encode pairwise ordering preferences over cheese, sauce, meat, and veggie.
For example, a positive value of \texttt{cheese\_before\_sauce} means that all target cheese should
be used before any target sauce, while a negative value represents the reverse ordering. The last
two dimensions encode preferences over the number of ingredients in the workspace/on-pizza area.
Weights are normalized to unit Euclidean norm.

\textbf{True preference options and environment initializations.}
Before the main trials, participants selected one of six true preference options and used it
consistently throughout the pizza-making task:
\begin{itemize}[wide]
    \item \textit{Preference A:} Sauce before cheese before veggie before meat.
    \item \textit{Preference B:} Sauce before meat before veggie.
    \item \textit{Preference C:} Cheese before meat before sauce and veggie.
    \item \textit{Preference D:} Veggie before meat.
    \item \textit{Preference E:} Prefer more than two ingredients in the workspace/on-pizza area.
    \item \textit{Preference F:} Prefer at most two ingredients in the workspace/on-pizza area.
\end{itemize}
The main trials used target recipes drawn from the following set:
\begin{itemize}[wide]
    \item \textit{Combo Pizza:} 1 cheese, 1 sauce, 1 pepperoni, 1 ham, 1 green pepper, 1 black olive.
    \item \textit{Hawaiian Pizza:} 1 pineapple, 1 green pepper, 1 ham, 1 pepperoni, 1 cheese, 2 sauce.
    \item \textit{Deluxe Pizza:} 1 broccoli, 1 black olive, 1 sausage, 1 pepperoni, 2 cheese, 1 sauce.
    \item \textit{Sausage Pizza:} 2 cheese, 2 sauce, 1 sausage, 1 green pepper, 1 mushroom.
    \item \textit{Meat Pizza:} 1 sausage, 1 ham, 1 pepperoni, 2 cheese, 2 sauce.
\end{itemize}

\textbf{Reward function.}
The shared reward is decomposed into goal and preference terms:
\[
R(s,a_R,a_H) = R_{\mathrm{goal}}(s,a_R,a_H) +
\gamma R_{\mathrm{pref}}(s,a_R,a_H),
\]
where \(\gamma\) controls the relative importance of the preference reward and goal reward. The goal
term is the sum of agent-specific rewards. Each agent receives a step cost of \(-0.3\), and
completing the pizza gives a \(+10\) bonus. The robot is rewarded for making target ingredients
available to the human: picking a still-needed target ingredient from storage gives \(+2\), and
placing such an ingredient from storage into the workspace gives \(+3\). The robot is penalized for
actions that do not advance the recipe: picking a needed target that is already in the workspace or
on the human side gives \(-2\) (or \(+0.5\) when the shared capacity is full), picking from storage
when the shared capacity is full gives \(-5\), moving a needed target back to storage gives \(-1\),
and cycling a target from the workspace or human side back to the workspace gives \(-4\). Acting on
an already-satisfied target gives \(-3\) or \(-5\) depending on the action, except that clearing an
already-used item from a full workspace gives \(+1\). Acting on a non-target ingredient gives \(-2\)
or \(-10\). Waiting while holding an ingredient, or while a needed target remains unavailable to the
human, gives \(-5\).

The human receives \(+2\) for picking a still-needed target ingredient from the workspace and \(+3\)
for placing it on the pizza. Picking or placing an already-satisfied target gives \(-5\), acting on
a non-target ingredient gives \(-10\), and attempting to place without holding an ingredient gives
\(-2\). These goal rewards encourage the robot to stage useful ingredients for the human and
encourage the human to add only ingredients that still contribute to the target recipe.

The preference reward is linear in interpretable feature vectors:
\[
R_{\mathrm{pref}}(s,a_R,a_H)
= \mathbf{w}^\top \phi_R(s,a_R) + \mathbf{w}^\top \phi_H(s,a_H)
= \mathbf{w}^\top(\phi_R(s,a_R)+\phi_H(s,a_H)).
\]
The feature vectors $\phi_R$ and $\phi_H$ use the same eight dimensions as \(\mathbf{w}\). Ordering features are positive when an
action supports the preferred ingredient ordering, negative when it violates that ordering, and zero
when the action is irrelevant. Workspace-size features evaluate whether the resulting
workspace/on-pizza occupancy is consistent with the selected occupancy preference.

\subsection{Demographic Information}

Participants completed a demographic survey at the beginning of the study. The statistics for the
20 participants are summarized below:
\begin{itemize}[leftmargin=1.5em]
    \item Gender: 16 participants were male, 3 were female, and 1 was non-binary/third gender.
    \item Age: The participants' mean age was 26 years old
    ($\mathrm{SD}=2.1$), with ages ranging from 21 to 31.
    \item Familiarity with robotics: Participants rated their familiarity with robotics on a
    7-point Likert scale, where 1 indicated ``not familiar at all'' and 7 indicated ``very familiar.''
    The mean rating was 4.6 ($\mathrm{SD}=2.01$).
    \item Frequency of interacting with robots: 7 participants reported interacting with
    robots less than once a month, 4 reported once a month, 1 reported once a week, 2 reported
    2--3 times a week, 3 reported 4--6 times a week, and 3 reported daily.
\end{itemize}

%% file: sections/simulation_table_explicit.tex
\definecolor{mplBlue}{rgb}{0.7176, 0.8431, 0.9490}
\definecolor{mplOrange}{rgb}{0.9843, 0.8902, 0.8353}

\begin{figure}[b!p]
  \centering

  \begin{minipage}[t]{0.3\textwidth}
    \centering
    \vspace{0pt}
    \includegraphics[width=0.92\linewidth]{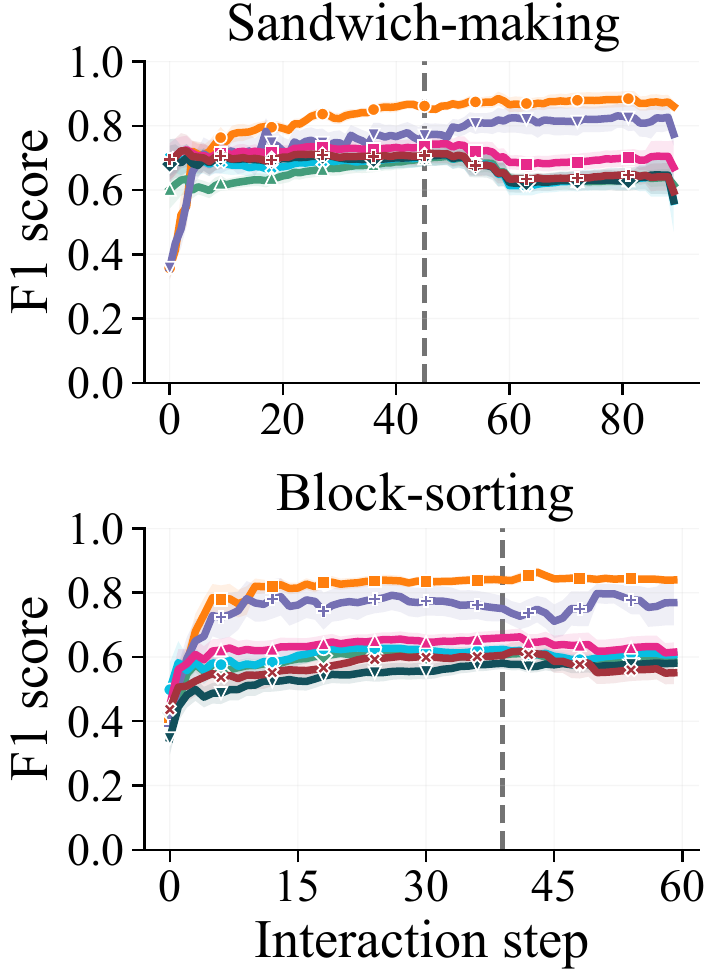}
    \captionof{figure}{
    F1 score for classifying unchosen robot actions over time with \textbf{explicit feedback only.} See first column of Tab.~\ref{tab:downstream_explicit} for legend.
    }
    \label{fig:explicit_f1}
  \end{minipage}
  \hfill
  \begin{minipage}[t]{0.69\textwidth}
    \centering
    \vspace{0pt}
    \captionof{table}{
    Downstream preference-learning performance using \textbf{explicit feedback only}. Learning is achieved per Eq.~(\ref{eq:MLE-objective}) with gradient ascent, but the actions of the human are not considered as evidence for learning. Each row in the table shows results for different implication approaches. Values are mean $\pm$ standard error at the earliest task-completion timestep across participants. \protect\colorbox{mplBlue}{Human} uses ground-truth implication labels for explicit feedback, and \protect\colorbox{mplOrange}{\implied{}} highlights our proposed method.
    }
    \label{tab:downstream_explicit}
    \scriptsize
    \setlength{\tabcolsep}{3pt}
    \begin{adjustbox}{width=\linewidth}
    \begin{tabular}{lcccc}
      \toprule
      & \multicolumn{2}{c}{Sandwich-making}
      & \multicolumn{2}{c}{Block-sorting} \\
      \cmidrule(r){2-3} \cmidrule(r){4-5}
      Implication
      & L2 Error $\downarrow$
      & Conflict Prop. $\downarrow$
      & L2 Error $\downarrow$
      & Conflict Prop. $\downarrow$ \\
      \cmidrule(r){1-1} \cmidrule(r){2-2} \cmidrule(r){3-3} \cmidrule(r){4-4} \cmidrule(r){5-5}

      Human 
      & \cellcolor{mplBlue}$0.83 \pm 0.05$
      & \cellcolor{mplBlue}$0.15 \pm 0.02$
      & \cellcolor{mplBlue}$0.62 \pm 0.06$
      & \cellcolor{mplBlue}$0.15 \pm 0.03$ \\

      IMPLIED \examplesymbol[FF7F0E] 
      & \cellcolor{mplOrange}$0.87 \pm 0.06$
      & \cellcolor{mplOrange}$0.16 \pm 0.02$
      & \cellcolor{mplOrange}$0.66 \pm 0.07$
      & \cellcolor{mplOrange}$0.14 \pm 0.02$ \\

      Fixed-Rule \examplesymbol[449D7B]
      & $1.03 \pm 0.06$
      & $0.25 \pm 0.04$
      & $0.79 \pm 0.06$
      & $0.26 \pm 0.04$ \\

      No-Prior \examplesymbol[7570B3]
      & $1.01 \pm 0.06$
      & $0.26 \pm 0.03$
      & $0.91 \pm 0.08$
      & $0.30 \pm 0.04$ \\

      LLM \examplesymbol[07BEE1]
      & $1.09 \pm 0.07$
      & $0.29 \pm 0.06$
      & $0.85 \pm 0.05$
      & $0.33 \pm 0.04$ \\

      LLM-H \examplesymbol[E7298A]
      & $1.07 \pm 0.08$
      & $0.27 \pm 0.04$
      & $0.78 \pm 0.06$
      & $0.26 \pm 0.04$ \\

      LLM-B \examplesymbol[12505B]
      & $1.08 \pm 0.07$
      & $0.30 \pm 0.05$
      & $0.87 \pm 0.06$
      & $0.31 \pm 0.05$ \\

      LLM-HB \examplesymbol[A4333D]
      & $1.05 \pm 0.08$
      & $0.24 \pm 0.05$
      & $0.84 \pm 0.06$
      & $0.34 \pm 0.04$ \\

      \bottomrule
    \end{tabular}
    \end{adjustbox}
  \end{minipage}
\end{figure}

%% file: sections/simulation_table_implicit.tex
\definecolor{mplBlue}{rgb}{0.7176, 0.8431, 0.9490}
\definecolor{mplOrange}{rgb}{0.9843, 0.8902, 0.8353}

\begin{figure}[tb!p]
  \centering

  \begin{minipage}[t]{0.31\textwidth}
    \centering
    \vspace{0pt}
    \includegraphics[width=0.92\linewidth]{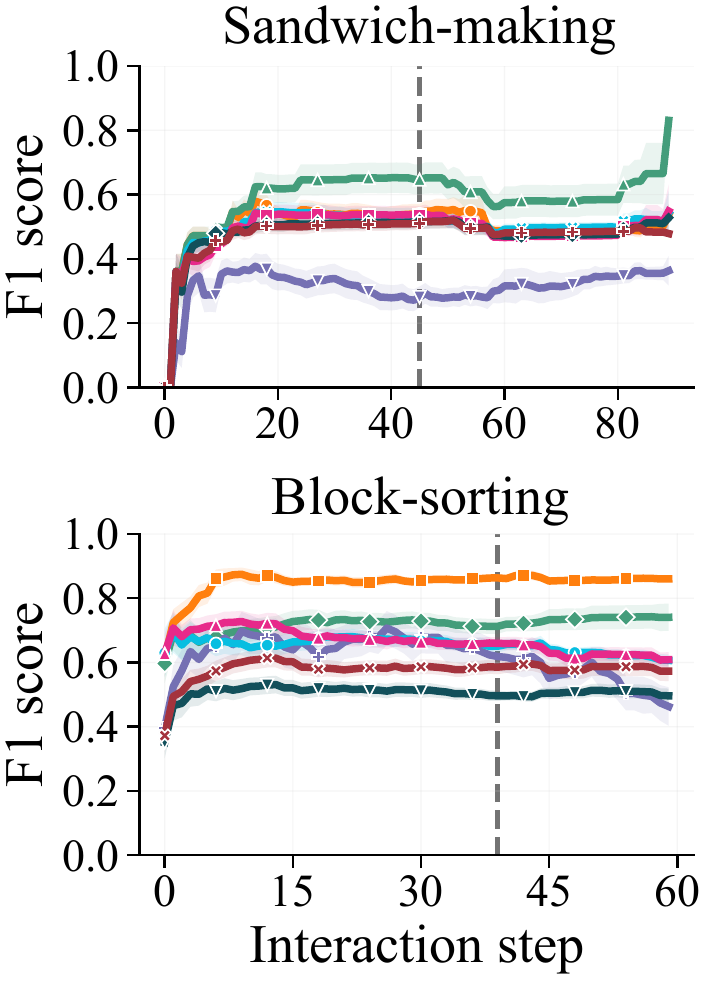}
    \captionof{figure}{
    F1 score for classifying unchosen human actions over time using \textbf{implicit feedback only}. See first column of Tab.~\ref{tab:downstream_implicit} for legend.
    }
    \label{fig:implicit_f1}
  \end{minipage}
  \hfill
  \begin{minipage}[t]{0.68\textwidth}
    \centering
    \vspace{0pt}
    \captionof{table}{
    Downstream preference-learning performance using \textbf{implicit feedback only}. Learning is achieved per Eq.~(\ref{eq:MLE-objective}) with gradient ascent, but no feedback about the robot actions is considered as evidence for learning. Each row in the table shows results for different implication approaches. Values are mean $\pm$ standard error at the earliest task-completion timestep across participants. \protect\colorbox{mplBlue}{Human} uses ground-truth implication labels for implicit feedback, and \protect\colorbox{mplOrange}{\implied{}} highlights our proposed method.
    }
    \label{tab:downstream_implicit}
    \vspace{0.5em}
    \scriptsize
    \setlength{\tabcolsep}{3pt}
    \begin{adjustbox}{width=\linewidth}
    \begin{tabular}{lcccc}
      \toprule
      & \multicolumn{2}{c}{Sandwich-making}
      & \multicolumn{2}{c}{Block-sorting} \\
      \cmidrule(r){2-3} \cmidrule(r){4-5}
      Implication
      & L2 Error $\downarrow$
      & Conflict Prop. $\downarrow$
      & L2 Error $\downarrow$
      & Conflict Prop. $\downarrow$ \\
      \cmidrule(r){1-1} \cmidrule(r){2-2} \cmidrule(r){3-3} \cmidrule(r){4-4} \cmidrule(r){5-5}

      Human 
      & \cellcolor{mplBlue}$0.81 \pm 0.05$
      & \cellcolor{mplBlue}$0.25 \pm 0.03$
      & \cellcolor{mplBlue}$0.53 \pm 0.08$
      & \cellcolor{mplBlue}$0.16 \pm 0.04$ \\

      IMPLIED \examplesymbol[FF7F0E] 
      & \cellcolor{mplOrange}$0.84 \pm 0.05$
      & \cellcolor{mplOrange}$0.25 \pm 0.03$
      & \cellcolor{mplOrange}$0.56 \pm 0.08$
      & \cellcolor{mplOrange}$0.18 \pm 0.05$ \\

      Fixed-Rule \examplesymbol[449D7B]
      & $0.86 \pm 0.06$
      & $0.24 \pm 0.03$
      & $0.66 \pm 0.07$
      & $0.28 \pm 0.06$ \\

      No-Prior \examplesymbol[7570B3]
      & $1.01 \pm 0.07$
      & $0.30 \pm 0.04$
      & $0.74 \pm 0.07$
      & $0.31 \pm 0.05$ \\

      LLM \examplesymbol[07BEE1]
      & $0.82 \pm 0.06$
      & $0.27 \pm 0.03$
      & $0.67 \pm 0.07$
      & $0.26 \pm 0.05$ \\

      LLM-H \examplesymbol[E7298A]
      & $0.85 \pm 0.06$
      & $0.29 \pm 0.04$
      & $0.71 \pm 0.08$
      & $0.26 \pm 0.05$ \\

      LLM-B \examplesymbol[12505B]
      & $0.77 \pm 0.06$
      & $0.26 \pm 0.03$
      & $0.68 \pm 0.07$
      & $0.25 \pm 0.06$ \\

      LLM-HB \examplesymbol[A4333D]
      & $0.83 \pm 0.06$
      & $0.26 \pm 0.03$
      & $0.65 \pm 0.07$
      & $0.27 \pm 0.05$ \\

      \bottomrule
    \end{tabular}
    \end{adjustbox}
  \end{minipage}
\end{figure}

%% file: references.bib
@inproceedings{hadfield2016cooperative,
  title={Cooperative inverse reinforcement learning},
  author={Hadfield-Menell, Dylan and Russell, Stuart J and Abbeel, Pieter and Dragan, Anca},
  booktitle={Advances in neural information processing systems},
  volume={29},
  year={2016}
}

@inproceedings{fitzgerald2022inquire,
  title={INQUIRE: INteractive querying for user-aware informative REasoning},
  author={Fitzgerald, Tesca and Koppol, Pallavi and Callaghan, Patrick and Wong, Russell Quinlan Jun Hei and Simmons, Reid and Kroemer, Oliver and Admoni, Henny},
  booktitle={6th Annual Conference on Robot Learning},
  year={2022}
}

@article{jeon2020reward,
  title={Reward-rational (implicit) choice: A unifying formalism for reward learning},
  author={Jeon, Hong Jun and Milli, Smitha and Dragan, Anca},
  journal={Advances in Neural Information Processing Systems},
  volume={33},
  pages={4415--4426},
  year={2020}
}

@inproceedings{abbeel2004apprenticeship,
  title={Apprenticeship learning via inverse reinforcement learning},
  author={Abbeel, Pieter and Ng, Andrew Y},
  booktitle={Proceedings of the twenty-first international conference on Machine learning},
  pages={1},
  year={2004}
}

@inproceedings{ziebart2008maximum,
  title={Maximum entropy inverse reinforcement learning.},
  author={Ziebart, Brian D and Maas, Andrew L and Bagnell, J Andrew and Dey, Anind K and others},
  booktitle={Aaai},
  volume={8},
  pages={1433--1438},
  year={2008},
  organization={Chicago, IL, USA}
}

@inproceedings{thompson2024shutter,
  author = {Thompson, Sydney and Narcomey, Austin and Lew, Alexander and V\'{a}zquez, Marynel},
  title = {Shutter: A Low-Cost and Flexible Social Robot Platform for In-the-Wild Deployments},
  year = {2024},
  booktitle = {Companion of the 2024 ACM/IEEE International Conference on Human-Robot Interaction},
  pages = {94--96},
  location = {Boulder, CO, USA},
  series = {HRI '24}
}

@inproceedings{losey2018including,
  title={Including uncertainty when learning from human corrections},
  author={Losey, Dylan P and O’Malley, Marcia K},
  booktitle={Conference on Robot Learning},
  pages={123--132},
  year={2018},
  organization={PMLR}
}

@inproceedings{brawerhri23-overlay,
  title={Interactive Policy Shaping for Human-Robot Collaboration with Transparent Matrix Overlays},
  author={Brawer, Jake and Ghose, Debasmita and Candon, Kate and Qin, Meiying and Roncone, Alessandro and V{\'a}zquez, Marynel and Scassellati, Brian},
  booktitle={Proceedings of HRI},
  year={2023}
}

@article{argall2009survey,
  title={A survey of robot learning from demonstration},
  author={Argall, Brenna D and Chernova, Sonia and Veloso, Manuela and Browning, Brett},
  journal={Robotics and autonomous systems},
  volume={57},
  number={5},
  pages={469--483},
  year={2009},
  publisher={Elsevier}
}

@article{christiano2017deep,
  title={Deep reinforcement learning from human preferences},
  author={Christiano, Paul F and Leike, Jan and Brown, Tom and Martic, Miljan and Legg, Shane and Amodei, Dario},
  journal={Advances in neural information processing systems},
  volume={30},
  year={2017}
}

@article{biyik2022learning,
  title={Learning reward functions from diverse sources of human feedback: Optimally integrating demonstrations and preferences},
  author={B{\i}y{\i}k, Erdem and Losey, Dylan P and Palan, Malayandi and Landolfi, Nicholas C and Shevchuk, Gleb and Sadigh, Dorsa},
  journal={The International Journal of Robotics Research},
  volume={41},
  number={1},
  pages={45--67},
  year={2022},
  publisher={SAGE Publications Sage UK: London, England}
}

@inproceedings{hejna2023few,
  title={Few-shot preference learning for human-in-the-loop rl},
  author={Hejna III, Donald Joseph and Sadigh, Dorsa},
  booktitle={Conference on Robot Learning},
  pages={2014--2025},
  year={2023},
  organization={PMLR}
}

@INPROCEEDINGS{mindmeld2022,
  author={Schrum, Mariah L. and Hedlund-Botti, Erin and Moorman, Nina and Gombolay, Matthew C.},
  booktitle={2022 17th ACM/IEEE International Conference on Human-Robot Interaction (HRI)}, 
  title={MIND MELD: Personalized Meta-Learning for Robot-Centric Imitation Learning}, 
  year={2022},
  pages={157--165},
  doi={10.1109/HRI53351.2022.9889616}}

@article{zhang2025predicting,
  title={Predicting Human Perceptions of Robot Performance during Navigation Tasks},
  author={Zhang, Qiping and Tsoi, Nathan and Nagib, Mofeed and Choi, Booyeon and Tan, Jie and Chiang, Hao-Tien Lewis and V{\'a}zquez, Marynel},
  journal={ACM Transactions on Human-Robot Interaction},
  volume={14},
  number={3},
  pages={1--27},
  year={2025},
  publisher={ACM New York, NY}
}

@inproceedings{tamer2009,
author = {Knox, W. Bradley and Stone, Peter},
title = {Interactively shaping agents via human reinforcement: the TAMER framework},
year = {2009},
isbn = {9781605586588},
publisher = {Association for Computing Machinery},
address = {New York, NY, USA},
url = {https://doi.org/10.1145/1597735.1597738},
doi = {10.1145/1597735.1597738},
booktitle = {Proceedings of the Fifth International Conference on Knowledge Capture},
pages = {9--16},
numpages = {8},
location = {Redondo Beach, California, USA},
series = {K-CAP '09}
}

@article{ma2024goal,
  title={Goal inference from open-ended dialog},
  author={Ma, Rachel and Qu, Jingyi and Bobu, Andreea and Hadfield-Menell, Dylan},
  journal={arXiv preprint arXiv:2410.13957},
  year={2024}
}

@article{jain2019probabilistic,
  title={Probabilistic human intent recognition for shared autonomy in assistive robotics},
  author={Jain, Siddarth and Argall, Brenna},
  journal={ACM Transactions on Human-Robot Interaction (THRI)},
  volume={9},
  number={1},
  pages={1--23},
  year={2019},
  publisher={ACM New York, NY, USA}
}

@article{ghose2025ve,
  title={I’ve Changed My Mind: Robots Adapting to Changing Human Goals during Collaboration},
  author={Ghose, Debasmita and Gitelson, Oz and Jin, Ryan and Abawe, Grace and V{\'a}zquez, Marynel and Scassellati, Brian},
  journal={IEEE Robotics and Automation Letters},
  volume={11},
  number={2},
  pages={1490--1497},
  year={2025},
  publisher={IEEE}
}

@inproceedings{ghose2026open,
  title={Open-Ended Goal Inference through Actions and Language for Human-Robot Collaboration},
  author={Ghose, Debasmita and Gitelson, Oz and Vazquez, Marynel and Scassellati, Brian},
  booktitle={In Proceedings of the 2026 ACM/IEEE International Conference on Human-Robot Interaction (HRI '26)},
  year={2026}
}

@inproceedings{candon2026learning,
  title={Learning human preferences over a human-robot collaboration based on explicit and implicit human feedback},
  author={Candon, Kate and Zhang, Qiping and Lew, Alexander and Claure, Houston and Qian, Lena and Quarles, Alyssa and Sarkar, Chayan and V{\'a}zquez, Marynel},
  booktitle={Proceedings of the 21st ACM/IEEE International Conference on Human-Robot Interaction},
  pages={1040--1049},
  year={2026}
}

@inproceedings{mandi2024roco,
  title={Roco: Dialectic multi-robot collaboration with large language models},
  author={Mandi, Zhao and Jain, Shreeya and Song, Shuran},
  booktitle={2024 IEEE International Conference on Robotics and Automation (ICRA)},
  pages={286--299},
  year={2024},
  organization={IEEE}
}

@inproceedings{agrawal2022task,
  title={The task specification problem},
  author={Agrawal, Pulkit},
  booktitle={Conference on Robot Learning},
  pages={1745--1751},
  year={2022},
  organization={PMLR}
}

@InProceedings{pmlr-v70-macglashan17a,
  title = 	 {Interactive Learning from Policy-Dependent Human Feedback},
  author =       {James MacGlashan and Mark K. Ho and Robert Loftin and Bei Peng and Guan Wang and David L. Roberts and Matthew E. Taylor and Michael L. Littman},
  booktitle = 	 {Proceedings of the 34th International Conference on Machine Learning},
  pages = 	 {2285--2294},
  year = 	 {2017},
  editor = 	 {Precup, Doina and Teh, Yee Whye},
  volume = 	 {70},
  series = 	 {Proceedings of Machine Learning Research},
  month = 	 {06--11 Aug},
  publisher =    {PMLR},
}

@inproceedings{ng2000algorithms,
  title={Algorithms for inverse reinforcement learning.},
  author={Ng, Andrew Y and Russell, Stuart and others},
  booktitle={Icml},
  volume={1},
  number={2},
  pages={2},
  year={2000}
}

@inproceedings{russell1998learning,
  title={Learning agents for uncertain environments},
  author={Russell, Stuart},
  booktitle={Proceedings of the eleventh annual conference on Computational learning theory},
  pages={101--103},
  year={1998}
}

@article{billard2016learning,
  title={Learning from humans},
  author={Billard, Aude G and Calinon, Sylvain and Dillmann, R{\"u}diger},
  journal={Springer handbook of robotics},
  pages={1995--2014},
  year={2016},
  publisher={Springer}
}

@article{mackay1998introduction,
  title={Introduction to Gaussian processes},
  author={MacKay, David JC and others},
  journal={NATO ASI series F computer and systems sciences},
  volume={168},
  pages={133--166},
  year={1998},
  publisher={Springer Verlag}
}

@book{williams2006gaussian,
  title={Gaussian processes for machine learning},
  author={Williams, Christopher KI and Rasmussen, Carl Edward},
  volume={2},
  number={3},
  year={2006},
  publisher={MIT press Cambridge, MA}
}

@INPROCEEDINGS{dennler2025hri,
  author={Dennler, Nathaniel and Nikolaidis, Stefanos and Matarić, Maja},
  booktitle={2025 20th ACM/IEEE International Conference on Human-Robot Interaction (HRI)}, 
  title={Contrastive Learning from Exploratory Actions: Leveraging Natural Interactions for Preference Elicitation}, 
  year={2025},
  volume={},
  number={},
  pages={778-788},
  }

@inproceedings{zhu2026interpret,
  title={InterPReT: Interactive Policy Restructuring and Training Enable Effective Imitation Learning from Laypersons},
  author={Zhu, Feiyu Gavin and Oh, Jean and Simmons, Reid},
  booktitle={Proceedings of the 21st ACM/IEEE International Conference on Human-Robot Interaction},
  pages={864--873},
  year={2026}
}

@inproceedings{ghose2023tailoring,
  title={Tailoring visual object representations to human requirements: A case study with a recycling robot},
  author={Ghose, Debasmita and Lewkowicz, Michal Adam and Gezahegn, Kaleb and Lee, Julian and Adamson, Timothy and V{\'a}zquez, Marynel and Scassellati, Brian},
  booktitle={Conference on Robot Learning},
  pages={583--593},
  year={2023},
  organization={PMLR}
}

@inproceedings{wang2025effects,
  title={Effects of robot competency and motion legibility on human correction feedback},
  author={Wang, Shuangge and Wang, Anjiabei and Goncharova, Sofiya and Scassellati, Brian and Fitzgerald, Tesca},
  booktitle={2025 20th ACM/IEEE International Conference on Human-Robot Interaction (HRI)},
  pages={789--799},
  year={2025},
  organization={IEEE}
}

@inproceedings{akrour2011preference,
  title={Preference-based policy learning},
  author={Akrour, Riad and Schoenauer, Marc and Sebag, Michele},
  booktitle={Joint European Conference on Machine Learning and Knowledge Discovery in Databases},
  pages={12--27},
  year={2011},
  organization={Springer}
}

@article{wilde2020improving,
  title={Improving user specifications for robot behavior through active preference learning: Framework and evaluation},
  author={Wilde, Nils and Blidaru, Alexandru and Smith, Stephen L and Kuli{\'c}, Dana},
  journal={The International Journal of Robotics Research},
  volume={39},
  number={6},
  pages={651--667},
  year={2020},
  publisher={Sage Publications Sage UK: London, England}
}

@inproceedings{wang2022skill,
  title={Skill preferences: Learning to extract and execute robotic skills from human feedback},
  author={Wang, Xiaofei and Lee, Kimin and Hakhamaneshi, Kourosh and Abbeel, Pieter and Laskin, Michael},
  booktitle={Conference on robot learning},
  pages={1259--1268},
  year={2022},
  organization={PMLR}
}

@article{hullermeier2021aleatoric,
  title={Aleatoric and epistemic uncertainty in machine learning: An introduction to concepts and methods},
  author={H{\"u}llermeier, Eyke and Waegeman, Willem},
  journal={Machine learning},
  volume={110},
  number={3},
  pages={457--506},
  year={2021},
  publisher={Springer}
}

@article{matthews2017gpflow,
  title={GPflow: A Gaussian process library using TensorFlow},
  author={Matthews, Alexander G de G and Van Der Wilk, Mark and Nickson, Tom and Fujii, Keisuke and Boukouvalas, Alexis and Le{\'o}n-Villagr{\'a}, Pablo and Ghahramani, Zoubin and Hensman, James},
  journal={Journal of Machine Learning Research},
  volume={18},
  number={40},
  pages={1--6},
  year={2017}
}

@inproceedings{bajcsy2018learning,
  title={Learning from physical human corrections, one feature at a time},
  author={Bajcsy, Andrea and Losey, Dylan P and O'Malley, Marcia K and Dragan, Anca D},
  booktitle={Proceedings of the 2018 ACM/IEEE International Conference on Human-Robot Interaction},
  pages={141--149},
  year={2018}
}

@inproceedings{cui2021empathic,
  title={The empathic framework for task learning from implicit human feedback},
  author={Cui, Yuchen and Zhang, Qiping and Knox, Brad and Allievi, Alessandro and Stone, Peter and Niekum, Scott},
  booktitle={Conference on robot learning},
  pages={604--626},
  year={2021},
  organization={PMLR}
}

@article{loftin2016learning,
  title={Learning behaviors via human-delivered discrete feedback: modeling implicit feedback strategies to speed up learning},
  author={Loftin, Robert and Peng, Bei and MacGlashan, James and Littman, Michael L and Taylor, Matthew E and Huang, Jeff and Roberts, David L},
  journal={Autonomous agents and multi-agent systems},
  volume={30},
  number={1},
  pages={30--59},
  year={2016},
  publisher={Springer}
}

@inproceedings{wang2026enhancing,
  title={Enhancing Goal Inference via Correction Timing},
  author={Wang, Anjiabei and Wang, Shuangge and Fitzgerald, Tesca},
  booktitle={Proc. of the 25th International Conference on Autonomous Agents and Multiagent Systems},
  pages={2744--2753},
  year={2026}
}

@article{fox2006robot,
  title={Robot introspection through learned hidden markov models},
  author={Fox, Maria and Ghallab, Malik and Infantes, Guillaume and Long, Derek},
  journal={Artificial Intelligence},
  volume={170},
  number={2},
  pages={59--113},
  year={2006},
  publisher={Elsevier}
}

@article{liang2024introspective,
  title={Introspective planning: Aligning robots' uncertainty with inherent task ambiguity},
  author={Liang, Kaiqu and Zhang, Zixu and Fisac, Jaime F},
  journal={Advances in Neural Information Processing Systems},
  volume={37},
  pages={71998--72031},
  year={2024}
}

@inproceedings{daftry2016introspective,
  title={Introspective perception: Learning to predict failures in vision systems},
  author={Daftry, Shreyansh and Zeng, Sam and Bagnell, J Andrew and Hebert, Martial},
  booktitle={2016 IEEE/RSJ international conference on intelligent robots and systems (IROS)},
  pages={1743--1750},
  year={2016},
  organization={IEEE}
}

@inproceedings{gorner2019moveit,
  title={Moveit! task constructor for task-level motion planning},
  author={G{\"o}rner, Michael and Haschke, Robert and Ritter, Helge and Zhang, Jianwei},
  booktitle={2019 International conference on robotics and automation (ICRA)},
  pages={190--196},
  year={2019},
  organization={IEEE}
}

@book{colledanchise2018behavior,
  title={Behavior trees in robotics and AI: An introduction},
  author={Colledanchise, Michele and {\"O}gren, Petter},
  year={2018},
  publisher={CRC Press}
}

@inproceedings{zhang2023self,
  title={Self-annotation methods for aligning implicit and explicit human feedback in human-robot interaction},
  author={Zhang, Qiping and Narcomey, Austin and Candon, Kate and V{\'a}zquez, Marynel},
  booktitle={Proceedings of the 2023 ACM/IEEE International Conference on Human-Robot Interaction},
  pages={398--407},
  year={2023}
}

@inproceedings{candon2024react,
  title={REACT: Two datasets for analyzing both human reactions and evaluative feedback to robots over time},
  author={Candon, Kate and Georgiou, Nicholas C and Zhou, Helen and Richardson, Sidney and Zhang, Qiping and Scassellati, Brian and V{\'a}zquez, Marynel},
  booktitle={Proceedings of the 2024 ACM/IEEE International Conference on Human-Robot Interaction},
  pages={885--889},
  year={2024}
}

@article{ghose2026robots,
  author = {Debasmita Ghose and Oz Gitelson and Michal Lewkowicz and Jake Brawer and Alessandro Roncone and Marynel Vazquez and Brian Scassellati},
  title = {Robots Influencing Humans to Reveal their Goals during Collaboration and Competition},
  journal = {Autonomous Robots},
  volume = {50},
  number = {3},
  pages = {38},
  year = {2026},
}

@article{zhan2021human,
  title={Human-guided robot behavior learning: A gan-assisted preference-based reinforcement learning approach},
  author={Zhan, Huixin and Tao, Feng and Cao, Yongcan},
  journal={IEEE Robotics and Automation Letters},
  volume={6},
  number={2},
  pages={3545--3552},
  year={2021},
  publisher={IEEE}
}

@article{park2022surf,
  title={Surf: Semi-supervised reward learning with data augmentation for feedback-efficient preference-based reinforcement learning},
  author={Park, Jongjin and Seo, Younggyo and Shin, Jinwoo and Lee, Honglak and Abbeel, Pieter and Lee, Kimin},
  journal={arXiv preprint arXiv:2203.10050},
  year={2022}
}

@article{liu2026learning,
  title={Learning from Noisy Preferences: A Semi-Supervised Learning Approach to Direct Preference Optimization},
  author={Liu, Xinxin and Li, Ming and Lyu, Zonglin and Shang, Yuzhang and Chen, Chen},
  journal={arXiv preprint arXiv:2604.24952},
  year={2026}
}

@inproceedings{brown2020better,
  title={Better-than-demonstrator imitation learning via automatically-ranked demonstrations},
  author={Brown, Daniel S and Goo, Wonjoon and Niekum, Scott},
  booktitle={Conference on robot learning},
  pages={330--359},
  year={2020},
  organization={PMLR}
}

@inproceedings{chen2021learning,
  title={Learning from suboptimal demonstration via self-supervised reward regression},
  author={Chen, Letian and Paleja, Rohan and Gombolay, Matthew},
  booktitle={Conference on robot learning},
  pages={1262--1277},
  year={2021},
  organization={PMLR}
}
